\documentclass[10pt,twocolumn,letterpaper]{article}

\usepackage[pagenumbers]{wacv} 

\usepackage[dvipsnames]{xcolor}

\usepackage{dsfont}
\usepackage{etoolbox}
\usepackage{color}

\newif\ifshowedits

\newcommand{\addeditor}[3]{%
  \definecolor{#1color}{rgb}{#3}
  \expandafter\newcommand\csname #1\endcsname[1]{%
  \ifshowedits
    {\color{#1color} ##1}%
  \else
    {##1}%
  \fi
  }%
  \expandafter\newcommand\csname #1rmk\endcsname[1]{%
  \ifshowedits
    {\color{#1color} {\bf [#2: ##1]}}
  \fi
  }%
  \expandafter\newcommand\csname #1rpl\endcsname[2]{%
  \ifshowedits
    {\color{#1color} ##1 \sout{##2}}
  \else
    {##1}
  \fi
  }%
}

\newcommand{\createtextvar}[1]{
  \expandafter\newcommand\csname #1\endcsname{%
  {\text{#1}}
}%
}
\newcommand{\mycomment}[1]{}

\newcommand{\calT}{{\cal T}}

\newcommand{\calV}{{\cal V}}

\DeclareMathOperator*{\argmin}{arg\,min}

\newcommand{\SE}{{\text{SE}}}

\addeditor{vincent}{VL}{0.0, 0.5, 0.0}
\addeditor{steve}{SB}{0.5, 0.5, 0.0}
\addeditor{boris}{BM}{0.7, 0, 0.5}
\addeditor{fabrice}{FM}{0, 0.5, 0.5}
\addeditor{asma}{AB}{1, 0., 0.7}
\showeditstrue
\showeditsfalse

\usepackage{colortbl}
\usepackage{multirow}
\usepackage{listings}

\definecolor{codegreen}{rgb}{0,0.6,0}
\definecolor{codegray}{rgb}{0.5,0.5,0.5}
\definecolor{codepurple}{rgb}{0.58,0,0.82}
\definecolor{backcolour}{rgb}{0.95,0.95,0.92}

\lstdefinestyle{mystyle}{
    backgroundcolor=\color{backcolour},   
    commentstyle=\color{codegreen},
    keywordstyle=\color{magenta},
    numberstyle=\tiny\color{codegray},
    stringstyle=\color{codepurple},
    basicstyle=\ttfamily\footnotesize,
    breakatwhitespace=false,         
    breaklines=true,                 
    captionpos=b,                    
    keepspaces=true,                 
    numbers=left,                    
    numbersep=5pt,                  
    showspaces=false,                
    showstringspaces=false,
    showtabs=false,                  
    tabsize=2,
    escapechar=\%
}

\newcommand{\epsSym}{\epsilon\text{-sym}}

\usepackage{pifont}

\definecolor{wacvblue}{rgb}{0.21,0.49,0.74}
\usepackage[pagebackref,breaklinks,colorlinks,allcolors=wacvblue]{hyperref}

\def\wacvPaperID{179} 
\def\confName{WACV}
\def\confYear{2026}

\title{
BOP-Distrib: Revisiting 6D Pose Estimation Benchmarks for\\ Better Evaluation under Visual Ambiguities
}

\newcommand{\namesep}{\hspace{0.8em}}

\author{
Boris Meden$^{1}$\namesep
Asma Brazi$^{1,2}$\namesep
Fabrice Mayran de Chamisso$^{1}$\namesep
Steve Bourgeois$^{1}$\namesep
Vincent Lepetit$^{2}$\\
{$^{1}$Université Paris-Saclay, CEA, List}\\{$^{2}$LIGM, Ecole des Ponts, Univ Gustave Eiffel, CNRS, Marne-la-vallée, France}\\
\url{https://cea-list.github.io/BOP-Distrib/}\\[-0.2cm]
}

\begin{document}

\twocolumn[{
\maketitle
\begin{center}
    \captionsetup{type=figure}
    \includegraphics[width=\textwidth]{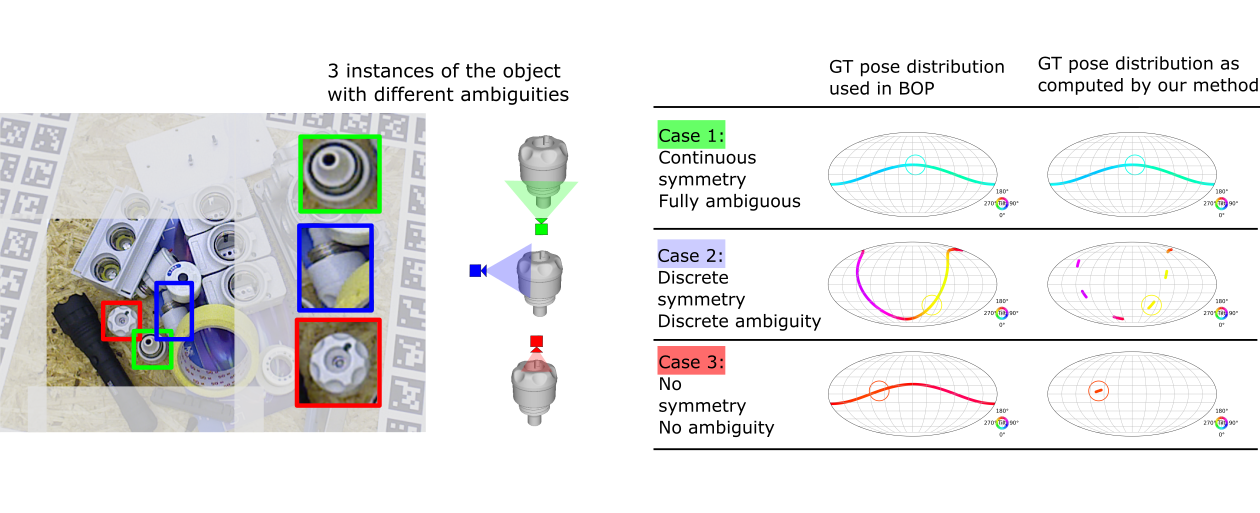}
\vspace{-1cm}
\captionof{figure}{
We provide for the first time 6D pose annotations in the form of a per-image object pose distribution. Current annotations in BOP~\cite{hodan2024bop} datasets are given as a single pose, shown here as a circle in the SO(3) representations. BOP also provides a symmetry pattern per object, from which a distribution can be computed~(the colored points in SO(3)). Such distribution however does not cover many cases~\cite{manhardt2019explaining}: In this example, when only the core is visible (\colorbox{green!60}{Case 1}), the pose is fully ambiguous and should be represented by a continuous distribution in SO(3). When the sides of the head are visible~(\colorbox{blue!20}{Case 2}), there are still ambiguities and the distribution is made of 6 modes. When the hole is visible~(\colorbox{red!60}{Case 3}), the pose distribution should be concentrated around one non-ambiguous pose. Our method annotates scenes with \emph{per-image} distributions, taking into account the partial occlusions and allowing us to evaluate a predicted pose properly. We show that considering these distributions for evaluation results in a significant change of ranking for the BOP challenge. Such ground truth distributions also become a key asset when it comes to evaluating pose distribution estimation methods~\cite{haugaard2023spyropose, hsiao2024confronting}. With appropriate metrics, we demonstrate the first quantitative evaluation of pose distribution methods on real images, as an extension to single pose methods.
}
  \label{fig:teaser}
\end{center}
}]
\maketitle

\begin{abstract}
6D pose estimation aims at determining the object pose that best explains the camera observation. The unique solution for non-ambiguous objects can turn into a multi-modal pose distribution for symmetrical objects or when occlusions 
of symmetry-breaking elements happen, depending on the viewpoint.
Currently, 6D pose estimation methods are benchmarked on datasets that consider, for their ground truth annotations, visual ambiguities as only related to global object symmetries, whereas they should be defined per-image to account for the camera viewpoint. We thus first propose an automatic method to re-annotate those datasets with a 6D pose distribution specific to each image, taking into account the object surface visibility in the image to correctly determine the visual ambiguities. Second, given this improved ground truth, we re-evaluate the state-of-the-art single pose methods and show that this greatly modifies the ranking of these methods. Third, as some recent works focus on estimating the complete set of solutions, we derive a precision/recall formulation to evaluate them against our image-wise distribution ground truth, making it the first benchmark for pose distribution methods on real images.
\end{abstract}    
\section{Introduction}
\label{sec:intro}

Visual 6D pose estimation of an object consists in determining the 3D position and orientation of this object with respect to the camera that explains the observed image. It is a key task in many application domains, such as robotics (\eg grasping or manipulation), augmented reality, industrial quality control, etc. In such contexts, datasets of images with ground truth poses are of primary importance, since they can be used to train learning based methods, but first and foremost to benchmark the performances of proposed approaches. The annotation accuracy of reference benchmarks, such as BOP~\cite{sundermeyerBopChallenge20222023} for 6D pose estimation, is thus crucial since it influences the research directions.

However, annotating images of objects with their 6D poses is a complex task. This is especially true when considered objects include symmetrical parts. Indeed, as illustrated in Figure~\ref{fig:teaser}, an image does not necessarily correspond to a single pose solution depending on the viewpoint and occlusions. While symmetrical objects naturally imply multiple solutions, non-symmetrical objects can also yield multiple solutions in case of partial occlusions. Current reference benchmarks~\cite{sundermeyerBopChallenge20222023} ignore these occlusion-induced symmetries, resulting in an imperfect evaluation of the methods. Considering better ground truth distributions results in a significant change of ranking of pose estimation methods.

In this paper, we propose a method that automatically annotates pose distributions for images that take into account ambiguities due to object symmetries but also to partial occlusions. Our method exploits the current annotations, 
 the ground truth visibility masks of the objects, and their 3D models, to produce a non-parametric distribution representation (see Supp. Mat. Section 2.2). Its genericity is demonstrated on T-LESS~\cite{hodanTLESSRGBDDataset2017} and YCB-V~\cite{xiangPoseCNNConvolutionalNeural2018}.
 
We also re-evaluate classical metrics with these annotations to assess the performance of pose estimation methods that return a single pose estimate per image. This yields a strongly modified ranking of current challenger methods. 

Moreover, we also consider the few methods that already predict multi-modal pose distributions~\cite{haugaard2023spyropose,hsiao2024confronting}.

Because of the absence of proper annotations, these methods have been evaluated quantitatively only on synthetic images and only qualitatively on real images. Thanks to our annotations, we are able to provide their first evaluation on real data.

In summary, our contributions are the following:
\begin{enumerate}
\item A novel automatic method computing a multi-modal 6D pose distribution ground truth from a unique ground truth pose and a proposal set of object symmetries;
\item A comprehensive re-evaluation of 6-DOF single pose estimation methods related to the re-annotation of the T-LESS~\cite{hodanTLESSRGBDDataset2017} and YCB-V~\cite{xiangPoseCNNConvolutionalNeural2018} ground truths;
\item A new evaluation framework of 6-DOF pose distribution estimation methods, which makes it the first multi-modal pose distribution benchmarking on real data.
\end{enumerate}

\section{Related Work}
\label{sec:RW}

Object 6D pose estimation aims at determining an object pose that best explains the camera observation.
Multiple equivalent solutions arise when the object is symmetrical or when occlusions (external or self)
prevent the observation of symmetry-breaking elements (\textit{e.g.} for a cup~\cite{hodanEvaluation6DObject2016, manhardt2019explaining}).

While the state of the art mostly focused on estimating a single pose~\cite{drostModelGloballyMatch2010, parkPix2PosePixelWiseCoordinate2019, hodanEPOSEstimating6D2020, labbeCosyPoseConsistentMultiview2020, sundermeyer2020augmented, wangGdrnetGeometryguidedDirect2021, haugaardSurfEmbDenseContinuous2022, su2022zebrapose}, some recent works  focus on estimating the complete set of solutions \cite{manhardt2019explaining, okorn2020learning, murphy2021implicit, klee2022image, mayrandechamissoHSPAHoughSpace2022, hofer2023hyperposepdf, haugaard2023spyropose, hsiao2024confronting, gilitschenski2019deep, deng2022deep, bui20206d, yin2022fishermatch, lee20243d, vutukur2024alignist}. 
As we argue below, to properly evaluate the performance of a method,
ground truth annotations including the complete set of solutions for each test image, and metrics that take into account the multiplicity of the solutions, are then required, even for methods that return a single pose per image.

\paragraph{Datasets and 6D pose annotation.}
Various techniques have been introduced to annotate image datasets with their 6D poses. For synthetic datasets~\cite{bregierSymmetryAwareEvaluation2017, tyree6DoFPoseEstimation2022}, the ground truth pose is directly available, while, for real datasets, this pose is usually determined with the help of user interaction~\cite{drostIntroducingMVTecITODD2017, hodanBOPBenchmark6D2018, doumanoglouRecovering6DObject2016}, markers \cite{hinterstoisserMultimodalTemplatesRealtime2011, hinterstoisserModelBasedTraining2013, brachmann2014learning}, or a robotic arm~\cite{wangPhoCaLMultiModalDataset2022, medenImsold2024, kalra2024towards}, and refined with ICP in the case of an RGB-D camera~\cite{drostIntroducingMVTecITODD2017, hodanBOPBenchmark6D2018, wangPhoCaLMultiModalDataset2022, medenImsold2024}.

However, such annotations only provide a single pose solution while multiple may exist. For fully symmetric objects, the solutions are always related with the same set of rigid transforms, independently of the viewpoint or the occlusions. This set of transforms, also called \textit{symmetries pattern}, can then be pre-computed offline and applied to the initial solution to recover the whole set of solutions~\cite{hodanBOPChallenge20202020}. 

For non-symmetrical objects, recovering the complete set of solutions is much more challenging since symmetries may arise from the non-visibility of disambiguating parts of the object, those non-visibilities being induced by the viewpoint (self-occlusion), by other elements of the scene (external occlusion), or by the lack of image resolution (resolution occlusion). 
The transformation set that relates the different solutions is then specific to each image and cannot be pre-computed from the 3D object model.

To our knowledge, no method has been proposed to determine this ground truth transformation set per image. Instead, the current gold standard in 6D pose estimation benchmarks still consists in approximating this per-image transformation set with a unique global transformation set, usually computed with the same method than fully symmetric object with large enough tolerance to ignore small disambiguating elements (\eg surface deviation tolerance of 1.5cm or 10\% of the object diameter~\cite{hodanBOPChallenge20202020}), and cannot be used to evaluate properly multi-modal pose distribution estimates like~\cite{haugaard2023spyropose,hsiao2024confronting}.

\paragraph{Evaluation metrics. }
Performance evaluation usually differs if the method outputs a single pose or a distribution. 

For single pose estimation, the accuracy is measured through the deviation of the object surface points when transformed by the estimate and by the ground truth pose~(registration error)~\cite{hodanEvaluation6DObject2016}. Depending on the sensor used---RGB or RGB-D---and the targeted application~(\eg robotics or Augmented Reality),  deviations can be measured in 3D space (3D distance) or in the image space~(reprojection error), and the registration error can be considered as the mean or maximal error over the object surface. In case of multi-valued ground truths due to a symmetrically tagged object, the accuracy corresponds to the minimal error with respect to the set of solutions~\cite{hodanEvaluation6DObject2016,xiangPoseCNNConvolutionalNeural2018,tyree6DoFPoseEstimation2022}. In practice, the most commonly used accuracy measures are : 
\begin{enumerate}
\item Average Distance metric (ADD)~\cite{hinterstoisserModelBasedTraining2013}, and variations for symmetrical objects (ADD-S~\cite{xiangPoseCNNConvolutionalNeural2018}, ADD-H~\cite{tyree6DoFPoseEstimation2022}), measure mean Euclidean error between estimated and ground truth surface points, but are replaced by MSSD.
  \item Visual Surface Discrepancy (VSD)~\cite{hodanBOPBenchmark6D2018} measures
misalignment over the visible surface of the object model.
  VSD is more expensive to compute than MSSD and MSPD, and requires a depth image. It is now omitted for new tasks as stated by BOP organizers\footnote{\url{https://bop.felk.cvut.cz/challenges/}}.
  \item Maximum Symmetry-Aware Surface Distance (MSSD)~\cite{hodanBOPChallenge20202020}, similar to ADD, considers the maximal Euclidean error with symmetry management, and provides a 3D error, useful to robotics applications.
  \item Maximum Symmetry-Aware Projection Distance (MSPD)~\cite{hodanBOPChallenge20202020} measures the maximum reprojection error between projections of the ground truth and estimated surface points, with symmetry management.
\end{enumerate}

Based on these accuracy measures, and inspired from the evaluation of the detection methods, single 6D pose estimation methods are usually evaluated through Precision/Recall. In such evaluation frameworks, an estimated pose is considered as correct if its registration error is below a predefined threshold. The precision corresponds to the rate of correct pose estimations, meaning the ratio of the estimated poses that are correct over the number of estimated poses. The recall corresponds to the rate of correctly registered objects, meaning the ratio between the number of object instances in the dataset whose pose was considered as correct over the number of object instances in the dataset.

Regarding the evaluation of multi-modal 6D pose distribution estimation methods, the accuracy is usually measured as a pose error, meaning the minimal deviation between the inferred rotation and translation and the corresponding nearest ground truth pose~\cite{murphy2021implicit,haugaard2023spyropose,iversen2022ki,periyasamy2022learning}.

If the method outputs a probability distribution over the whole pose space, the method is usually evaluated through its spread, corresponding to the expectation of the pose error and the log-likelihood between the inferred distribution and the multi-valued solution~\cite{murphy2021implicit,haugaard2023spyropose,iversen2022ki,periyasamy2022learning,hofer2023hyperposepdf}, meaning the sum of the mean log probability at ground truth solutions. While the spread provides a probabilistic measure of the accuracy, the log-likelihood measures how similar the probability distributions are. This penalizes the method if some ground truth solutions are missing in the estimate.

However, not all multi-modal 6D pose estimation methods actually provide a full probability distribution over the whole pose space. Instead, some methods output a set of poses corresponding to some local maxima of the underlying probability distribution~\cite{hsiao2024confronting}. For such methods, performances can be evaluated through the Precision/Recall of the multi-valuated estimation. A pose estimate is then considered as correct with respect to one ground truth pose if its distance in pose space does not exceed a threshold $\delta$. The Precision for a given image is then defined as the ratio of the number of estimated poses whose distance have at least one ground truth pose that does not exceed the threshold $\delta$ over the total number of estimated poses. The Recall for an object image is then defined as the ratio between the number of ground truth poses whose distance have at least one estimated pose that does not exceed the threshold $\delta$ over the total number of ground truth poses. Similarly to the spread and log likelihood, these precision and recall are related to the multi-modal pose distribution.

Our method combines a ground truth pose and a symmetries pattern to represent the pose distribution for each image. Unlike previous works, the  symmetries pattern is adjusted to each image, taking into account its specific  viewpoint and  occluded objects~(Section~\ref{sec:method}). 
Moreover, whereas single pose and multi-modal distribution pose methods are currently evaluated with non-comparable metrics, we introduce new evaluation metrics for multi-modal methods that are homogeneous with single pose methods~(Section~\ref{sec:distributionComparison}).

\begin{figure*}
  \centering
  \includegraphics[width=0.8\textwidth]{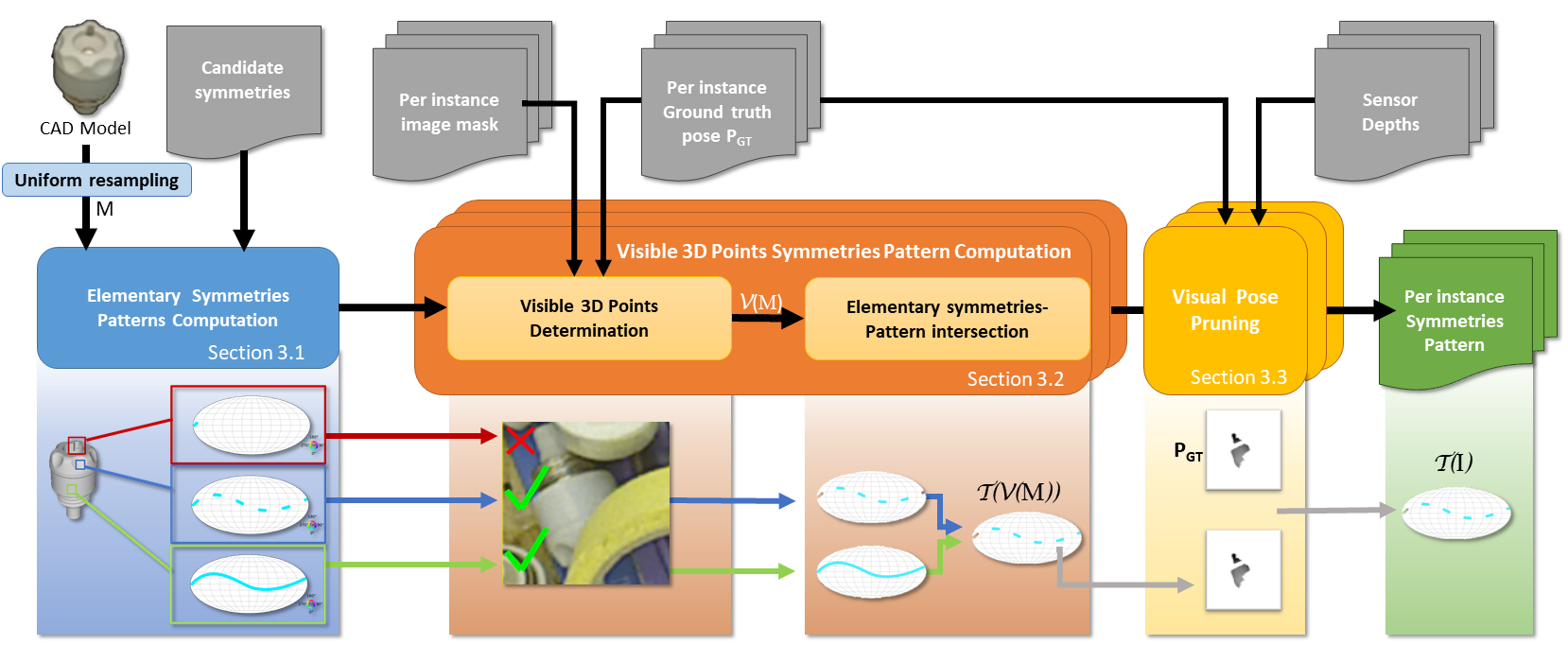}
  \caption{\textbf{Method overview.} From a symmetry candidate set, we pre-compute the object per-vertex $\epsSym$. Then for a given scene, we compute the vertices visibility (\textcolor{PineGreen}{$\checkmark$} and \textcolor{BrickRed}{\ding{55}} illustrate respectively if the visibility test passed or not for the vertex) and perform a robust intersection between their $\epsSym$. This intersection \boris{is then pruned with a depth comparison and the result} constitutes the symmetries pattern of this object instance for this image. When multiplied by the ground truth, we obtain the \SE(3) distribution of the object instance.
  }
  \label{fig:overview}
\end{figure*}

\section{Method}
\label{sec:method}

    Consider a 3D object model $M$ observed from a pose $P_\text{GT}$. We call symmetry-pattern the set $\calT=\{T_i \in \SE(3) \}$ of rigid transformations that, once combined with $P_\text{GT}$, generate representations that are geometrically indistinguishable from the one generated by $P_\text{GT}$. Depending on the object geometry $M$ and the representation type, the set $\calT$ can be limited to the identity transform or can include multiple transformations in addition to the identity.

    Previous work well studied the representation corresponding to a 3D surface, or a subset $\calV(M) \in M$ of the 3D object surface. In such case, the symmetries pattern $\calT(\calV(M))$ can be defined through $\epsilon$-symmetries ~\cite{hodanEvaluation6DObject2016}:
    \begin{align}
    \epsSym(\calV(M),M) = \{T_i : d(T_i*\calV(M),M)< \epsilon \} \> ,
    \label{eq:epsymVM}
    \end{align}
    where $\epsilon$ is the maximal deviation tolerance, $*$ represents the application of the transformation $T_i$ onto the point set $\calV(M)$ and $d$ measures the max deviation between the sets:\\
    $d(X,M) = \max_{x\in X}||x - \argmin_{m\in M}(||x-m||_2)||_2$.

    In this paper, we propose to consider the case of representation corresponding to an image $I$. The symmetries pattern $\calT(I)$ associated to the image $I$ corresponds to a subset of the symmetries pattern $\calT(\calV(M))$ associated to the visible part of the object $\calV(M)$ in the image. Indeed, since $\epsilon$-symmetries work in 3D space, they do not take into account the potential self or external occlusions that can be induced by the projection process that produces the image. 

    Consequently, to estimate $\calT(I)$, our solution computes $\calT(\calV(M))$ then prune the pose visually-inconsistent (Section~\ref{subsec:refine}). As illustrated in Figure~\ref{fig:overview}, to achieve this estimation efficiently, our solution decomposes the estimation of $\calT(\calV(M))$ in a two step process: a first pre-computation (Section~\ref{subsec:accelerate}) independent of $\calV(M)$ and executed only once, and a second step that computes $\calT(\calV(M))$ (Section~\ref{subsec:robustIntersect}).

\subsection{Speeding up the computation of $\epsilon$-sym}
\label{subsec:accelerate}

Computing \steve{$\calT(\calV(M))$}
for each instance in each image of a dataset can quickly become time consuming since it implies to sample the 6D pose space around the ground truth pose. \boris{This can explain why such annotation process was never used to our knowledge.}
We accelerate the $\epsilon$-symmetries computation by introducing \steve{terms independent from $\calV(M)$} that can be pre-computed.

First, by definition of the $\epsilon$-symmetries, for any subsets $V_1 \subset \calV(M)$ and $V_2 \subset \calV(M)$, we have: 
\begin{align}
\epsSym(V_1 \cup V_2,M)= \epsSym(V_1,M) \cap \epsSym(V_2,M) \> .
\end{align}
Therefore, we can reformulate \autoref{eq:epsymVM} as: 
\begin{align}
\epsSym(\calV(M),M) = \bigcap_{v_j \in \calV(M)} \epsSym(v_j,M) \> ,
\label{eq:epssymPrecompute}
\end{align}
where $v_j$ is a 3D point and $\epsSym(v_j,M)$ is the symmetries pattern when the point $v_j$ is the only visible point, which we name \textit{elementary symmetries pattern} of $v_j$.

With such expression of the $\epsilon$-symmetries, it appears that elementary symmetries patterns can be pre-computed for the whole 3D object model $M$ since they are independent of the image. Only the set of visible points and the intersection of the corresponding pre-computed elementary symmetries patterns need to be computed per image.

\subsection{Robust symmetries patterns intersection} \label{subsec:robustIntersect}

In practice, using a strict intersection might reject too many transformations since a transformation $T_i$ would be kept only if it is part of all the elementary symmetries patterns. Instead, we prefer to use a soft intersection, meaning that the transformation $T_i$ will be included even if a few patterns do not include $T_i$. This soft intersection is achieved by counting, for each transformation $T_i$, the number of elementary symmetries patterns that include $T_i$. This counting is defined as follows, in the form of a histogram over $\calT$: 
\begin{multline}
	\forall i, \text{H}(T_i) = \text{Card} \{v_j ~|~ v_j \in \calV(M),  T_{i} \in \epsSym(v_j,M) \} \> . 
	\label{ref:eqHist}
\end{multline}

A strict intersection would only keep the $T_i$ with the maximum value of $\text{H}(T_i)$ which is represented by $\text{H}(Id)$ as the identity of \SE(3) and is always in the symmetries pattern. Our soft intersection tolerates a threshold and keeps all the $T_i$ such that $\text{H}(T_i) > \text{H}(Id) - \tau$, where $\tau$ represents the minimum size (expressed in number of 3D points) of the symmetry-breaking elements.

\subsection{Refining $\epsilon$-sym: visual occlusions consideration
} \label{subsec:refine}
Since $\epsSym$ does not consider visual occlusions, $\calT(\calV(M))$ may contain a small set of pose that could induce more visible part of $M$ than $\calV(M)$ (see Supp. Mat. Section 2.1). 

We follow the VSD metric~\cite{hodanEvaluation6DObject2016} and determine if a pose $T_i$ should be pruned by rendering the depth image $D_{T_i}$. For annotation purposes, the minimal disambiguation element size to reject a candidate pose should be the same for all objects. So we remove VSD dependence to object size and count pixels with a depth deviation to ground truth greater than $\delta$ (mm). If the number of pixels exceeds $\tau_{pix}$, the pose is pruned.

\small
\begin{multline}
Prune
\big(D_{GT}, D_{T_i}, V_{GT}, V_{T_i}, \delta\big) =\\
	\sum_{p \in V_{GT} \cup V_{T_i}}
	\begin{cases}
		0 & \text{if} \; p \in V_{GT} \cap V_{T_i} \, \wedge \, \big|D_{GT}(p) -
		D_{T_i}(p)\big| < \delta \\
		1 & \text{otherwise,}
	\end{cases}
	\label{ref:eqrefine}
\end{multline}
\normalsize

where $D_{GT}$ and $V_{GT}$ are distance image and visible mask of the ground truth, $D_{T_i}$ and $V_{T_i}$ are their $T_i$ counterparts, with visible masks computed from sensor depth.

\subsection{Generalization to textured 3D objects}
\label{subsec:extensionTexture}

While the process previously described considered only texture-less 3D objects, the method can be extended to textured ones by simply redefining the $\epsilon$-symmetries as: 
\begin{multline}
\epsSym(\calV(M),M) = \{T_i : d(T_i*\calV(M),M)< \epsilon,\\ d_\text{color}(T_i*\calV(M),M)<\zeta)\} \> ,
\end{multline}
%
where $\zeta$ is the color deviation tolerance and with:\\
$d_\text{color}(X,M)=\max_{x\in X}d_\text{col}(x, \argmin_{m\in M} (||x-m||_2))$,\\
with $d_\text{col}(x, m)$ the color distance between vertices $x$ and $m$.

\section{Evaluating 6D Pose Distribution Predictions}
\label{sec:distributionComparison}

Currently, single pose estimation methods and distribution pose estimation methods are evaluated on different datasets~(BOP challenge for single poses, SYMSOL for 6D pose distributions) with different metrics (registration error for the first, pose error for the others).
With our annotation method that provides the full set of 6D pose solutions even on real data, it becomes possible to evaluate both tasks on the same dataset. However, it would be beneficial if both tasks can be evaluated with comparable metrics.

We therefore propose to keep the evaluation process of single pose method unchanged, but on a more accurate ground truth, and to extend it to pose distributions evaluation. 
First, whereas pose error is commonly used in 6D pose distribution evaluation process~\cite{murphy2021implicit,iversen2022ki,haugaard2023spyropose}, we propose to use the registration error. The latter takes the object 3D shape into account, making the accuracy more meaningful for most applications. It is also the type of error used by the current gold standard benchmark for single pose~\cite{sundermeyerBopChallenge20222023}.  Typically, we suggest to use the Maximal Surface Distance~(MSD) and Maximum Projective Distance~(MPD) metrics.
Second, we propose to keep the metrics of Precision and Recall, but we adapt them to measure Precision over the whole estimated distribution of poses instead of a unique pose, and Recall over the whole set of solutions instead of considering a pose to be found if one of its multiple solutions was found.

\paragraph{Precision for pose distribution.}
We define the precision in the case of pose distribution prediction as: 
\begin{multline}
\mathbf{P_{d}}(\text{Est}, \text{GT}, \tau_d) =\\ \sum\limits_{P_\text{Est}^i \in \text{Est}}p(P_\text{Est}^i)\left(\min_{P_\text{GT} \in \text{GT}}\mathbf{d}(P_\text{Est}^i, P_\text{GT}) < \tau_{\mathbf{d}}\right) \> ,
\label{eq:precisionDist}
\end{multline}
where $\text{Est}$ and $\text{GT}$ are respectively the estimated and the ground truth distributions to compare, and $p(P_\text{Est}^i)$ represents the probability associated to the $i$-th element of the estimated distribution \footnote{If the method outputs solutions without assigning a probability to each pose, such as~\cite{hsiao2024confronting}, a uniform probability over the solutions is used.}, $\mathbf{d}$ is the chosen registration distance, and $\tau_{\mathbf{d}}$ the associated threshold.

\paragraph{Recall for pose distribution.}
Similarly to the precision, we define the recall in the case of pose distribution prediction as:
 \begin{multline}
    \mathbf{R_d}(\text{Est}, \text{GT}, \tau_d)=\\ \sum\limits_{P_\text{GT}^i \in \text{GT}}\min\left(p(\hat{P}_\text{Est}^i),\frac{1}{\text{Card}(\text{GT})}\right)[\mathbf{d}(\hat{P}_\text{Est}^i, P_\text{GT}^i) < \tau_{\mathbf{d}}],\\
    \text{with }\hat{P}_\text{Est}^i=\argmin_{P_\text{Est} \in \text{Est}} \mathbf{d}(P_\text{Est},P_\text{GT}^i) \> .
    \label{eq:recallDist}
 \end{multline}
The probability of $\hat{P}_\text{Est}^i$ is clamped to $\frac{1}{\text{Card}(\text{GT})}$ since ground truth poses are considered as equiprobable.

\section{Experiments}
\label{sec:exp}

In this section, after evaluating the performances of our annotation method~(Section~\ref{subsec:expValidation}), we present and discuss the impact of our ground truth annotations and evaluation metrics onto the evaluation of state-of-the-art solutions for both single pose estimation~(Section~\ref{subsec:expSinglepose}) and multi-modal pose distribution estimation~(Section~\ref{subsec:expDistpose}). 

Since our annotation process and metrics are related to objects with intrinsic or occlusion-induced symmetries, we perform our evaluations on the T-LESS dataset. Indeed, among the  datasets of the BOP challenge~\cite{hodan2024bop},  T-LESS appears to be the only dataset to exhibit symmetrical objects, with occlusion-induced symmetries, real data and public ground truth annotations:
ITODD~\cite{drostIntroducingMVTecITODD2017} and HB~\cite{kaskmanHomebrewedDBRGBDDataset2019} feature symmetrical objects, with occlusion-induced symmetries, but do not publicly share their ground truths. HOPE~\cite{tyree6DoFPoseEstimation2022}, Omni6D~\cite{zhang2024omni6d} and IC-BIN~\cite{doumanoglouRecovering6DObject2016} have symmetrical shapes, but texture information disambiguates them.

The first experiment focus on a qualitative evaluation of our annotation method~(Section~\ref{subsec:expValidation}). The second evaluates the impact of our re-annotation of T-LESS  and YCB-V datasets on the performance evaluation of the methods of the BOP benchmark~\cite{hodan2024bop} on single pose estimation~(Section~\ref{subsec:expSinglepose}). The last experiment evaluates for the first time 6D pose distribution methods on a non-synthetic dataset with proper annotations and metrics~(Section~\ref{subsec:expDistpose}).

\subsection{Implementation details}

In our experiments, we sample uniformly surface points from the CAD models with a resolution of 0.5mm. For the pre-computation of elementary symmetries patterns for a given object, we use the per-object symmetries pattern given by the BOP challenge~\cite{hodan2024bop} as the initial symmetry candidates, with a tolerance factor $\epsilon\text{-sym}$ set to 1mm. Following BOP, continuous symmetries are discretized such as the furthest vertex from the axis of symmetry travels not more than 1\% of the object diameter.

Surface visibility $\calV(M)$ is computed by Z-buffering using ground truth pose $P_\text{GT}$ and the 3D model of the object. 
For the robust symmetry pattern intersection, the soft intersection tolerance factor $\tau$ was experimentally adjusted to 28 3D points, resulting in a minimal disambiguating size of roughly 2.5$\times$2.5mm$^2$, $\delta=5$mm and $\tau_{pix}=30$pix.

\subsection{Validation of the method}
\label{subsec:expValidation}
\begin{figure*}[h]
\begin{center}
  \includegraphics[width=0.9\textwidth]{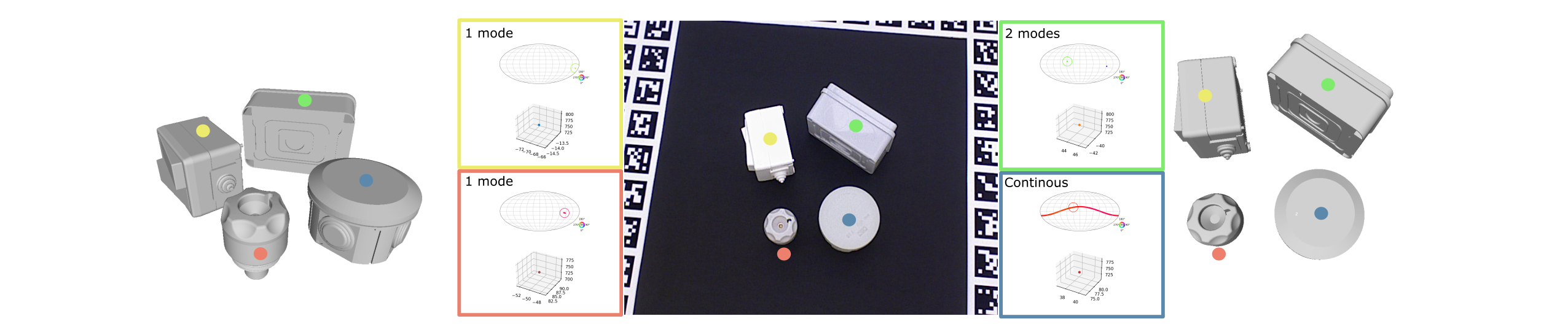}
\end{center}
  \caption{\textbf{Visualizations of our ground truth.} We display \SE(3) ground truth distributions for scene 1 of T-LESS~\cite{hodanTLESSRGBDDataset2017}. Circle on orientation diagram represents the unique ground truth pose provided as input to our method. Colors link objects to their distributions.
  }
  \label{fig:visu_tless_annotations}
\end{figure*}

Figure~\ref{fig:visu_tless_annotations} illustrates the coherence between the  distribution and the visibility of the disambiguating elements.
The ground truth translation is always reduced to a single 3D point as T-LESS exhibit no translational ambiguity.

Regarding quantitative evaluation of the symmetry pattern accuracy, Table~\ref{tab:quantitativeAccuracyTless} reports the max deviation error (in mm) between the depth image rendered from the GT pose $P_{GT}$ and the ones rendered from the combination of $P_{GT}$ with any pose of the symmetry pattern. We observe that the rendered depths are really close.

\begin{table}
 \begin{center}
 \resizebox{\columnwidth}{!}{
 \begin{tabular}{c c c c c}
 \toprule
$\mu_{\lvert \delta \rvert}$ (mm) & $\text{med}_{\lvert \delta \rvert}$ (mm) & max$_{\lvert \delta \rvert}$ (mm) & max$_{VSD}$  & max$_{Prune}$ (pix)\\
 \midrule
    0.218       &      5.85e-3  & 0.497 & 0.061 & 30 \\
  \bottomrule
 \end{tabular}
 }
 \end{center}
 \caption{\textbf{Quantitative accuracy analysis of our new pose distribution annotations on T-LESS.} We report the differences between rendered depths at the ground truth pose and all the points of the corresponding distribution. $\mu_{\lvert \delta \rvert}$, $\text{med}_{\lvert \delta \rvert}$ and max$_{\lvert \delta \rvert}$ are the mean, median and max of the absolute differences, while max$_{VSD}$ and max$_{Prune}$ are max errors of VSD@0.05 and of our pruning.}

 \label{tab:quantitativeAccuracyTless}
\end{table}

\begin{table*}
 \begin{center}
 \resizebox{0.78\textwidth}{!}{
 \begin{tabular}{l l c c c c c c c c l}
 \toprule
 &&\multicolumn{4}{c}{BOP \cite{sundermeyerBopChallenge20222023} (object-wise annotations)} & \multicolumn{5}{c}{BOP-Distrib (our image-wise annotations)} \\
 \cmidrule(lr){3-6}\cmidrule(lr){7-11}
 &BOP Contenders  &  $\mathbf{MSSD}$ & $\mathbf{MSPD}$ & $\mathbf{Mean}$ & Rank & $\mathbf{MSSD}$ & $\mathbf{MSPD}$ & $\mathbf{Mean}$ & $\mathbf{Loss}$ & Rank \\
 \hline
 \\
  \multicolumn{11}{c}{T-LESS~\cite{hodanTLESSRGBDDataset2017} \textbf{BOP Challenge 2023~\cite{hodan2024bop}: pose estimation of unseen objects}} \\
 \hline
& \cellcolor{blue!10}foundpose~\cite{ornek2023foundpose}& 55.0& 62.3 & 58.6 & \cellcolor{orange!61.388888888888886}1 &45.5& 52.1 & 48.8 &\cellcolor{red!61.70} -9.8&\cellcolor{orange!57.77777777777777}3({\color{purple} $\downarrow$-2})\\ 
\rowcolor{lightgray!40}\cellcolor{white!100}& \cellcolor{blue!10}gigaposemegapose-5-hypothesis~\cite{nguyen2024gigapose, labbe2022megapose}& 54.3& 62.0 & 58.1 & \cellcolor{orange!57.77777777777777}2 &45.3& 52.3 & 48.8 &\cellcolor{red!58.55} -9.3&\cellcolor{orange!61.388888888888886}2\\ 
& \cellcolor{blue!10}gigaposemegapose-1-hypothesis~\cite{nguyen2024gigapose, labbe2022megapose}& 51.5& 59.2 & 55.3 & \cellcolor{orange!54.16666666666667}3 &38.5& 45.1 & 41.8 &\cellcolor{red!85} -13.5&\cellcolor{orange!36.111111111111114}9({\color{purple} $\downarrow$-6})\\ 
\rowcolor{lightgray!40}\cellcolor{white!100}& \cellcolor{blue!10}genflow-multihypo16-rgb~\cite{moon2024genflow}& 50.9& 57.9 & 54.4 & \cellcolor{orange!50.55555555555556}4 &45.9& 52.4 & 49.1 &\cellcolor{red!33.37} -5.3&\cellcolor{orange!65.0}1({\color{teal} $\uparrow$3})\\ 
& \cellcolor{blue!10}genflow-multihypo-rgb~\cite{moon2024genflow}& 50.4& 57.3 & 53.9 & \cellcolor{orange!46.94444444444444}5 &44.9& 51.4 & 48.1 &\cellcolor{red!36.51} -5.8&\cellcolor{orange!54.16666666666667}4({\color{teal} $\uparrow$1})\\ 
\rowcolor{lightgray!40}\cellcolor{white!100}& \cellcolor{blue!10}genflow-multihypo16~\cite{moon2024genflow}& 52.8& 54.8 & 53.8 & \cellcolor{orange!43.333333333333336}6 &46.7& 48.6 & 47.7 &\cellcolor{red!38.40} -6.1&\cellcolor{orange!50.55555555555556}5({\color{teal} $\uparrow$1})\\ 
& \cellcolor{blue!10}genflow-multihypo~\cite{moon2024genflow}& 51.6& 53.6 & 52.6 & \cellcolor{orange!39.72222222222223}7 &45.2& 47.2 & 46.2 &\cellcolor{red!40.29} -6.4&\cellcolor{orange!46.94444444444444}6({\color{teal} $\uparrow$1})\\ 
\rowcolor{lightgray!40}\cellcolor{white!100}& \cellcolor{blue!10}foundpose~\cite{ornek2023foundpose}& 49.1& 55.6 & 52.4 & \cellcolor{orange!36.111111111111114}8 &37.7& 43.4 & 40.6 &\cellcolor{red!74.29} -11.8&\cellcolor{orange!28.888888888888886}11({\color{purple} $\downarrow$-3})\\ 
& \cellcolor{blue!10}foundpose~\cite{ornek2023foundpose}& 49.1& 55.6 & 52.3 & \cellcolor{orange!32.5}9 &37.7& 43.4 & 40.5 &\cellcolor{red!74.29} -11.8&\cellcolor{orange!25.27777777777778}12({\color{purple} $\downarrow$-3})\\ 
\rowcolor{lightgray!40}\cellcolor{white!100}& \cellcolor{blue!10}cnos-fastsammegapose-multihyp-10~\cite{nguyen2023cnos, labbe2022megapose}& 48.5& 55.9 & 52.2 & \cellcolor{orange!28.888888888888886}10 &40.6& 47.1 & 43.8 &\cellcolor{red!52.88} -8.4&\cellcolor{orange!43.33333333333333}7({\color{teal} $\uparrow$3})\\ 
& \cellcolor{blue!10}cnos-fastsammegapose-multihyp~\cite{nguyen2023cnos,labbe2022megapose}& 48.4& 55.6 & 52.0 & \cellcolor{orange!25.277777777777775}11 &40.5& 47.1 & 43.8 &\cellcolor{red!51.62} -8.2&\cellcolor{orange!39.72222222222223}8({\color{teal} $\uparrow$3})\\ 
\rowcolor{lightgray!40}\cellcolor{white!100}\multirow{-12}{*}{\cellcolor{white!100}\rotatebox[origin=c]{90}{RGB}}& \cellcolor{blue!10}cnos-fastsammegapose-multihyp-teaserpp~\cite{nguyen2023cnos, labbe2022megapose}& 48.4& 51.3 & 49.9 & \cellcolor{orange!21.666666666666668}12 &40.3& 43.0 & 41.7 &\cellcolor{red!51.62} -8.2&\cellcolor{orange!32.5}10({\color{teal} $\uparrow$2})\\ 
\hline
\rowcolor{lightgray!40}\cellcolor{white!100}& \cellcolor{white!100}foundationpose~\cite{wen2024foundationpose}& 62.9& 64.1 & 63.5 & \cellcolor{orange!59.090909090909086}1 &56.7& 58.0 & 57.4 &\cellcolor{red!37.57} -6.1&\cellcolor{orange!65.0}1\\ 
& \cellcolor{white!100}gigaposegenflowkabsch~\cite{nguyen2024gigapose}& 56.3& 58.6 & 57.5 & \cellcolor{orange!53.18181818181818}2 &49.4& 51.7 & 50.6 & \cellcolor{red!42.5}-6.9&\cellcolor{orange!59.090909090909086}2\\ 
\rowcolor{lightgray!40}\cellcolor{white!100}& \cellcolor{white!100}sam6d~\cite{lin2024sam}& 52.9& 54.5 & 53.7 & \cellcolor{orange!47.27272727272727}3 &39.7& 41.2 & 40.4 &\cellcolor{red!81.92} -13.3&\cellcolor{orange!53.18181818181819}3\\ 
& \cellcolor{white!100}sam6d~\cite{lin2024sam}& 52.1& 53.5 & 52.8 & \cellcolor{orange!41.36363636363636}4 &38.3& 39.7 & 39.0 & \cellcolor{red!85}-13.8&\cellcolor{orange!41.36363636363636}5({\color{purple} $\downarrow$-1})\\ 
\rowcolor{lightgray!40}\cellcolor{white!100}& \cellcolor{white!100}sam6d-fastsam~\cite{lin2024sam}& 49.9& 51.1 & 50.5 & \cellcolor{orange!35.45454545454545}5 &37.9& 39.1 & 38.5 & \cellcolor{red!73.91}-12.0&\cellcolor{orange!35.45454545454545}6({\color{purple} $\downarrow$-1})\\ 
& \cellcolor{white!100}sam6d-cnosmask~\cite{lin2024sam}& 50.0& 50.9 & 50.4 & \cellcolor{orange!29.545454545454547}6 &39.3& 40.3 & 39.8 & \cellcolor{red!65.28}-10.6&\cellcolor{orange!47.27272727272727}4({\color{teal} $\uparrow$2})\\ 
\rowcolor{lightgray!40}\cellcolor{white!100}& \cellcolor{white!100}sam6d-cnosfastsam~\cite{lin2024sam}& 49.3& 50.5 & 49.9 & \cellcolor{orange!23.636363636363637}7 &37.2& 38.4 & 37.8 &\cellcolor{red!74.52} -12.1&\cellcolor{orange!29.545454545454543}7\\ 
& \cellcolor{white!100}sam6d-zeropose~\cite{lin2024sam}& 44.4& 45.6 & 45.0 & \cellcolor{orange!17.727272727272727}8 &33.5& 34.8 & 34.2 & \cellcolor{red!66.52}-10.8&\cellcolor{orange!23.636363636363637}8\\ 
\rowcolor{lightgray!40}\cellcolor{white!100}\multirow{-9}{*}{\cellcolor{white!100}\rotatebox[origin=c]{90}{RGB-D}}& \cellcolor{white!100}zeropose-multi-hypo-refinement-defaultseg~\cite{chen2024zeropose}& 41.5& 42.0 & 41.8 & \cellcolor{orange!11.818181818181815}9 &29.4& 30.1 & 29.7 &\cellcolor{red!74.52} -12.1&\cellcolor{orange!17.727272727272727}9\\ 
\hline
 \\
  \multicolumn{10}{c}{T-LESS~\cite{hodanTLESSRGBDDataset2017} \textbf{BOP Challenge 2022~\cite{sundermeyerBopChallenge20222023}: pose estimation}} \\
 \hline
\rowcolor{lightgray!40}\cellcolor{white!100}& \cellcolor{blue!10}hccepose-2024-pbr& 82.9& 95.8 & 89.4 & \cellcolor{orange!60.0}1 &53.8& 65.2 & 59.5 &\cellcolor{red!85.0} -29.9&\cellcolor{orange!45.0}5({\color{purple} $\downarrow$-4})\\ 
& \cellcolor{blue!10}hcceposeefficientnet-b4-default-2d-bbox& 81.4& 94.8 & 88.1 & \cellcolor{orange!55.0}2 &53.1& 64.5 & 58.8 &\cellcolor{red!83.29} -29.3&\cellcolor{orange!40.0}6({\color{purple} $\downarrow$-4})\\ 
\rowcolor{lightgray!40}\cellcolor{white!100}& \cellcolor{blue!10}cosypose-eccv20-syntreal-8views~\cite{labbeCosyPoseConsistentMultiview2020}& 83.6& 90.7 & 87.2 & \cellcolor{orange!49.99999999999999}3 &57.2& 64.0 & 60.6 &\cellcolor{red!75.61} -26.6&\cellcolor{orange!50.0}4({\color{purple} $\downarrow$-1})\\ 
& \cellcolor{blue!10}hccepose-default-2d-bbox& 79.2& 93.2 & 86.2 & \cellcolor{orange!45.0}4 &51.4& 63.1 & 57.2 & \cellcolor{red!82.44}-29.0&\cellcolor{orange!30.0}8({\color{purple} $\downarrow$-4})\\ 
\rowcolor{lightgray!40}\cellcolor{white!100}& \cellcolor{blue!10}gdrnppdet-pbrrealgenflow-multihypo-rgb~\cite{wangGdrnetGeometryguidedDirect2021, gdrnpp2022}& 78.4& 92.2 & 85.3 & \cellcolor{orange!40.0}5 &69.9& 82.9 & 76.4 &\cellcolor{red!25.30} -8.9&\cellcolor{orange!60.0}2({\color{teal} $\uparrow$3})\\ 
& \cellcolor{blue!10}leroy-fuseocclu-rgb& 78.0& 91.7 & 84.8 & \cellcolor{orange!35.0}6 &70.4& 83.3 & 76.9 &\cellcolor{red!22.45} -7.9&\cellcolor{orange!65.0}1({\color{teal} $\uparrow$5})\\ 
\rowcolor{lightgray!40}\cellcolor{white!100}& \cellcolor{blue!10}gpose2023-rgb~\cite{gpose2023}& 76.6& 92.9 & 84.7 & \cellcolor{orange!30.0}7 &50.2& 64.0 & 57.1 &\cellcolor{red!78.46} -27.6&\cellcolor{orange!25.0}9({\color{purple} $\downarrow$-2})\\ 
& \cellcolor{blue!10}gdrnpp-pbr-rgb-mmodel~\cite{wangGdrnetGeometryguidedDirect2021, gdrnpp2022}& 76.3& 92.4 & 84.4 & \cellcolor{orange!24.999999999999996}8 &50.0& 63.6 & 56.8 &\cellcolor{red!78.46} -27.6&\cellcolor{orange!20.0}10({\color{purple} $\downarrow$-2})\\ 
\rowcolor{lightgray!40}\cellcolor{white!100}& \cellcolor{blue!10}gdrnpp-pbrreal-rgb-mmodel~\cite{wangGdrnetGeometryguidedDirect2021, gdrnpp2022}& 76.0& 91.3 & 83.6 & \cellcolor{orange!20.0}9 &48.8& 61.7 & 55.2 &\cellcolor{red!80.73} -28.4&\cellcolor{orange!5.0}13({\color{purple} $\downarrow$-4})\\ 
& \cellcolor{blue!10}cosypose-eccv20-syntreal-4views~\cite{labbeCosyPoseConsistentMultiview2020}& 79.5& 86.4 & 83.0 & \cellcolor{orange!14.999999999999996}10 &52.7& 59.3 & 56.0 &\cellcolor{red!76.75} -27.0&\cellcolor{orange!10.0}12({\color{purple} $\downarrow$-2})\\ 
\rowcolor{lightgray!40}\cellcolor{white!100}& \cellcolor{blue!10}mrc-net~\cite{liMrcnet2024}& 78.5& 87.1 & 82.8 & \cellcolor{orange!10.0}11 &52.3& 60.3 & 56.3 &\cellcolor{red!75.33} -26.5&\cellcolor{orange!15.0}11\\ 
& \cellcolor{blue!10}gdrnppdet-pbrrealmegapose-multihyp~\cite{wangGdrnetGeometryguidedDirect2021, gdrnpp2022}& 74.0& 87.4 & 80.7 & \cellcolor{orange!4.9999999999999964}12 &61.6& 73.8 & 67.7 &\cellcolor{red!36.95} -13.0&\cellcolor{orange!55.0}3({\color{teal} $\uparrow$9})\\ 
\rowcolor{lightgray!40}\cellcolor{white!100}\multirow{-13}{*}{\cellcolor{white!100}\rotatebox[origin=c]{90}{RGB}}& \cellcolor{blue!10}sc6d~\cite{cai2022sc6d}& 72.1& 85.3 & 78.7 & \cellcolor{orange!0.0}13 &52.6& 64.1 & 58.3 &\cellcolor{red!57.99} -20.4&\cellcolor{orange!35.0}7({\color{teal} $\uparrow$6})\\ 
\hline

\rowcolor{lightgray!40}\cellcolor{white!100}& \cellcolor{white!100}gpose2023~\cite{gpose2023}& 92.1& 94.6 & 93.4 & \cellcolor{orange!61.9047619047619}1 &86.5& 89.1 & 87.8 & \cellcolor{red!15.81}-5.6&\cellcolor{orange!65.0}1\\ 
& \cellcolor{white!100}gdrnppv2-rgbd-pbrreal~\cite{wangGdrnetGeometryguidedDirect2021, gdrnpp2022}& 90.2& 92.6 & 91.4 & \cellcolor{orange!58.80952380952381}2 &84.8& 87.2 & 86.0 &\cellcolor{red!15.24} -5.4&\cellcolor{orange!58.80952380952381}3({\color{purple} $\downarrow$-1})\\ 
\rowcolor{lightgray!40}\cellcolor{white!100}& \cellcolor{white!100}gpose2023-pbr~\cite{gpose2023}& 90.3& 92.6 & 91.4 & \cellcolor{orange!55.714285714285715}3 &84.7& 87.2 & 86.0 & \cellcolor{red!15.24}-5.4&\cellcolor{orange!61.9047619047619}2({\color{teal} $\uparrow$1})\\ 
& \cellcolor{white!100}modalocclusion-rgbd& 89.7& 92.9 & 91.3 & \cellcolor{orange!52.61904761904762}4 &83.2& 86.4 & 84.8 & \cellcolor{red!18.35}-6.5&\cellcolor{orange!55.71428571428571}4\\ 
\rowcolor{lightgray!40}\cellcolor{white!100}& \cellcolor{white!100}hcceposebf-2024-ref& 89.0& 91.7 & 90.3 & \cellcolor{orange!49.52380952380952}5 &71.7& 74.7 & 73.2 &\cellcolor{red!48.28} -17.1&\cellcolor{orange!49.52380952380952}6({\color{purple} $\downarrow$-1})\\ 
& \cellcolor{white!100}gdrbpp-pbrreal-rgbd-mmodel-fast \textbf{v1.4}~\cite{wangGdrnetGeometryguidedDirect2021, gdrnpp2022}& 88.7& 91.5 & 90.1 & \cellcolor{orange!46.42857142857143}6 &58.1& 61.9 & 60.0 &\cellcolor{red!85} -30.1&\cellcolor{orange!27.857142857142854}13({\color{purple} $\downarrow$-7})\\ 
\rowcolor{lightgray!40}\cellcolor{white!100}& \cellcolor{white!100}gdrnpp-pbrreal-rgbd-mmodel \textbf{v1.3}~\cite{wangGdrnetGeometryguidedDirect2021, gdrnpp2022}& 88.4& 90.9 & 89.6 & \cellcolor{orange!43.333333333333336}7 &60.6& 63.7 & 62.2 & \cellcolor{red!77.37}-27.4&\cellcolor{orange!37.14285714285714}10({\color{purple} $\downarrow$-3})\\ 
& \cellcolor{white!100}gdrnpp-pbrreal-rgbd-smodel \textbf{v1.2}~\cite{wangGdrnetGeometryguidedDirect2021, gdrnpp2022}& 87.1& 90.3 & 88.7 & \cellcolor{orange!40.23809523809524}8 &57.5& 61.2 & 59.3 & \cellcolor{red!83.02}-29.4&\cellcolor{orange!21.666666666666664}15({\color{purple} $\downarrow$-7})\\ 
\rowcolor{lightgray!40}\cellcolor{white!100}& \cellcolor{white!100}zebraposesat-effnetb4-refineddefdet-2023~\cite{su2022zebrapose}& 87.3& 90.2 & 88.7 & \cellcolor{orange!37.14285714285714}9 &60.2& 63.4 & 61.8 & \cellcolor{red!75.96}-26.9&\cellcolor{orange!34.04761904761905}11({\color{purple} $\downarrow$-2})\\ 
& \cellcolor{white!100}gdrnpp-pbr-rgbd-mmodel \textbf{v1.1}~\cite{wangGdrnetGeometryguidedDirect2021, gdrnpp2022}& 87.2& 90.1 & 88.7 & \cellcolor{orange!34.04761904761905}10 &60.8& 64.2 & 62.5 & \cellcolor{red!73.98}-26.2&\cellcolor{orange!40.23809523809524}9({\color{teal} $\uparrow$1})\\ 
\rowcolor{lightgray!40}\cellcolor{white!100}& \cellcolor{white!100}hipose-cvpr24~\cite{lin2024hipose}& 86.6& 88.8 & 87.7 & \cellcolor{orange!30.95238095238095}11 &58.0& 60.8 & 59.4 & \cellcolor{red!79.91}-28.3&\cellcolor{orange!24.76190476190476}14({\color{purple} $\downarrow$-3})\\ 
& \cellcolor{white!100}defaultdetection-pfa-mixpbr-rgb-d& 85.5& 87.7 & 86.6 & \cellcolor{orange!27.857142857142858}12 &59.2& 62.0 & 60.6 & \cellcolor{red!73.42}-26.0&\cellcolor{orange!30.95238095238095}12\\ 
\rowcolor{lightgray!40}\cellcolor{white!100}& \cellcolor{white!100}zebraposesat-effnetb4defdet-2023~\cite{su2022zebrapose}& 79.1& 93.0 & 86.1 & \cellcolor{orange!24.76190476190476}13 &51.8& 63.4 & 57.6 & \cellcolor{red!80.48}-28.5&\cellcolor{orange!15.476190476190474}17({\color{purple} $\downarrow$-4})\\ 
& \cellcolor{white!100}zebraposesat-effnetb4pbr-only-defdet-2023& 77.7& 92.0 & 84.9 & \cellcolor{orange!21.666666666666668}14 &50.7& 62.8 & 56.8 & \cellcolor{red!79.35}-28.1&\cellcolor{orange!12.38095238095238}18({\color{purple} $\downarrow$-4})\\ 
\rowcolor{lightgray!40}\cellcolor{white!100}& \cellcolor{white!100}gdpnpp-pbrreal-rgbd-mmodel-officialdet \textbf{v1.0}~\cite{wangGdrnetGeometryguidedDirect2021, gdrnpp2022}& 83.4& 85.6 & 84.5 & \cellcolor{orange!18.57142857142857}15 &57.5& 60.2 & 58.9 & \cellcolor{red!72.29}-25.6&\cellcolor{orange!18.57142857142857}16({\color{purple} $\downarrow$-1})\\ 
& \cellcolor{white!100}surfemb-pbr-rgbd-lin~\cite{haugaardSurfEmbDenseContinuous2022}& 82.9& 85.9 & 84.4 & \cellcolor{orange!15.476190476190478}16 &75.8& 78.7 & 77.2 & \cellcolor{red!20.33}-7.2&\cellcolor{orange!52.61904761904762}5({\color{teal} $\uparrow$11})\\ 
\rowcolor{lightgray!40}\cellcolor{white!100}\multirow{-17}{*}{\cellcolor{white!100}\rotatebox[origin=c]{90}{RGB-D}}& \cellcolor{white!100}poseio& 81.7& 86.0 & 83.8 & \cellcolor{orange!12.38095238095238}17 &69.3& 73.6 & 71.4 & \cellcolor{red!35.01}-12.4&\cellcolor{orange!43.33333333333333}8({\color{teal} $\uparrow$9})\\ 
 \hline
 \\
 \multicolumn{10}{c}{YCB-V~\cite{xiangPoseCNNConvolutionalNeural2018} \textbf{BOP Challenge 2022~\cite{sundermeyerBopChallenge20222023}: pose estimation}} \\
 \hline

&\cellcolor{blue!10}cosypose-eccv20-syntreal-8views~\cite{labbeCosyPoseConsistentMultiview2020} &88.5& 88.0 & 88.2 & \cellcolor{orange!48.75}1 &83.3& 82.4 & 82.9 & \cellcolor{red!85}-5.3&\cellcolor{orange!32.5}2({\color{purple} $\downarrow$-1})\\ 
\rowcolor{lightgray!40}\cellcolor{white!100}&\cellcolor{blue!10}cosypose-eccv20-syntreal-4views~\cite{labbeCosyPoseConsistentMultiview2020} &86.9& 86.6 & 86.8 & \cellcolor{orange!32.5}2 &82.1& 81.3 & 81.7 & \cellcolor{red!81.79}-5.1&\cellcolor{orange!0}4({\color{purple} $\downarrow$-2})\\ 
&\cellcolor{blue!10}mrpe-pbrreal-rgb-smodel &85.7& 87.3 & 86.5 & \cellcolor{orange!16.25}3 &83.1& 84.4 & 83.7 & \cellcolor{red!44.90}-2.8&\cellcolor{orange!48.75}1({\color{teal} $\uparrow$2})\\ 
\rowcolor{lightgray!40} \multirow{-4}{*}{\cellcolor{white!100}\rotatebox[origin=c]{90}{RGB}}&\cellcolor{blue!10}magic &85.9& 85.4 & 85.6 & \cellcolor{orange!0.0}4 &82.9& 82.3 & 82.6 & \cellcolor{red!48.11}-3.0&\cellcolor{orange!16.5}3({\color{teal} $\uparrow$1})\\ 
\hline
 
& \cellcolor{white!100}gdrnppv2-rgbd-pbrreal~\cite{wangGdrnetGeometryguidedDirect2021, gdrnpp2022}&96.3& 92.7 & 94.5 & \cellcolor{orange!48.75}1 &94.2& 90.2 & 92.2 & \cellcolor{red!34.29}-2.3&\cellcolor{orange!48.75}1\\ 
\rowcolor{lightgray!40}\cellcolor{white!100}&\cellcolor{white!100}gpose2023~\cite{gpose2023} &96.2& 92.6 & 94.4 & \cellcolor{orange!32.5}2 &94.2& 90.2 & 92.2 & \cellcolor{red!32.80}-2.2&\cellcolor{orange!32.5}2\\ 
&\cellcolor{white!100}gdrnpp-pbrreal-rgbd-mmodel~\cite{wangGdrnetGeometryguidedDirect2021, gdrnpp2022} &95.7& 91.9 & 93.8 & \cellcolor{orange!16.25}3 &90.1& 86.2 & 88.1 & \cellcolor{red!85}-5.7&\cellcolor{orange!16.25}3\\ 
\rowcolor{lightgray!40}\multirow{-4}{*}{\cellcolor{white!100}\rotatebox[origin=c]{90}{RGB-D}}&\cellcolor{white!100}dsgc6d &93.0& 90.1 & 91.5 & \cellcolor{orange!0.0}4 &87.5& 85.2 & 86.4 & \cellcolor{red!76.05}-5.1&\cellcolor{orange!0}4\\
 \hline
 \\
 \multicolumn{10}{c}{T-LESS~\cite{hodanTLESSRGBDDataset2017} \textbf{Best mode of pose distribution methods}} \\
 \hline
 \rowcolor{lightgray!40}\multirow{2}{*}{\cellcolor{white!100}\rotatebox[origin=c]{90}{RGB}}
&\cellcolor{blue!10}LiePose Diffusion~\cite{hsiao2024confronting} & 60.1 & 92.2 & 76.1 & \cellcolor{orange!65.0} 1  & 48.3 & 76.1 & 62.2 &\cellcolor{red!85} -13.9 & \cellcolor{orange!30.0}2({\color{purple} $\downarrow$-1})\\ 
&\cellcolor{blue!10}SpyroPose~\cite{haugaard2023spyropose}  & 61.2 & 75.3 & 68.3  & \cellcolor{orange!30.0} 2 & 57.8 & 71.6 & 64.7 &\cellcolor{red!22.01} -3.6 & \cellcolor{orange!65.0}1({\color{teal} $\uparrow$1})\\
 \bottomrule
 \end{tabular}
 } 
 \end{center}
 \caption{
 \textbf{Results of BOP 2023 on T-LESS~\cite{hodanTLESSRGBDDataset2017} and YCB-V~\cite{xiangPoseCNNConvolutionalNeural2018} with our annotations.} We report the top contenders results of the \textbf{pose estimation of unseen objects challenge} and the \textbf{pose estimation challenge} evaluated with $\mathbf{MSSD}$ and $\mathbf{MSPD}$ on both the official ground truth (object-wise) and our ground truth (image-wise). 
 We rank the methods with the mean of $\mathbf{MSSD}$ and $\mathbf{MSPD}$ (similar to BOP, except we exclude $\mathbf{VSD}$ as BOP starts to abandon it) and re-rank them with the mean of $\mathbf{MSSD}$ and $\mathbf{MSPD}$ computed on our ground truth distributions. The drop in the scores highlighted by the $\mathbf{Loss}$ column ($\mathbf{Loss} = \mathbf{Mean}_{\mathbf{BOP-Distrib}} - \mathbf{Mean}_{\mathbf{BOP}}$) produces drastic changes in the rankings. We also show the same scores for the \textbf{best mode of distribution estimation methods}, however, both of them use ground truth bounding boxes instead of a detector, so direct comparison would be unfair.
 This highlights the importance to consider image level symmetries as we proposed, to evaluate accurately pose estimation methods. We also processed YCB-V. The $\mathbf{Loss}$ is much lower as there are fewer symmetrical objects. Names of the methods are automatically processed from BOP csv result files.
 }
 \label{tab:vtless}
 \end{table*}

\subsection{Single Pose Estimation Evaluation}
\label{subsec:expSinglepose}

Since May 2023, raw results of BOP Challenge~\cite{hodan2024bop} submissions have become publicly available to allow in-depth analysis. This experiment reprocess these results to compare the performance evaluations of state-of-the-art methods for both the original 6D annotations and our annotations. We considered top contenders of the BOP challenge for the pose estimation  and pose estimation for unseen objects tasks. We also included two methods for the multi-modal pose distribution estimation task~\cite{haugaard2023spyropose,hsiao2024confronting} for which we kept only the highest confidence mode.

\paragraph{Experimental protocol.}
We evaluate the current (July 2024) top contenders of the BOP Challenge on the T-LESS dataset~\cite{hodanTLESSRGBDDataset2017} on the \textbf{pose estimation} task and the \textbf{pose estimation of unseen object} task together with \cite{haugaard2023spyropose,hsiao2024confronting}. We conduct a similar evaluation with YCB-V~\cite{xiangPoseCNNConvolutionalNeural2018}.

This evaluation is based directly on the raw pose results provided by the contenders and the official evaluation scripts of BOP challenges, both being publicly available on the Challenge website. 
Our method ranking is based on the mean of Recalls computed on MSPD and MSSD accuracies.

\paragraph{Results.}
Table~\ref{tab:vtless} reports the results. 
We can observe a large impact of the ground truth on the methods results (see \textbf{Loss} column) and ranking.  
In terms of ranking, the well established \textbf{pose estimation} task faces the biggest changes, with gdrnppdet-pbrrealmegapose-multihyp moving from the 12th to the 3rd position for RGB methods and surfemb-pbr-rgbd-lin~\cite{haugaardSurfEmbDenseContinuous2022} moving from the 16th to the 5th place for the RGB-D methods.
For \textbf{pose estimation of unseen objects}, the difference of ranking can reach 6 places. 

We observe large impacts on the recall performances as well. Whereas 29 of the 30 contenders of \textbf{pose estimation} task have a mean recall up to 80\% and even up to 90\% for 6 methods with the original annotations, only 4 methods exceed this score with our annotations, with a majority of methods having less than 65\%. Similarly, most of the contenders for the \textbf{pose estimation of unseen objects} task have a mean recall that exceeds 50\% with the original annotations whereas only the 2 best RGB-D methods exceed this score once evaluated with our annotations. As shown in Figure~\ref{fig:teaser} (\colorbox{red!60}{Case 3}), the image-wise annotation rejects poses when the image is not ambiguous. The Loss is explained by inaccurate pose estimates that were mistakenly validated by previous object-wise annotations (see supplementary materials for visualizations).

\subsection{Pose Distribution Evaluation}
\label{subsec:expDistpose}

We provide now the first evaluation of 6D pose distribution estimation methods on a dataset of real images.

\paragraph{Experimental protocol.}
Similarly to Section~\ref{subsec:expSinglepose}, the methods were trained on the train set of T-LESS. The ground truth used for this dataset is the same than the one of Section~\ref{subsec:expSinglepose} and is constituted of a discrete set of 6D pose obtained with the 6D pose annotation process introduced in Section~\ref{sec:method}. 
The performances of the methods are evaluated in Precision and Recall, using accuracy measures MPD and MSD as introduced in Section~\ref{sec:distributionComparison}.
More details on methods and results processing are given as supplementary material.
\paragraph{Results.}
We evaluated SpyroPose~\cite{haugaard2023spyropose} and LiePose-Diffusion~\cite{hsiao2024confronting} (either PBR + real images and PBR only to compare with SpyroPose) since they provide 6D pose distributions, unlike \cite{murphy2021implicit,hofer2023hyperposepdf} that provide distributions only over the rotation. To show genericity, we also evaluated the top 3 RGB methods of single \textbf{pose estimation} task (with image-wise annotations). Their distribution is considered as a Dirac distribution on the single estimate. Results are given on the complete dataset, and the subset of objects with cylindrical symmetry (1-4, 13-18, 24 and 30).

Quantitative results are reported on Table~\ref{tab:tlessDistribution} whereas qualitative results are provided in the supplementary material.
It appears that LiePose-Diffusion outperforms SpyroPose, except for the Precision MSD. Lower accuracy are obtained with MSD error since those methods use RGB images only and no depth. Single pose methods outperform distribution methods in terms of Precision, as they are optimized for that. Yet, distribution methods produce much stronger Recalls, even more on only symmetrical objects, highlighting their ability to retrieve multiple meaningfull poses from the targeted image-wise distribution. 

\begin{table}
 \begin{center}
 \resizebox{\columnwidth}{!}{
 \begin{tabular}{l l c c c c c }
 \toprule
 &Methods    & Train & $\mathbf{P_{MSD}}$ & $\mathbf{R_{MSD}}$ & $\mathbf{P_{MPD}}$ & $\mathbf{R_{MPD}}$ \\
 \midrule

 \multicolumn{7}{c}{T-LESS~\cite{hodanTLESSRGBDDataset2017} \textbf{Complete dataset}} \\
 \hline
\multirow{3}{*}{\cellcolor{white!100}\rotatebox[origin=c]{90}{Distrib}}&\cellcolor{blue!10}SpyroPose~\cite{haugaard2023spyropose} & \cellcolor{lightgray!40}PBR & \cellcolor{lightgray!40}\textbf{32.8} & \cellcolor{lightgray!40}48.1 & \cellcolor{lightgray!40}55.9 & \cellcolor{lightgray!40}55.5\\
&\cellcolor{blue!10}LiePose-Diffusion~\cite{hsiao2024confronting} & PBR  & 24.6 & \textbf{71.4}& \textbf{61.2} & \textbf{89.5}\\
&\cellcolor{blue!10}LiePose-Diffusion~\cite{hsiao2024confronting} & \cellcolor{lightgray!40}PBR + real  & \cellcolor{lightgray!40}29.9 & \cellcolor{lightgray!40}75.4& \cellcolor{lightgray!40}68.4 & \cellcolor{lightgray!40}90.6\\
\hline
\rowcolor{lightgray!40}\cellcolor{white!100}&\cellcolor{white!100} leroy-fuseocclu-rgb & PBR + real & \textbf{55.4} & \textbf{35.7} & \textbf{80.9} & \textbf{54.2} \\
&\cellcolor{white!100}gdrnppdet-pbrrealgenflow-multihypo-rgb~\cite{wangGdrnetGeometryguidedDirect2021, gdrnpp2022} & PBR + real &  52.4 & 34.9 & 77.7 & 52.2 \\
\rowcolor{lightgray!40}\multirow{-3}{*}{\cellcolor{white!100}\rotatebox[origin=c]{90}{Single}}&\cellcolor{white!100}cosypose-eccv20-syntreal-8views~\cite{labbeCosyPoseConsistentMultiview2020} & \cellcolor{lightgray!40}PBR + real & \cellcolor{lightgray!40}43.5 & \cellcolor{lightgray!40}25.7 & \cellcolor{lightgray!40}53.1 & \cellcolor{lightgray!40}30.9 \\
\hline
 \\
 \multicolumn{7}{c}{T-LESS~\cite{hodanTLESSRGBDDataset2017} \textbf{Only symmetrical objects}} \\
 \hline
 \multirow{3}{*}{\cellcolor{white!100}\rotatebox[origin=c]{90}{Distrib}}&\cellcolor{blue!10}SpyroPose~\cite{haugaard2023spyropose} & \cellcolor{lightgray!40}PBR & \cellcolor{lightgray!40}\textbf{31.5} & \cellcolor{lightgray!40}42.7 & \cellcolor{lightgray!40}70.5 & \cellcolor{lightgray!40}60.8\\
&\cellcolor{blue!10}LiePose-Diffusion~\cite{hsiao2024confronting} & PBR & 18.3 & \textbf{61.6} & \textbf{63.6} & \textbf{87.3}\\
&\cellcolor{blue!10}LiePose-Diffusion~\cite{hsiao2024confronting} & \cellcolor{lightgray!40}PBR + real & \cellcolor{lightgray!40}24.1 & \cellcolor{lightgray!40}67.0 & \cellcolor{lightgray!40}70.3 & \cellcolor{lightgray!40}88.1\\
\hline
\rowcolor{lightgray!40}\cellcolor{white!100}&\cellcolor{white!100}leroy-fuseocclu-rgb & \cellcolor{lightgray!40}PBR + real &\cellcolor{lightgray!40} \textbf{52.4} & \cellcolor{lightgray!40}\textbf{25.7} & \cellcolor{lightgray!40}\textbf{86.1} & \cellcolor{lightgray!40}\textbf{45.3}\\
&\cellcolor{white!100}gdrnppdet-pbrrealgenflow-multihypo-rgb~\cite{wangGdrnetGeometryguidedDirect2021, gdrnpp2022} & PBR + real & 45.8 & 20.8 & 76.5 & 37.8 \\
\rowcolor{lightgray!40}\multirow{-3}{*}{\cellcolor{white!100}\rotatebox[origin=c]{90}{Single}}&\cellcolor{white!100}cosypose-eccv20-syntreal-8views~\cite{labbeCosyPoseConsistentMultiview2020} & \cellcolor{lightgray!40}PBR + real & \cellcolor{lightgray!40}29.1 & \cellcolor{lightgray!40}6.5 & \cellcolor{lightgray!40}40.0 & \cellcolor{lightgray!40}10.5 \\

 \bottomrule
 \end{tabular}
 } 
 \end{center}
 \caption{
 \textbf{Comparison of pose distribution estimation methods on T-LESS using our ground truth pose distributions.}
 }
 \label{tab:tlessDistribution}
 \end{table}

\section{Limitations}
Our annotation method relies on geometric analysis. Sensor resolution, sensor noise, field of view or motion blur may also affect disambiguating parts visibility. Considering these effects would improve the ground truth even further.

\section{Conclusion}
For simplicity, most of the 6D pose estimation benchmarks rely on a single 6D pose annotation per image, completed by a per-object symmetry pattern to transform this unique pose into a distribution. We argued that ignoring the per-image nature of the symmetry pattern is prone to bias in the resulting ground truth and  performances evaluations. 

We then proposed a method to annotate 6D pose distribution with a per-image analysis of the object symmetries. 
We illustrated that the resulting ground truth is more accurate. Moreover, when using this ground truth to re-evaluate current state-of-the-art methods, we showed that the ranking of these methods changes drastically. 

We also introduced metrics to evaluate methods that estimate a pose distribution and provided their first evaluation on real data. Such pose distribution qualification is crucial for downstream tasks. Indeed, access to multiple accurate pose solutions allows the selection of the best one for the task (e.g., considering obstacles) or, when the object is known to be asymmetrical, helps determine if the task can’t be completed from the current viewpoint due to occlusion of the elements that break the symmetry and select the next best view.

\section*{Acknowledgment}
This work was partly funded by the European Union's Horizon Europe Research and Innovation Program under Grant 101070227 (CONVINCE) and partly funded by the European Union’s Horizon Europe Research and Innovation program under grant agreement nº 101135708 (JARVIS Project).
The authors would like to thank Rasmus Laurvig Haugaard for providing SpyroPose~\cite{haugaard2023spyropose} implementation for baseline evaluation.
{
    \small
    \bibliographystyle{ieeenat_fullname}
    \bibliography{cleaned_main}
}
\clearpage
\setcounter{page}{1}
\setcounter{section}{0}

\maketitlesupplementary

\section{\boris{Pseudo-code of Per-image Ground Truth Pose Annotation}}
\boris{In this section, we give the key elements of our annotation method with the following pseudo-codes.}
\label{sec:supp_pseudo}
\small
\begin{lstlisting}[language=Python, caption={Offline $\epsSym$ pre-computation with all vertices visible, pseudo-code of \autoref{eq:epssymPrecompute} from Section~\ref{subsec:accelerate}.}]
# Input: CAD model, candidate transformation set, threshold
# Output: per point epsilon sym set
def EpsSym(ModelCAD, TransformSet, %$\epsilon$%, %$\zeta$%, resolution):
  SampledModel = uniform_resampling(ModelCAD, resolution)
  for point in SampledModel:
    for T in TransformSet:
      # Texture-less case
      if testColor == False:
        if knnSearch(T*point, SampledModel, %$\epsilon$%) == True:
          EpsSym[point].append(T)
      # Textured case
      else:
        if knnSearch(T*point, SampledModel, %$\epsilon$%) == True:
          if %$d_{Color}$%(point, T*point) < %$\zeta$%:
            EpsSym[point].append(T)
  
  return EpsSym, SampledModel
\end{lstlisting}

The geometric ($d_{Geom}$) and colorimetric ($d_{Color}$) distances are implemented as follows:\\
$d_{Geom}(x, m) = ||x-m||_2$ and\\
\begin{multline*}
d_{Color}(x, m) < \zeta \Leftrightarrow \min \Big(|h(x)-h(m)|,\\
|h(x)-h(m)-360|,|h(x)-h(m)+360|\Big) < \zeta_h,\\
\text{AND~} |s(x)-s(m)| < \zeta_s,\\
\text{AND~} |v(x)-v(m)| < \zeta_v.   
\end{multline*}
with $h(.), s(.), v(.)$ being hue, saturation and value of the point, and $\zeta_h, \zeta_s, \zeta_v$, the respective hue, saturation and value thresholds.

\begin{lstlisting}[language=Python, caption={Image annotation, pseudo-code of \autoref{ref:eqHist} from Section~\ref{subsec:robustIntersect}.}]
# Input: SampledModel, EpsSym, PoseGT, MaskVisib, %\color{codegreen}{$\tau$}%
# Output: EpsSymImage
def SoftIntersection(SampledModel, EpsSym, PoseGT, MaskVisib, %$\tau$%):
  # Count the vertices that vote for a transform
  for point in SampledModel:
    if K*[%$R_{gt}$%, %$T_{gt}$%]*point in mask_visib:
      for T in EpsSym[point]:
        H[T]++
  Sort(H)
  # Count the vertices that vote for a transform
  for i in size(H):
    if H[0]-H[i] < %$\tau$%:
      EpsSymImage.append(i)
     
  return EpsSymImage
\end{lstlisting}
\begin{lstlisting}[language=Python, caption={Image annotation depth post-processing, pseudo-code of \autoref{ref:eqHist} from Section~\ref{subsec:refine}, based on VSD implementation from BOP toolkit.}]
# Input: ModelCAD, EpsSymImage, PoseGT, SensorDepth, %$\delta$%
# Output: EpsSymImageGlobal
def PostProcessAnnotateImage(ModelCAD, EpsSymImage, PoseGT, SensorDepth, %$\delta$%, threshold):
  depthGT = render(ModelCAD, PoseGT)
  distGT = depthToDistImage(depthGT)
  visibleMaskGT = generateMask(depthGT, SensorDepth)
  for Ti in EpsSymImage:
    depthEst = render(ModelCAD, Ti)
    distEst = depthToDistImage(depthEst)
    visibleMaskEst = generateMask(depthEst, SensorDepth)
    maskIntersection = intersection(visibleMaskGT, visibleMaskEst)
    maskUnion = union(visibleMaskGT, visibleMaskEst)
    nbPixelsOutlier += maskUnion-maskIntersection
    nbPixelsOutlier += abs(distGt - distEst)[maskIntersection] >= %$\delta$%
    if nbPixelsOutlier < threshold:
      EpsSymImageGlobal.append(Ti)
      
  return EpsSymImageGlobal
\end{lstlisting}
\normalsize

Section~\ref{subsec:refine} does not use the VSD metric, as VSD is normalized by the number of pixels in the union of the visible masks. In our case, we need an absolute score and not a relative one, so that the counting of outliers has a metric meaning and can represent the size of a minimal disambiguating element.

\section{Analysis of our BOP-Distrib Annotations}

\subsection{\boris{False visible points and false occluded points detected by the pruning stage}}
\label{suppmat:discussionLimitEps}

\begin{figure}
  \includegraphics[width=0.45\textwidth]{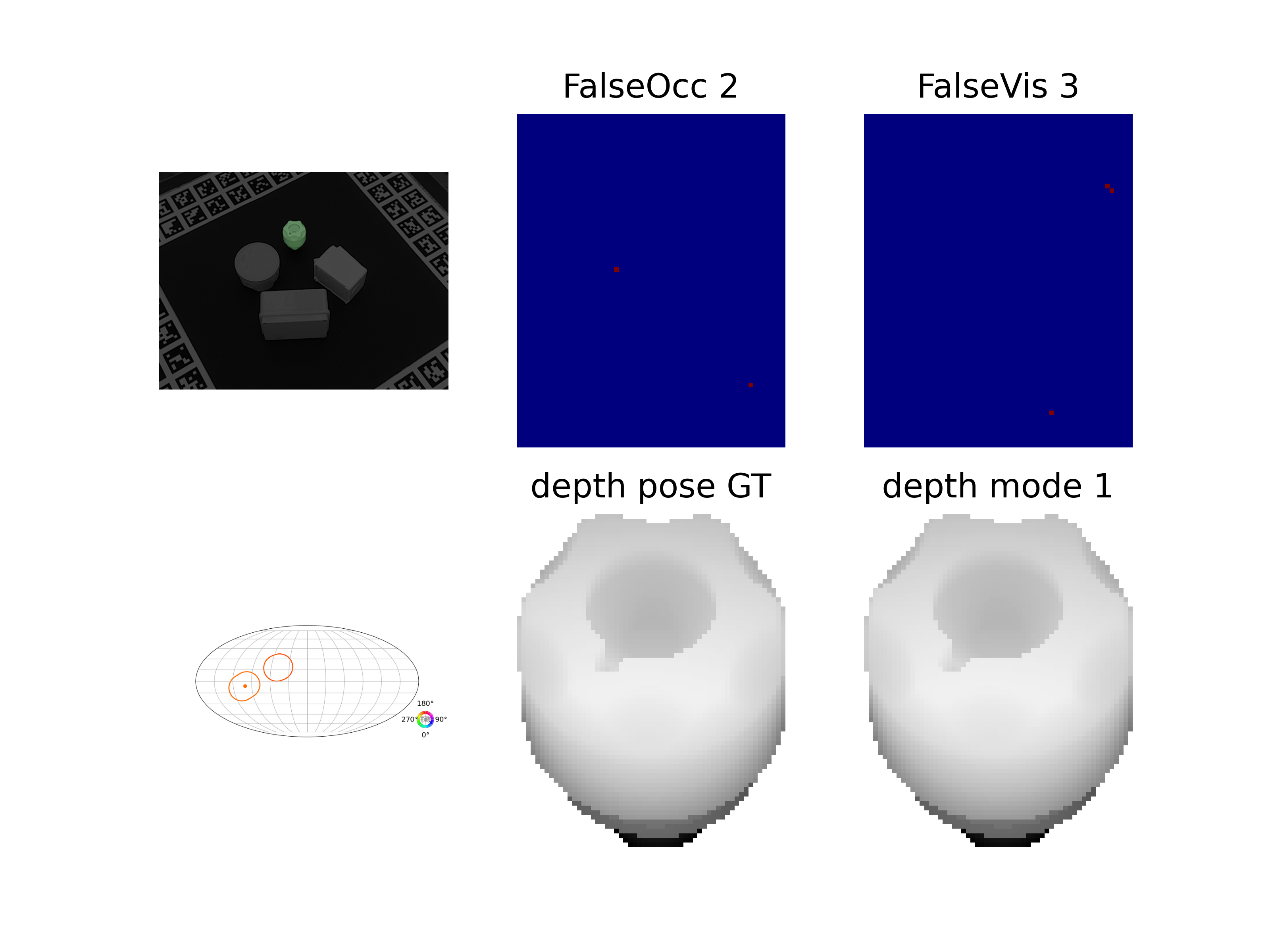}
  \caption{\textbf{Depth deviation post-processing analysis.} For a given image, we display the depth renderings of the ground truth pose and of one $\epsSym$ mode (1 here). They align well.}
  \label{fig:pruningNo}
\end{figure}

\begin{figure}
  \includegraphics[width=0.45\textwidth]{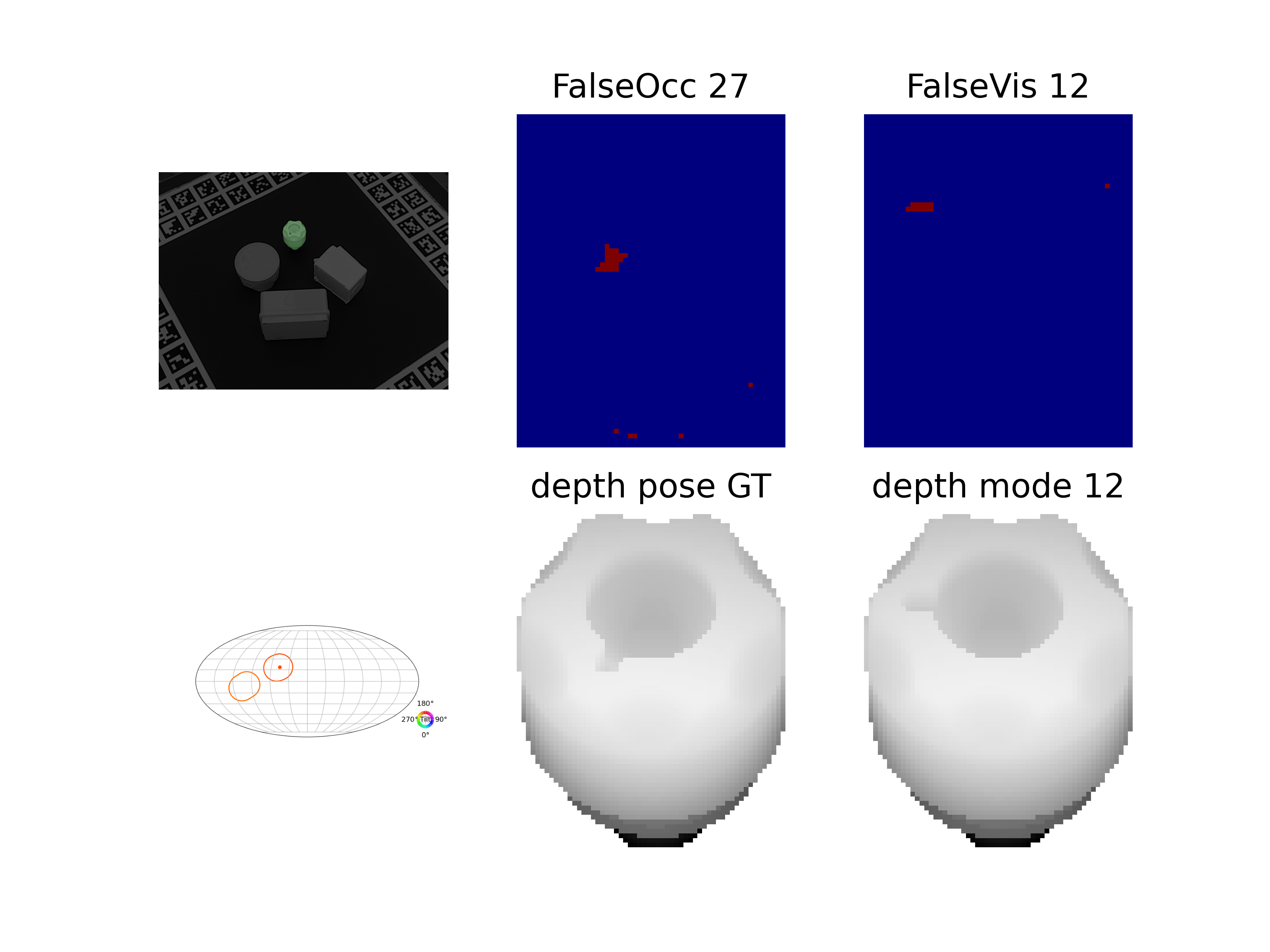}
  \caption{\textbf{Depth deviation post-processing analysis.} For a given image, we display the depth renderings of the ground truth pose and of one $\epsSym$ mode (12). Mode 12 generates several falsely occluded pixels (where the hole should be) and falsely visible pixels (where the hole is but shouldn't be). Mode 12 is rejected by or pruning stage.}
  \label{fig:pruningYes}
\end{figure}

Figures~\ref{fig:pruningNo} and \ref{fig:pruningYes} present the results of our pruning procedure. Each $\epsSym$ mode is rendered to be compared to the ground truth pose depth rendering. The pixel with deviation greater than $\delta$ are counted, and if they too numerous (more than $\tau_{pix}$), the mode is pruned, as in figure~\ref{fig:pruningYes}.

\subsection{Choice of Distribution Representation}
\label{suppmat:discussionRepresentation}
Unlike approaches based on Bingham distributions~\cite{deng2022deep, bui20206d, gilitschenski2019deep}, implicit representations \cite{murphy2021implicit, hofer2023hyperposepdf}, Wigner harmonics~\cite{lee20243d}, matrix Fisher distributions~\cite{yin2022fishermatch} or continuous symmetry groups \cite{bregierDefiningPoseAny2017}, we do not have a continuous distribution representation. We represent the ground truth pose distributions in a non-parametric way with a set samples of it.
Representing a distribution with a large set of samples is common practice and has the advantage of being general: it can represent distributions with no clear analytical form which happen in case of visual ambiguities beyond symmetries, \ie, it can represent distributions with no clear analytical form which happen in case of visual ambiguities beyond symmetries. Moreover, a set of samples permits an efficient performance evaluation.

\subsection{Additionnal BOP-Distrib Ground Truths Visualizations}

We first provide more visualizations for qualitative appreciation of the new ground truth annotations accuracy in Figure~\ref{fig:bop-d_gt}.

These images are taken from a video compilation of all ground truth annotations sorted by object identifier, also provided as supplementary material (\url{BOP\_Distrib\_id8513\_supp\_newGT\_visualizations.mp4}), to convince the reader of their quality. We invite the reader to stop on some frames and check that the distribution recovered by our method does correspond to the ambiguities in the image for the object in the bounding box.

\subsection{T-LESS~\cite{hodanTLESSRGBDDataset2017} Annotation Details}
In our experiments, we sample surface points from the CAD models with a resolution of 0.5mm. For the pre-computation of elementary symmetries patterns for a given object, we use the per-object symmetries pattern given by the BOP challenge~\cite{hodan2024bop} as the initial symmetry candidates, with a tolerance factor $\epsilon\text{-sym}$ set to 1mm. 

The object surface visibility $\calV(M)$ is computed by Z-buffering using ground truth pose $P_\text{GT}$ and the 3D model of the object. 
For the robust symmetry pattern intersection, the soft intersection tolerance factor $\tau$ was experimentally adjusted to 28 3D points, resulting in a minimal disambiguating element size of roughly 2.5$\times$2.5mm$^2$, $\delta=5mm$ and $\tau_{pix}=30pix$.

\subsection{YCB-V~\cite{xiangPoseCNNConvolutionalNeural2018} Annotation Details}\label{subsec:ycbv}
In our experiments, we sample surface points from the CAD models with a resolution of 1mm. For the pre-computation of elementary symmetries patterns for a given object, we use the per-object symmetries pattern given by the BOP challenge~\cite{hodan2024bop} as the initial symmetry candidates, with a tolerance factor $\epsilon\text{-sym}$ set to 2mm. The color tolerance is set in the HSV color space to have the chrominance on a single channel (hue) and only luminance on the other two (saturation and value). It is empirically set to 4$^{\circ}$ in hue and 0.1 in saturation and value.

The object surface visibility $\calV(M)$ is computed by Z-buffering using ground truth pose $P_\text{GT}$ and the 3D model of the object. 
For the robust symmetry pattern intersection, the soft intersection tolerance factor $\tau$ was experimentally adjusted to 28 3D points, resulting in a minimal disambiguating element size of roughly 5.3$\times$5.3mm$^2$. The pruning stage was not necessary for YCB-V, as objects are simpler, with much less occlusions.

\subsection{Differences between Original Object-wise BOP Annotations and Our Image-wise Annotations}
\begin{figure*}[h!]
\begin{tabular}{c}
  \includegraphics[width=0.95\textwidth]{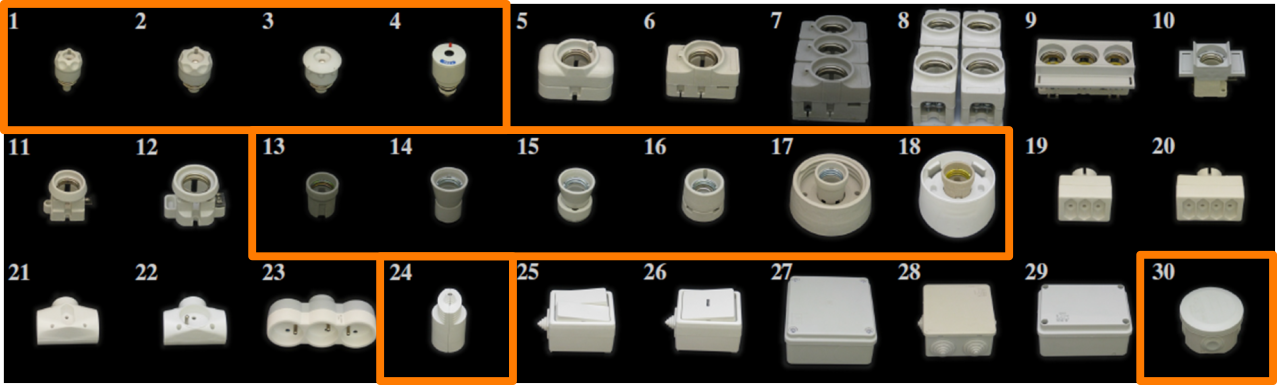} \\
  \includegraphics[width=0.95\textwidth]{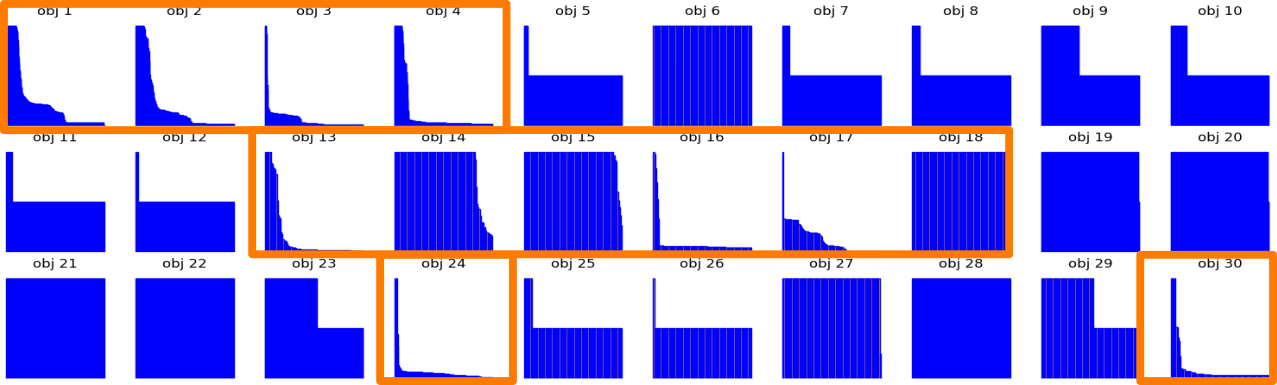}
\end{tabular}
  \caption{
  \textbf{Visualization of annotation changes compared to the T-LESS original annotations.}
  \textbf{Top:} T-LESS objects with their identifiers. \textbf{Bottom:} For each object, we plot the percentages of poses kept from the original annotations by our method over the images~(sorted by percentages). T-LESS assumes full rotational symmetries, while our annotation method captures more complex symmetry patterns. Only Object 18 is perfectly symmetrical and our method retrieves the same poses as the original annotations. For the other objects, in particular the objects with complex symmetry patterns like the first 4 objects, our annotations significantly change the original annotations. These changes in the annotations explain the score changes for T-LESS in Table~\ref{tab:vtless}. The objects annotated as 'circular' in BOP are highlighted in \colorbox{orange!60}{orange}.
  }
  \label{fig:tless_syms}
\end{figure*}

\begin{figure*}[h!]
\centering
\begin{tabular}{c}
  \includegraphics[width=0.7\textwidth]{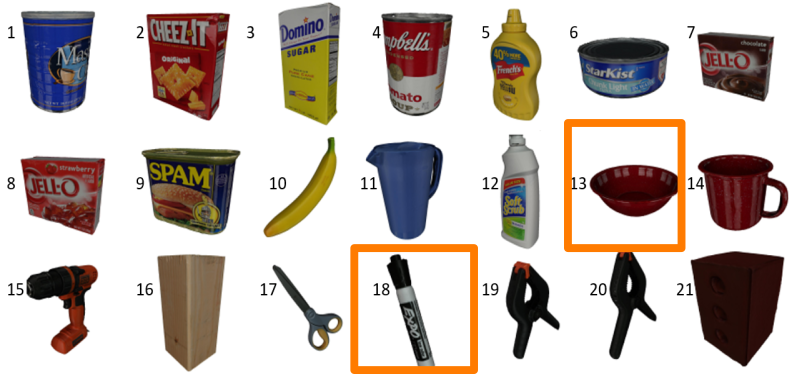} \\
  \includegraphics[width=0.7\textwidth]{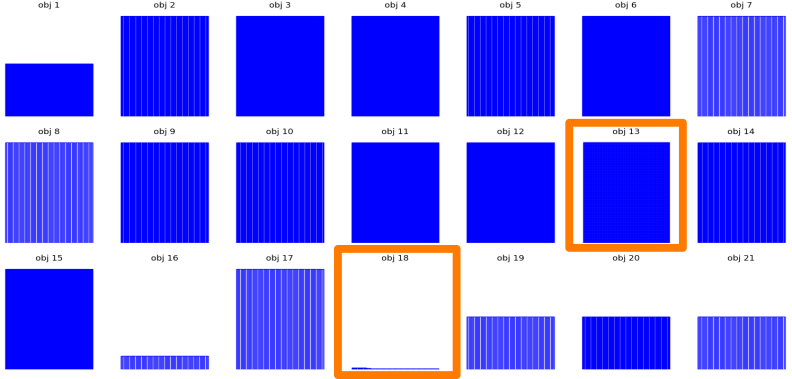}
\end{tabular}
  \caption{
  \textbf{Visualization of annotation changes compared to the YCB-V original annotations.}
  \textbf{Top:} YCB-V objects with their identifiers. \textbf{Bottom:} For each object, we plot the percentages of poses kept from the original annotations by our method over the images~(sorted by percentages). YCB-V assumes full rotational symmetries, while our annotation method captures more complex symmetry patterns. Only Object 13 is perfectly symmetrical, both in terms of geometry and texture, and our method retrieves the same poses as the original annotations. Object 18 is given as completely symmetrical but our method tags it as non-ambiguous due to its texture. Similarly, objects 1, 16, 19, 20 and 21 have few symmetrical poses for BOP annotations where our method keep always only one pose, as the texture disambiguate them. These changes in the annotations explain the score changes for YCB-V in Table~\ref{tab:vtless}. The objects annotated as 'circular' in BOP are highlighted in \colorbox{orange!60}{orange}.
  }
  \label{fig:ycbv_syms}
\end{figure*}

Figure~\ref{fig:tless_syms} for T-LESS~\cite{hodanTLESSRGBDDataset2017} and Figure~\ref{fig:ycbv_syms} for YCB-V~\cite{xiangPoseCNNConvolutionalNeural2018} highlight the differences between the poses that are accepted by BOP and the ones accepted when using our annotations. For this purpose, for each object, we provide a bar plot. This bar plot shows, for each image where the object appears, the percentage of poses considered correct by BOP that are also considered correct with our annotations. The bar plots show the percentages after sorting them, i.e., the bar on the left corresponds to the image with the smallest difference.

Concerning T-LESS, the annotations for some objects remain mostly unchanged. But because our method analyses more finely the possible object symmetries, many poses are actually not accepted with our annotations for the other objects (mainly the 'circular' ones). T-LESS is a very good dataset for our per-image pose distribution annotation method as it features objects with complex symmetries as well as a lot inter-object occlusions.

\begin{figure*}[h!]
\centering
\includegraphics[width=0.9\textwidth]{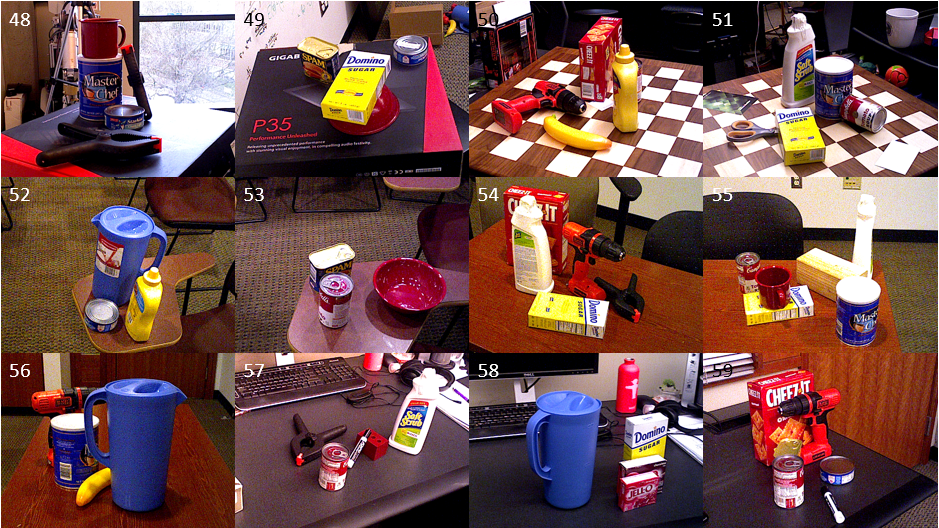}
\caption{\textbf{Visualization of YCB-V scenes.} The scenes present at most 6 different objects, with one instance of each. The objects with potential ambiguities (mainly objects 14 and 18) have limited occlusions. YCB-V is not really suited for pose distribution evaluation as no object becomes ambiguous, apart from object 13.}
\label{fig:ycbv_scenes}
\end{figure*}

Concerning YCB-V, some objects such as the mug (object 14 on \autoref{fig:ycbv_syms}) could yield very interesting symmetries with occlusions~\cite{manhardt2019explaining}. However, the disambiguating handle of the mug is never occluded. Hence the BOP symmetries that tag the mug as unambiguous. Overall, YCB-V has less potential for displaying visual ambiguities. Most objects are disambiguated by texture and the few ambiguous YCB-V objects do not face sufficient occlusions to become ambiguous (objects 11, 14 and 18 mainly), as depicted by the visualization of the scenes in Figure~\ref{fig:ycbv_scenes}. We show here that our per-image annotation method is able to retrieve finer symmetries patterns, which have an effect when evaluating Single Pose Estimation methods as in Table~\ref{tab:vtless}.


\begin{figure*}[h!]
\centering
\includegraphics[width=0.6\textwidth]{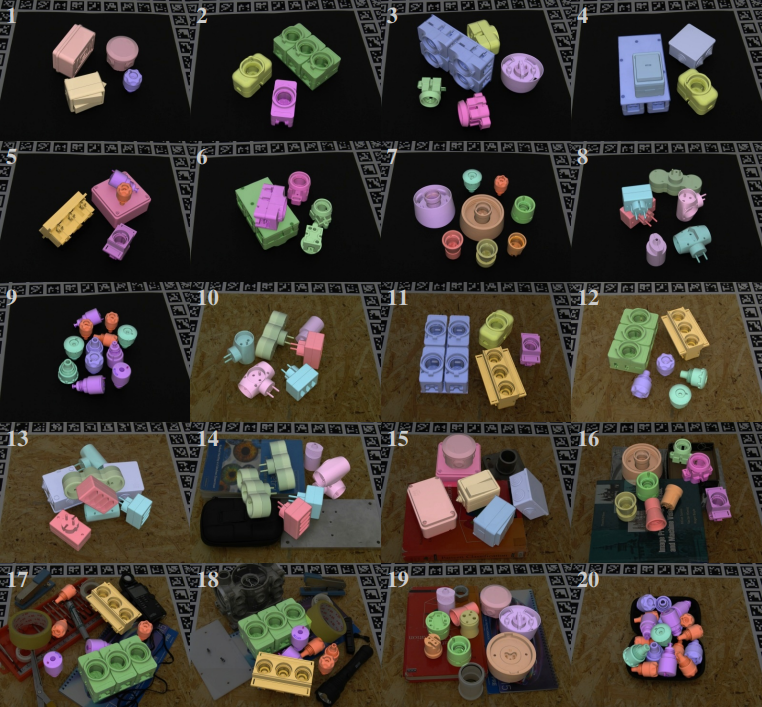}
\caption{\textbf{Visualization of T-LESS scenes.} The scenes present up to 18 instances of different objects, with a lot of occlusions and with ambiguous objects. T-LESS is really suited for pose distribution evaluation. (illustration borrowed from~\cite{hodanTLESSRGBDDataset2017})}
\label{fig:tless_scenes}
\end{figure*}

\begin{figure*}
  \centering

  \includegraphics[width=0.65\textwidth]{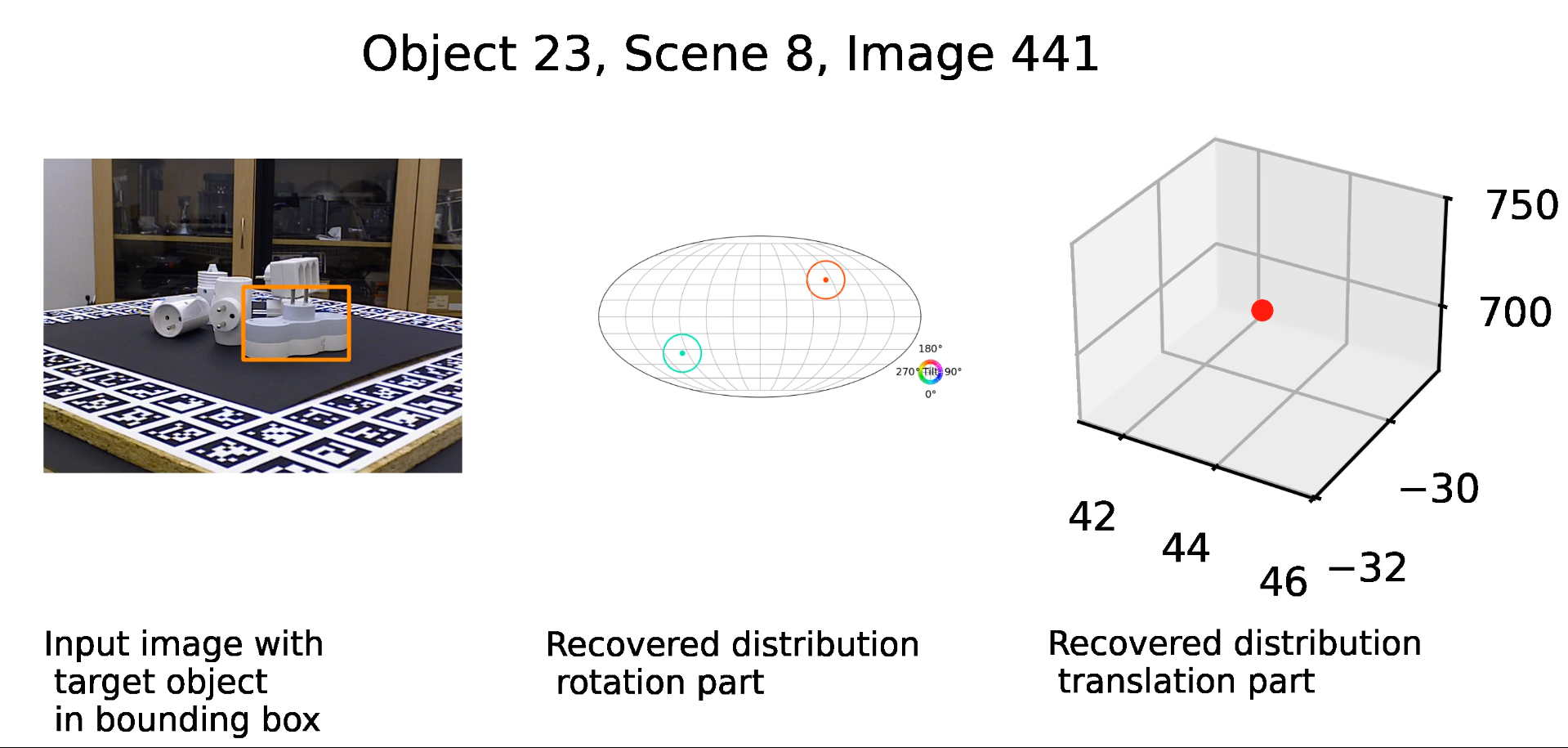}
  
  \includegraphics[width=0.65\textwidth]{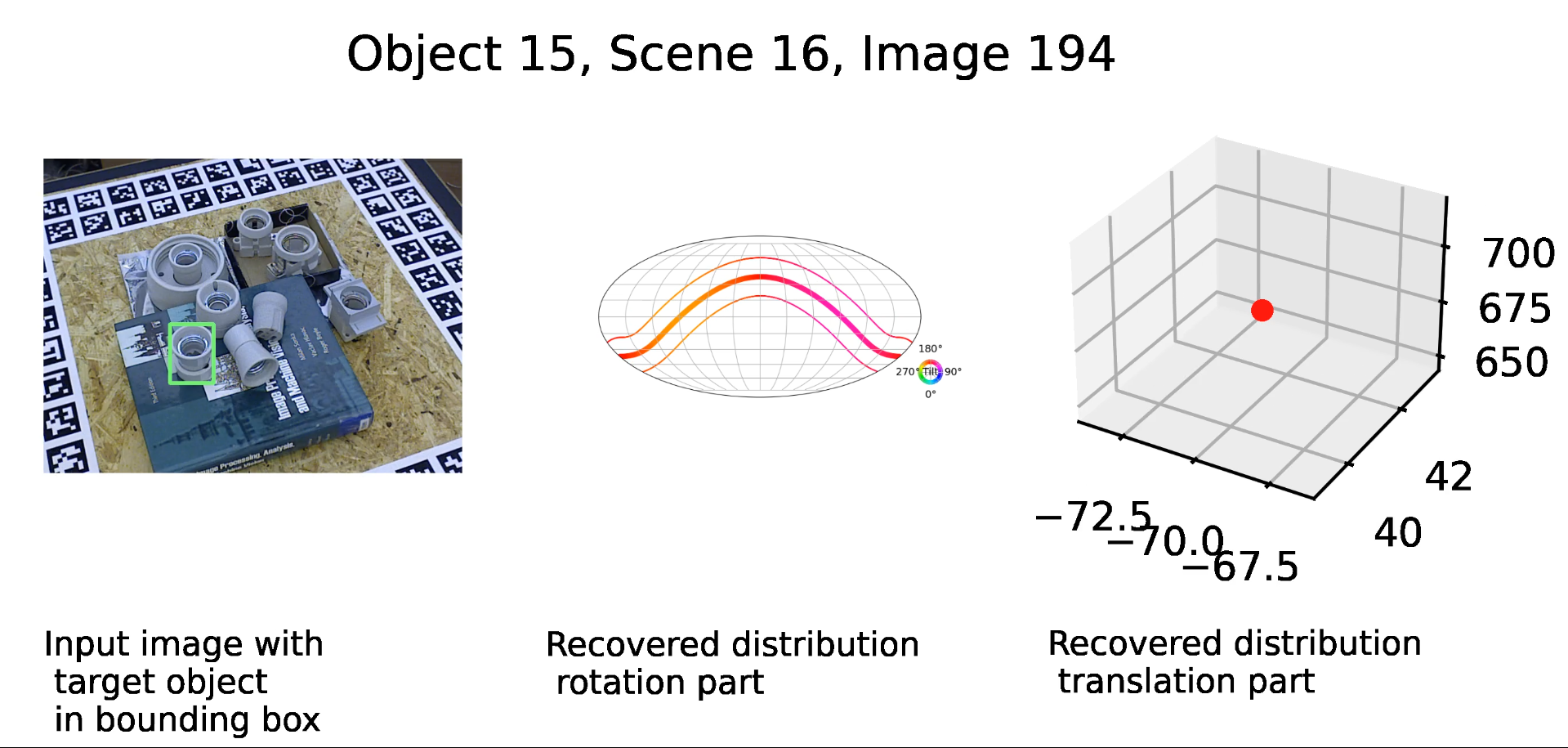}
  
  \includegraphics[width=0.65\textwidth]{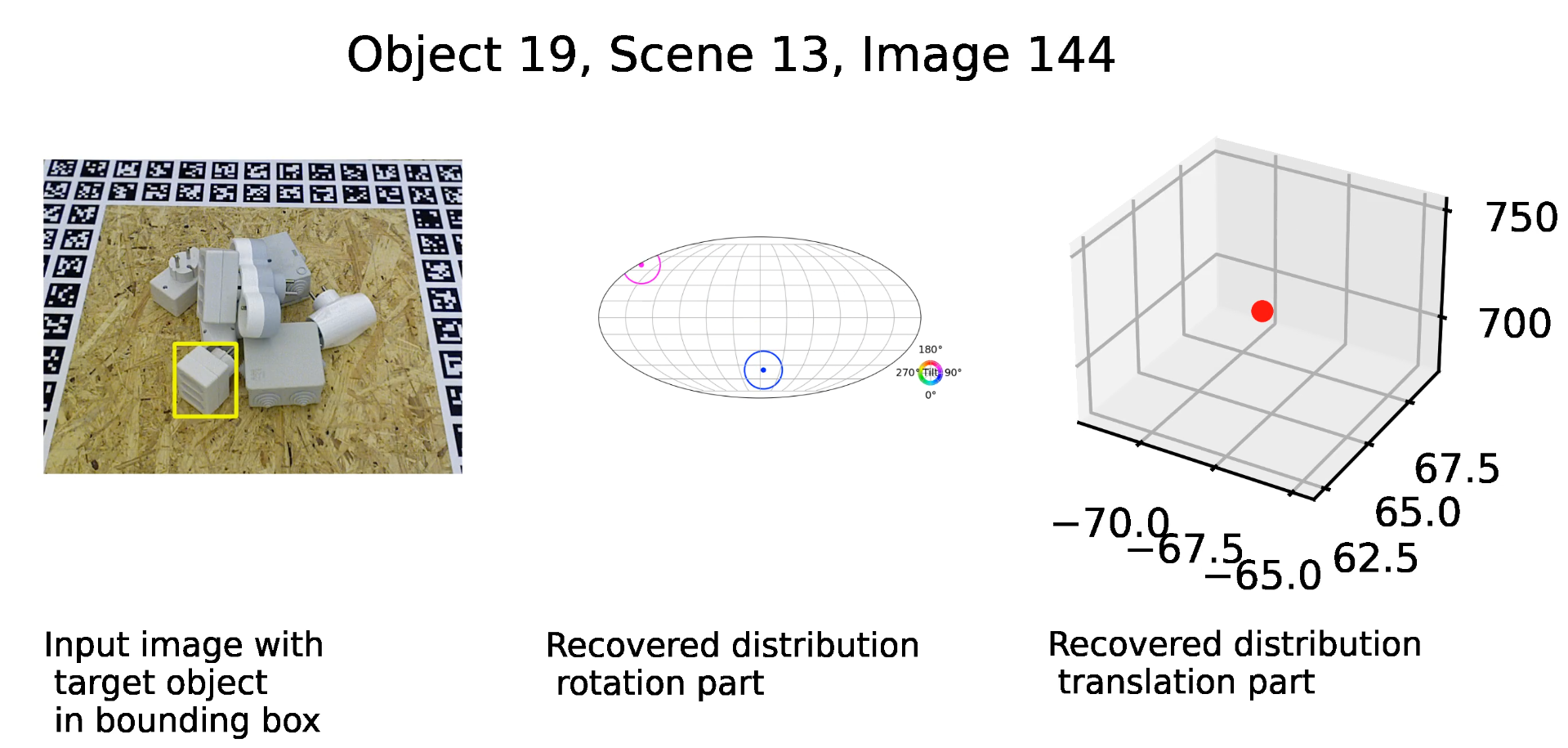}
  
  \includegraphics[width=0.65\textwidth]{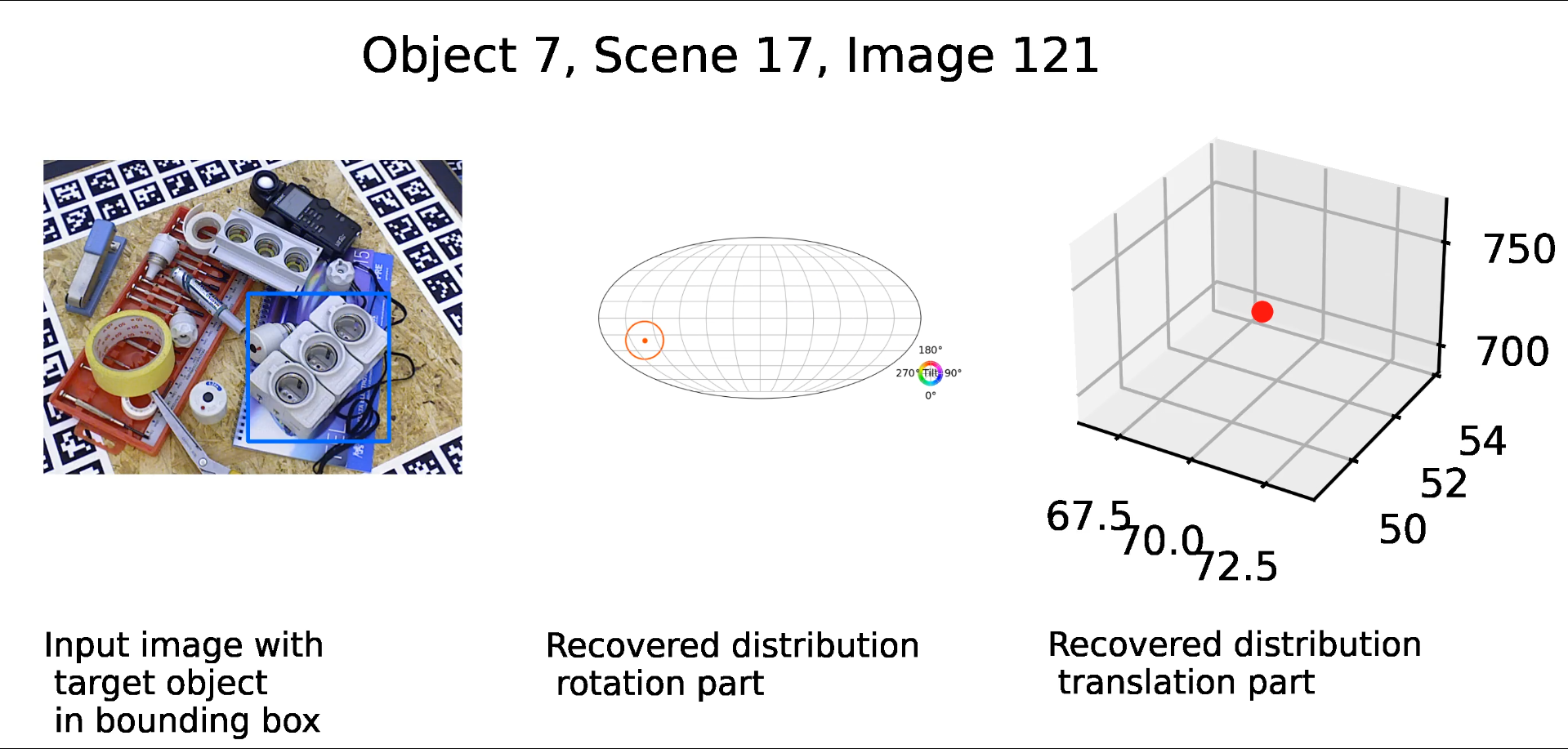}
  \caption{\textbf{Visualizing some BOP-Distrib ground truths as computed by our method.} Each example features an object of interest, in the bounding box in the left image, and its BOP-Distrib pose distribution, split between the rotation part (center) and translation part (right). As no object present symmetries in translation here, the translation part of the distribution is always on the same point of $\mathbb{R}^{3}$. We provide much more examples in the accompanying video.}
  \label{fig:bop-d_gt}
\end{figure*}


  
  
  
  

\section{Computing the Pose Estimation Results}
\subsection{T-LESS Pose Estimation and Unseen Objects Pose Estimation}

Recently, the BOP challenge~\cite{sundermeyerBopChallenge20222023} made public the pose estimation results of the methods evaluated in the leaderboard~\footnote{\url{https://bop.felk.cvut.cz/leaderboards/}}. Based on these results and the BOP toolkit~\footnote{\url{https://github.com/thodan/bop_toolkit/tree/master}} in which we implemented the variations of \textbf{MSSD} and \textbf{MSPD}, that use our per-image symmetries patterns instead of BOP global object symmetries, we were able to reprocess these pose estimates against our new ground truths. Our results on T-LESS have been presented in Table~\ref{tab:vtless}. Section~\ref{subsec:singleSupp} of supplementary material illustrates some of the failure cases with the new and more accurate ground truth.


\subsection{YCB-V Pose Estimation}
We conducted a similar experience of reprocessing BOP competitors on our re-annotation of the YCB-V.
Our results have been presented in Table~\ref{tab:vtless}.

\subsection{Computation of the SpyroPose~\cite{haugaard2023spyropose} Distribution Results}
\label{sec:supp_spyro}
Beyond finer evaluation of Single Pose Estimation methods, our new per-image annotations allow us to propose the first evaluation on real data of Pose Distribution Estimation methods.
As no pose distribution evaluation existed, the authors of~\cite{haugaard2023spyropose} reported only per-object averaged log-likelihood against BOP original ground truth.

The implementation provided by the authors of SpyroPose allows to train a network for one object of T-LESS. We batched the training stage for all objects, and then batched the inference. For one object, SpyroPose produces more than 100 000 estimates, sorted by their probabilities, as it samples SO(3) $\times$ $\mathbb{R}^{3}$ (using a slice of $\mathbb{R}^{3}$). As the probabilities of these estimates quickly tend to zero, we reduce their distributions to the 400 best estimates. These 400 estimates are used for the evaluations of Section~\ref{subsec:expDistpose} of the article.

\subsection{Computation of the Lie-Pose Diffusion~\cite{hsiao2024confronting} Distribution Results}
\label{sec:supp_liepose}

Similarly, as no pose distribution evaluation existed, the authors of~\cite{hsiao2024confronting} reported only few qualitative illustrations of pose distribution results on T-LESS. They evaluated their method as a Single Pose Estimation method, with a single run of the method.

The implementation provided by the authors of Lie-Pose allows to train a network for all objects of T-LESS. The inference phase produces one pose estimate per image crop. We batched the inference phase, with varying seeds. We then merged all these estimates into a single set of poses per image crop. The authors report 1000 runs to produce a distribution. For our evaluation, due to important computation time, we ran the code 100 times with different input noises to produce Lie-Pose distribution results.



\section{Illustrations of Single Pose Re-evaluation}
\label{subsec:singleSupp}

We provide here more cues about 
the changes in the Single Pose ranking, based on the  \textbf{MSSD} and \textbf{MSPD} metrics using our new more accurate ground truth.

To do so, we look at the pose estimates of the method gdrnpp-pbrreal-rgbd-mmodel\textbf{v1.3}. When we evaluated gdrnpp-pbrreal-rgbd-mmodel\textbf{v1.3} estimates against our new ground truths, it changed the method ranking from rank 7 to rank 10 in Table~\ref{tab:vtless}. Even more interestingly, the metrics went down from an \textbf{MSSD} of 88.4 and an \textbf{MSPD} of 90.9 to an \textbf{MSSD} of 60.6 and an \textbf{MSPD} of 63.7.

\begin{figure*}
\centering
\begin{tabular}{p{0.48\textwidth}p{0.48\textwidth}}
\includegraphics[trim={2.75cm 4.5cm 8cm 5cm},clip, height=1.2in]{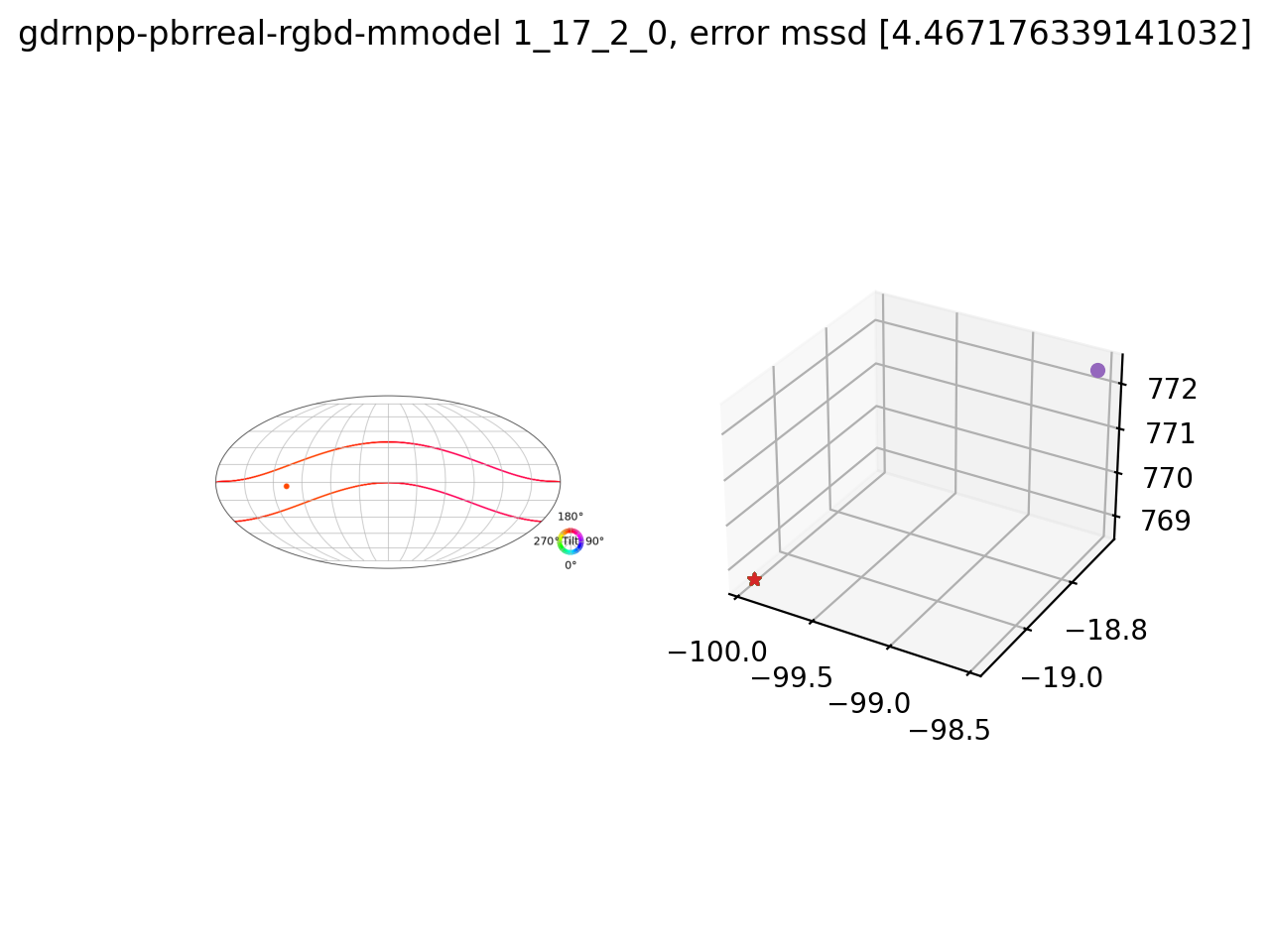}
&
\includegraphics[trim={2.75cm 4.5cm 8cm 5cm},clip,height=1.2in]{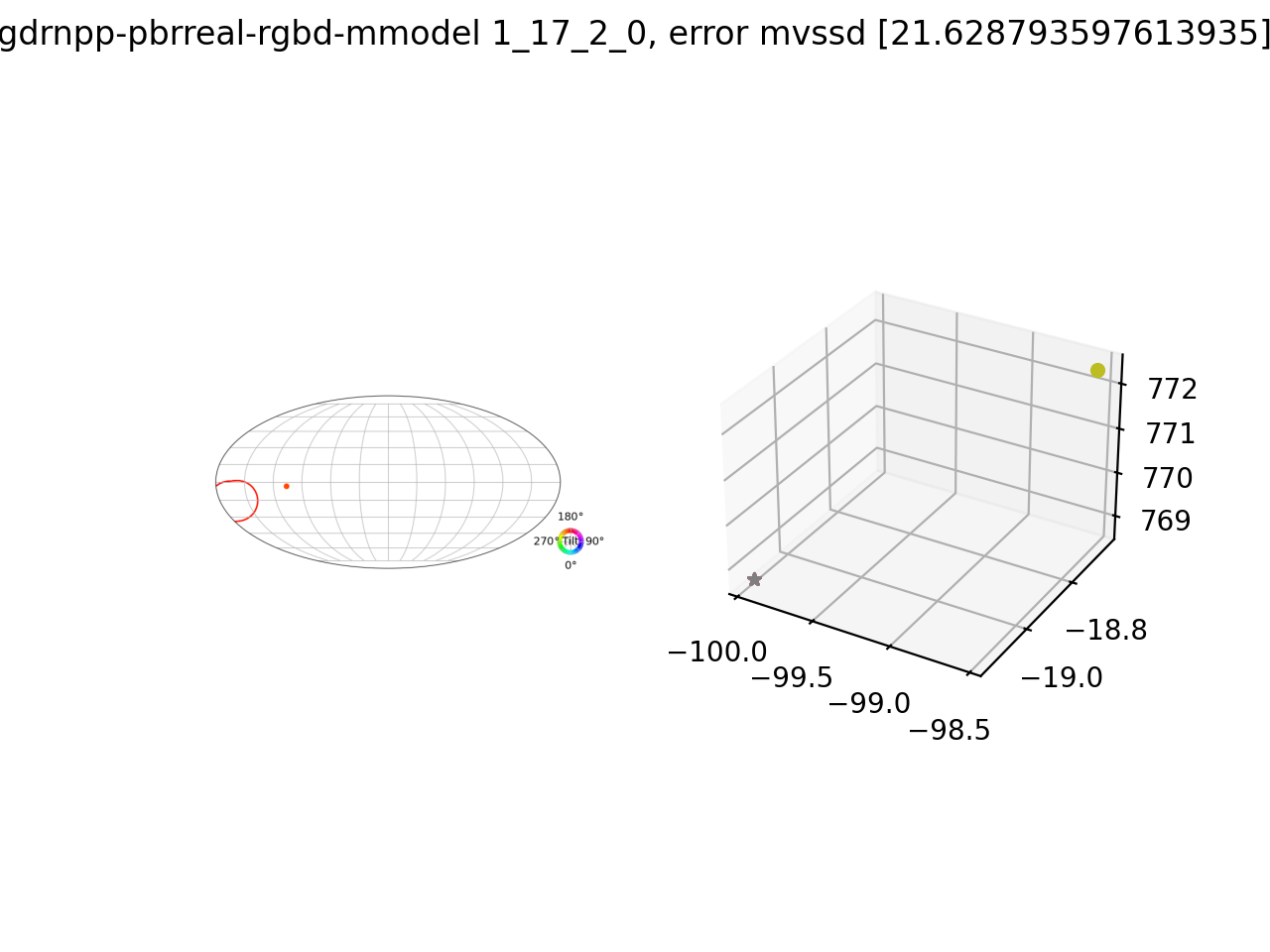}
\\
(a) BOP continuous symmetry pattern of Object 2~(envelop) and gdrnpp-pbrreal-rgbd-mmodel\textbf{v1.3} estimate~(plain circle). The circle belongs to the envelop, yielding a low \textbf{MSSD} error of 4.46.
&
(b) Our visual symmetry pattern (the much smaller envelop) and GDRNPP estimate (plain circle). The circle does not belong to the envelop anymore, the \textbf{MSSD} error becomes 21.62.
\\
\end{tabular}
\begin{tabular}{p{0.9\textwidth}}
\includegraphics[height=1.2in]{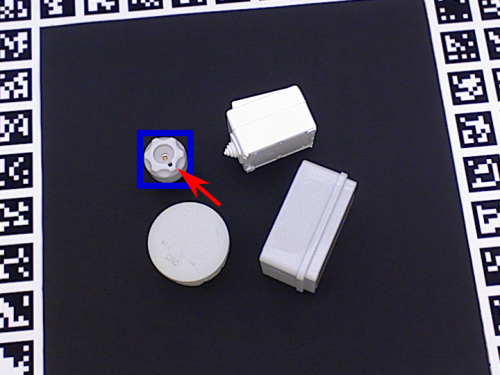}
\includegraphics[height=1.2in]{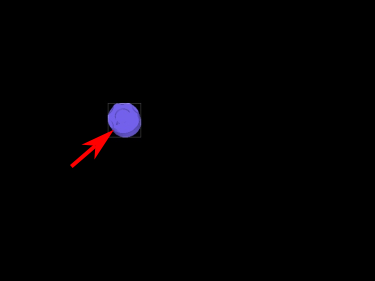}
\\
(c) Corresponding image for (a) and (b): T-LESS Scene 1, Image 17, Object 2 (in bounding box) and rendering of the pose (red arrow highlights disambiguating element).
\end{tabular}
\begin{tabular}{p{0.48\textwidth}p{0.48\textwidth}}
\includegraphics[trim={2.75cm 4.5cm 7.5cm 5cm},clip,height=1.2in]{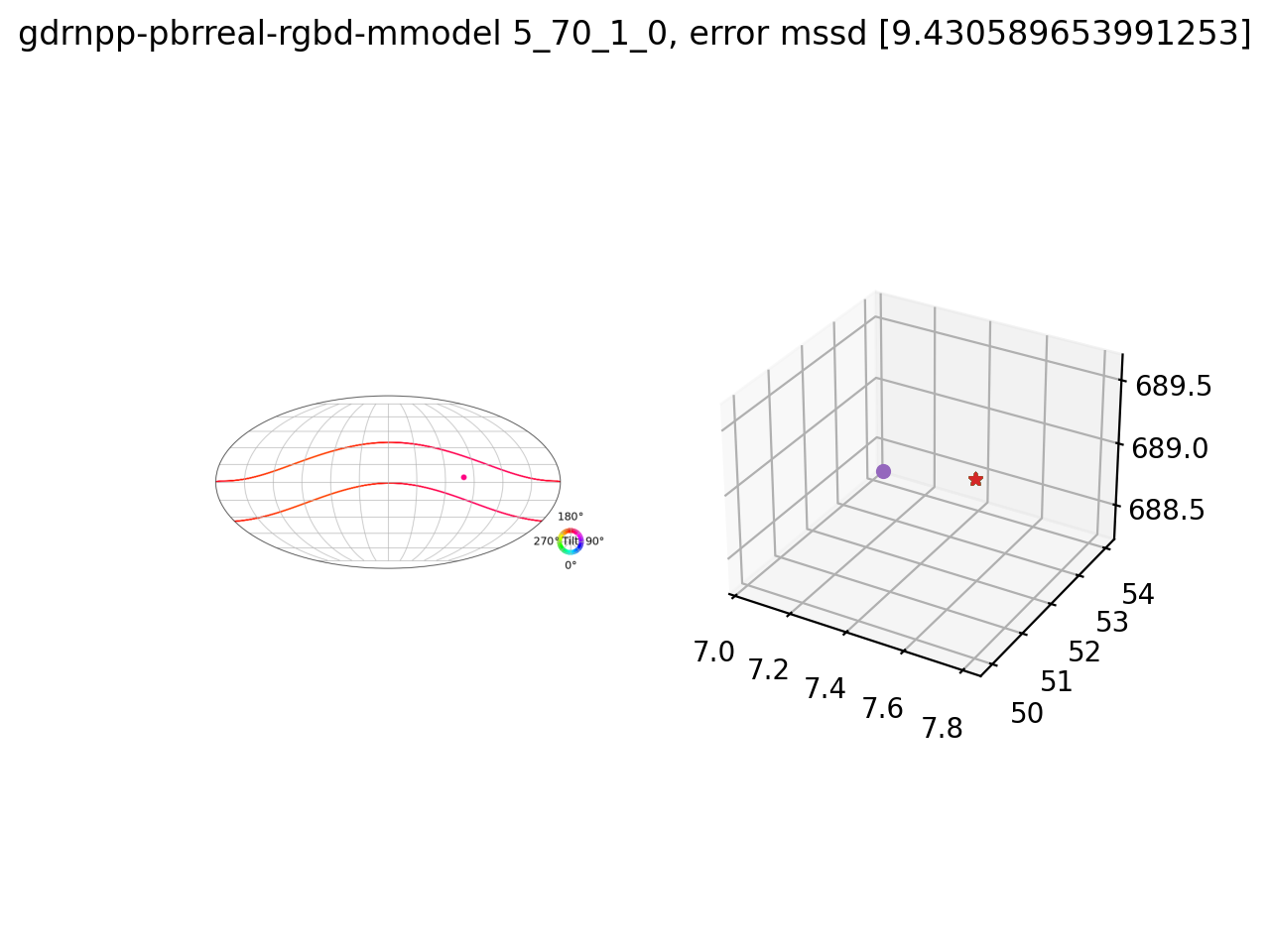}
&
\includegraphics[trim={2.75cm 4.5cm 7.5cm 5cm},clip,height=1.2in]{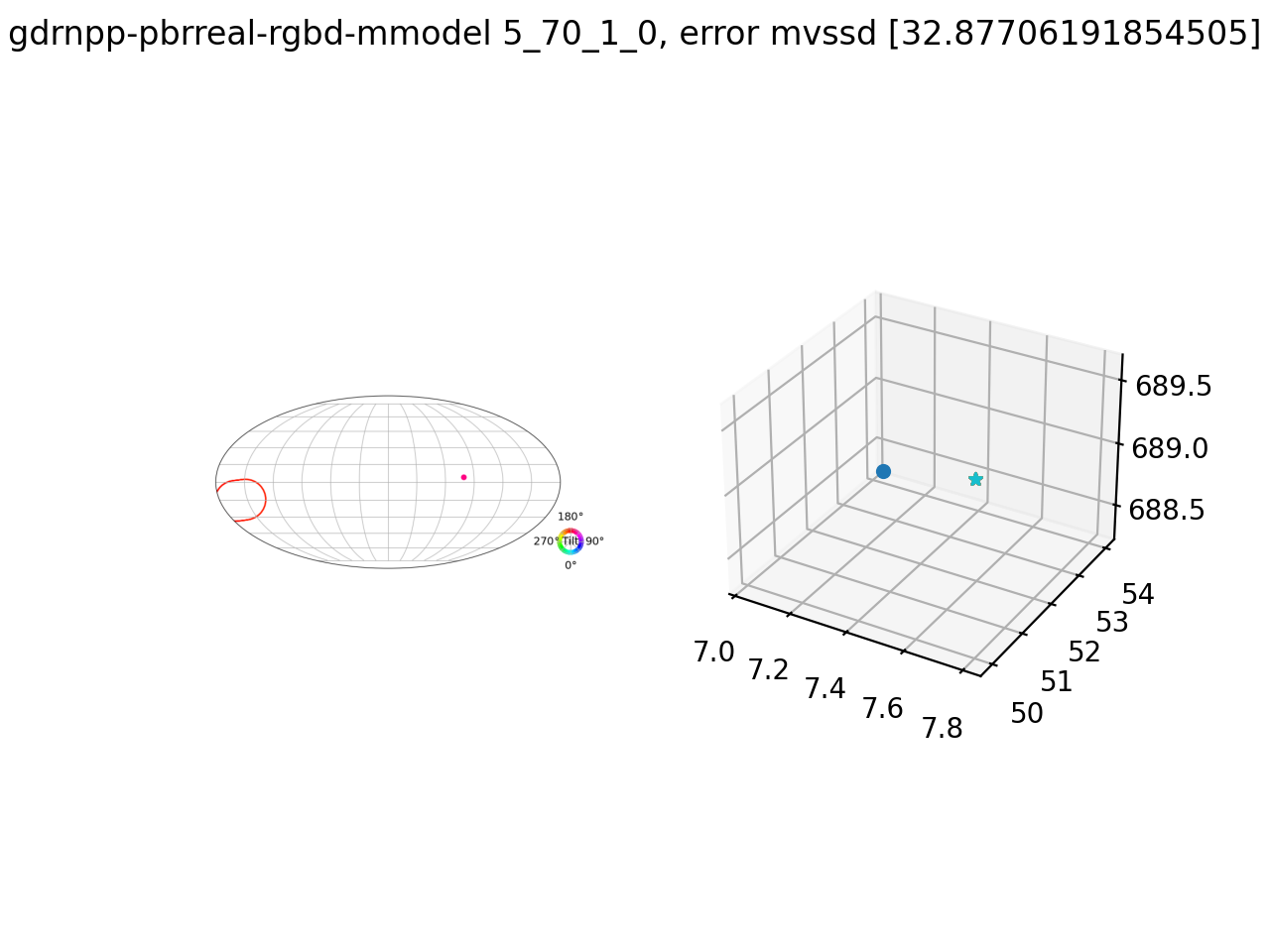}
\\
(d) BOP continuous symmetry pattern for Object 1~(envelop) and gdrnpp-pbrreal-rgbd-mmodel\textbf{v1.3} estimate~(plain circle). The circle belongs to the envelop, yielding a low \textbf{MSSD} error of 9.43.
&
(e) Our visual symmetry pattern (much smaller envelop) and GDRNPP estimate (plain circle). The circle does not belong to the envelop anymore, the \textbf{MSSD} error becomes 32.87.
\end{tabular}
\begin{tabular}{p{0.48\textwidth}p{0.48\textwidth}}
\includegraphics[trim={2.75cm 4.5cm 7.5cm 5cm},clip,height=1.2in]{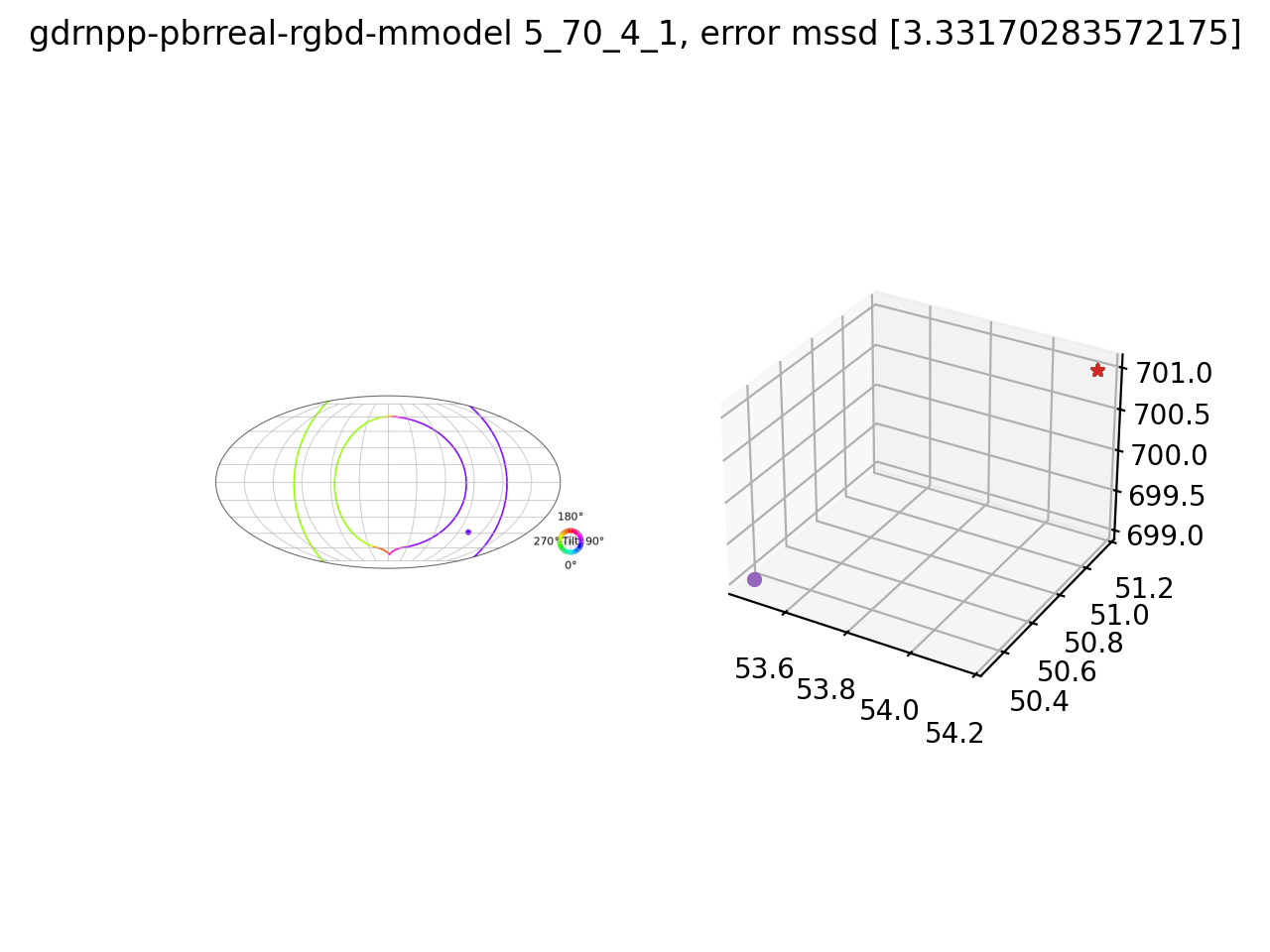}
&
\includegraphics[trim={2.75cm 4.5cm 7.5cm 5cm},clip,height=1.2in]{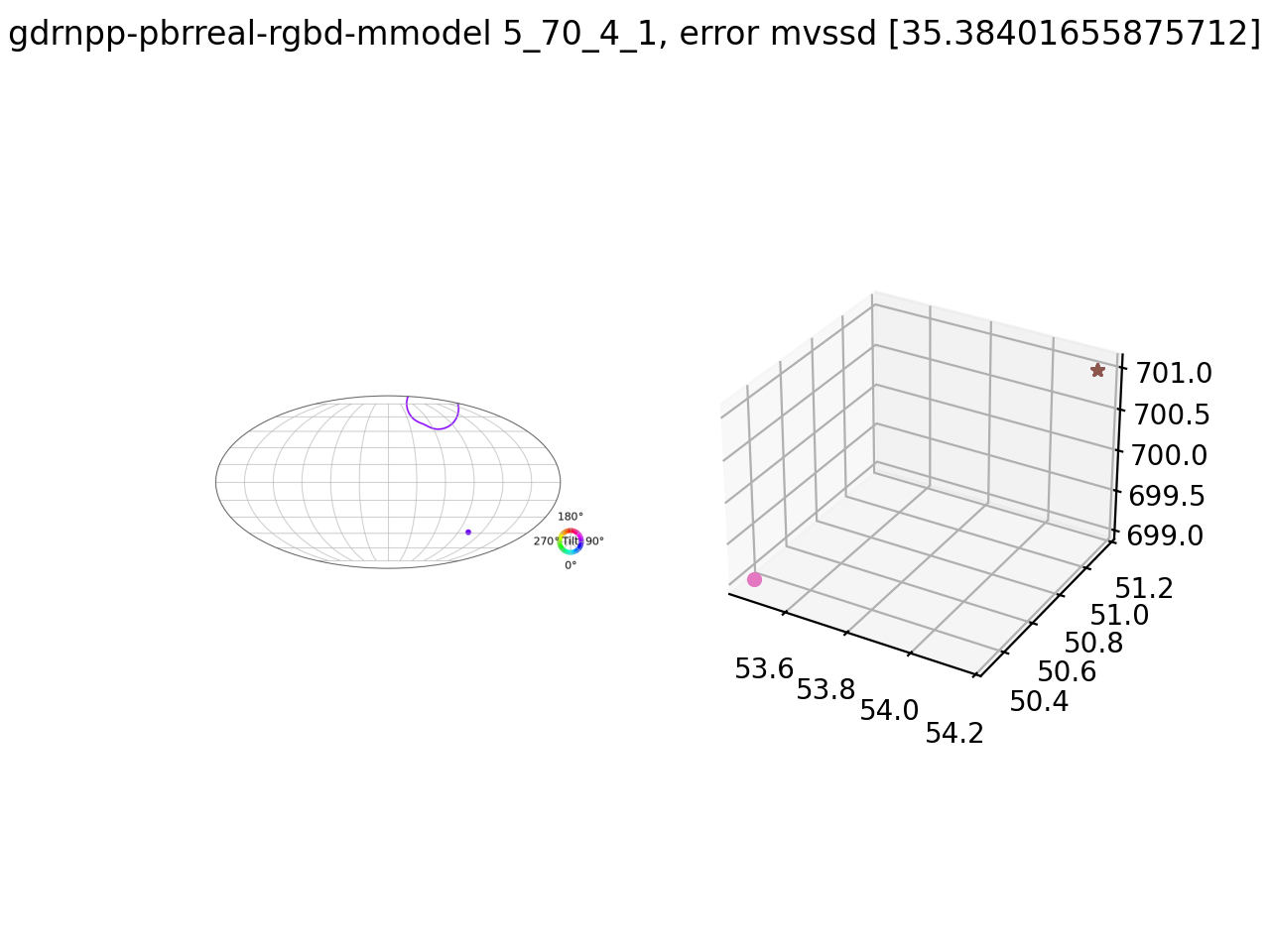}
\\
BOP continuous symmetry pattern of Object 4 (envelop) and gdrnpp-pbrreal-rgbd-mmodel\textbf{v1.3} estimate (plain circle). The circle belongs to the envelop, yielding a low \textbf{MSSD} error of 3.33.
&
Our visual symmetry pattern (much smaller envelop) and GDRNPP estimate (plain circle). The circle does not belong to the envelop anymore, the \textbf{MSSD} error becomes 35.38.
\end{tabular}
\begin{tabular}{p{0.9\textwidth}}
\includegraphics[height=1.2in]{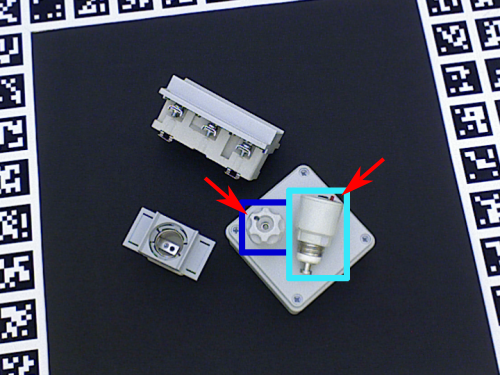}
\includegraphics[height=1.2in]{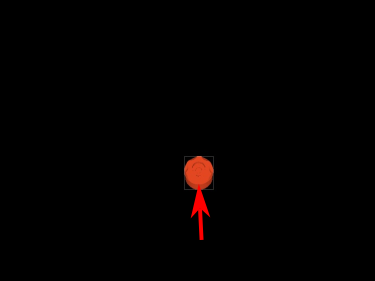}
\includegraphics[height=1.2in]{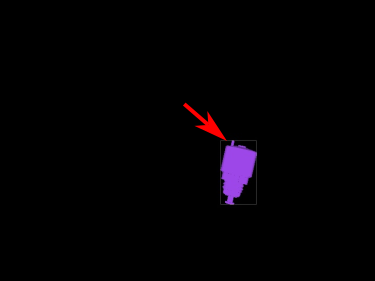}
\\
(h) Corresponding image for T-LESS Scene 5, Image 70, Objects 1 (d-e) and and 4 (f-g), in bounding boxes) and renderings of the poses (red arrow highlights disambiguating elements).
\end{tabular}
 \caption{\textbf{Impact of our annotations on Single Pose evaluation.} We show here cases where the estimates by a state-of-the-art method~(gdrnpp-pbrreal-rgbd-mmodel\textbf{v1.3}) produces fairly good \textbf{MSSD} errors when considering ground truth provided by BOP. For these cases, our more accurate ground truth yields worse \textbf{MSSD} errors, as it appears that the estimate belongs to the global symmetry pattern, but does not explain what is visible in the image.
 }
 \label{fig:symVSvsym}
\end{figure*}

Figure~\ref{fig:symVSvsym} illustrates why \textbf{MSSD} and \textbf{MSPD} changed for this method. It appears that, although gdrnpp-pbrreal-rgbd-mmodel\textbf{v1.3} estimates were close to the ground truth poses, its rotations were not precise. When they are evaluated against a symmetries pattern that is not precise, the evaluation appears correct. Our new ground truth shows that gdrnpp-pbrreal-rgbd-mmodel\textbf{v1.3} tends not to align correctly some of the objects. Hence the drop in performances.


\section{Visualizing results by SpyroPose~\cite{haugaard2023spyropose}}
\begin{figure}

  \vspace{-2cm}
  \includegraphics[width=0.45\textwidth]{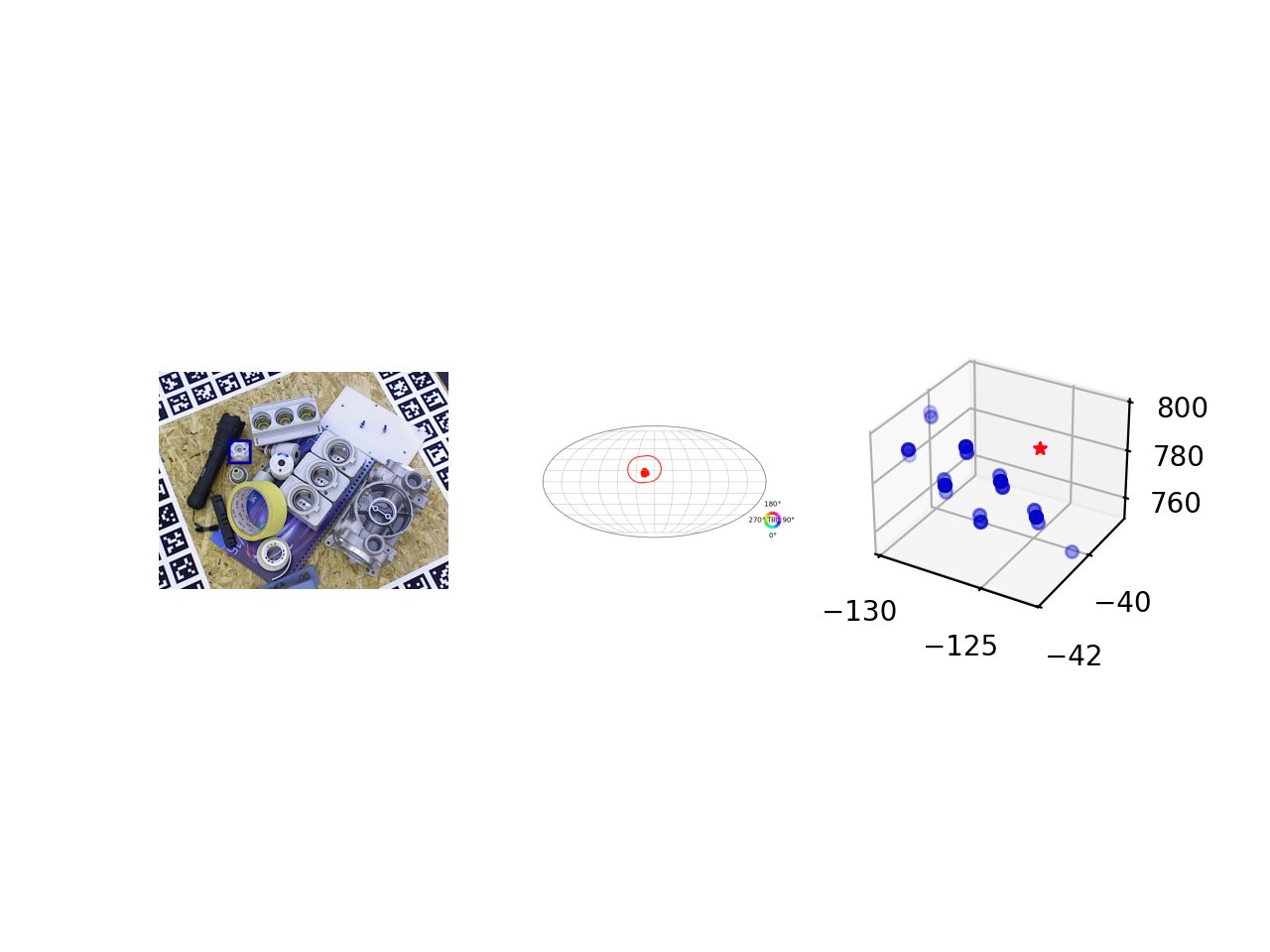}
  
  \vspace{-4cm}
  \includegraphics[width=0.45\textwidth]{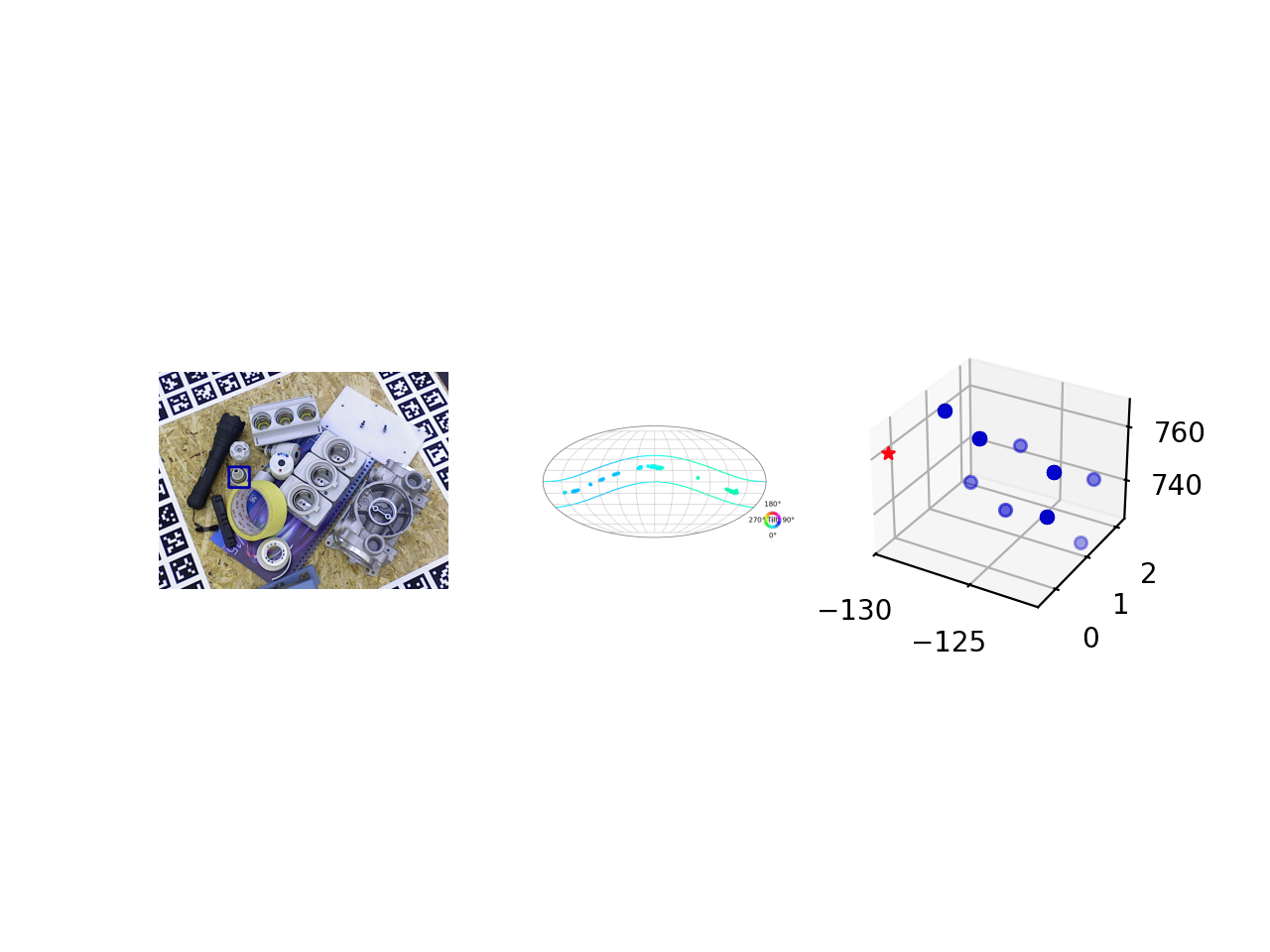}
  
  \vspace{-4cm}
  \includegraphics[width=0.45\textwidth]{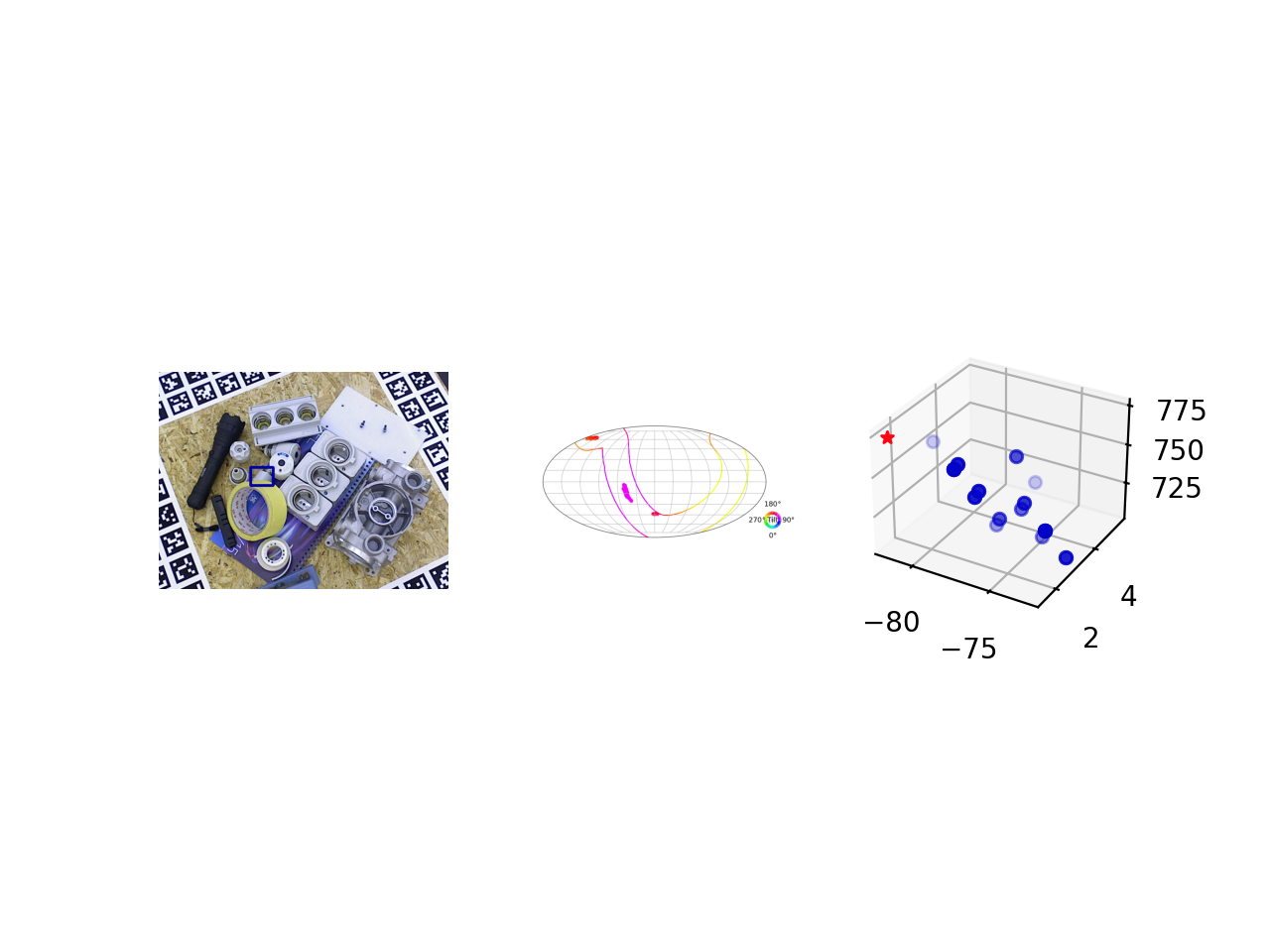}
  
  \vspace{-4cm}
  \includegraphics[width=0.45\textwidth]{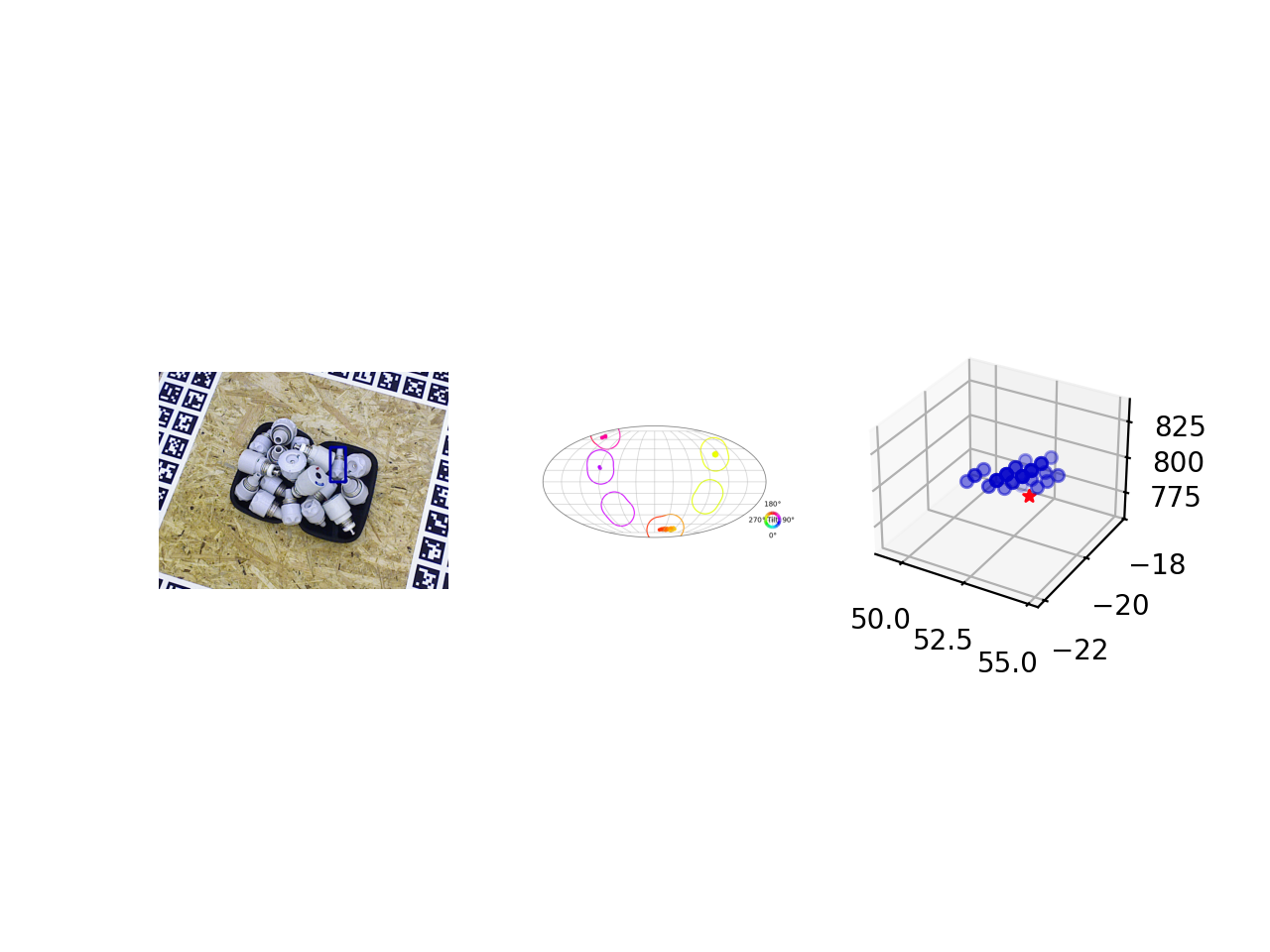}
  
  \caption{\textbf{Illustration of SpyroPose~\cite{haugaard2023spyropose} results on T-LESS.}. We show the distribution estimates for the 3 instances of Object 1 (in the bounding box). The ground truth distribution is displayed as an envelop for the rotation part and as red star for the translation part.} 
  \label{fig:spyro_res}
\end{figure}

We display some of SpyroPose distribution estimates against our ground truth in Figure~\ref{fig:spyro_res} on T-LESS. For the case of the three instances of Object 1, SpyroPose correctly retrieves the single mode for Instance 1. The continuous symmetry of Instance 2 is partially retrieved.

\section{Visualizing results by LiePose diffusion~\cite{hsiao2024confronting}}

\begin{figure}

  \vspace{-2cm}
  \includegraphics[width=0.45\textwidth]{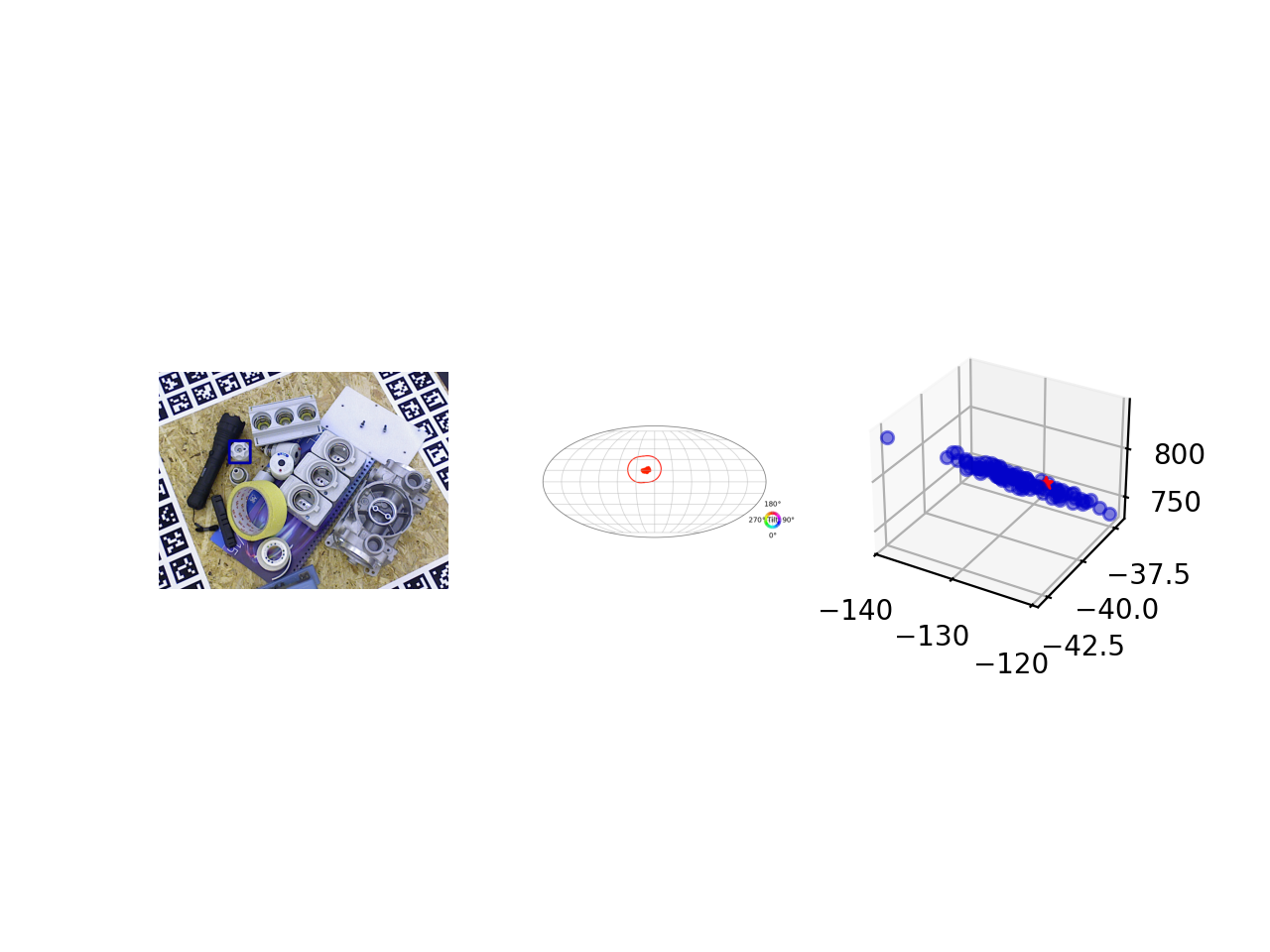}
  
  \vspace{-4cm}
  \includegraphics[width=0.45\textwidth]{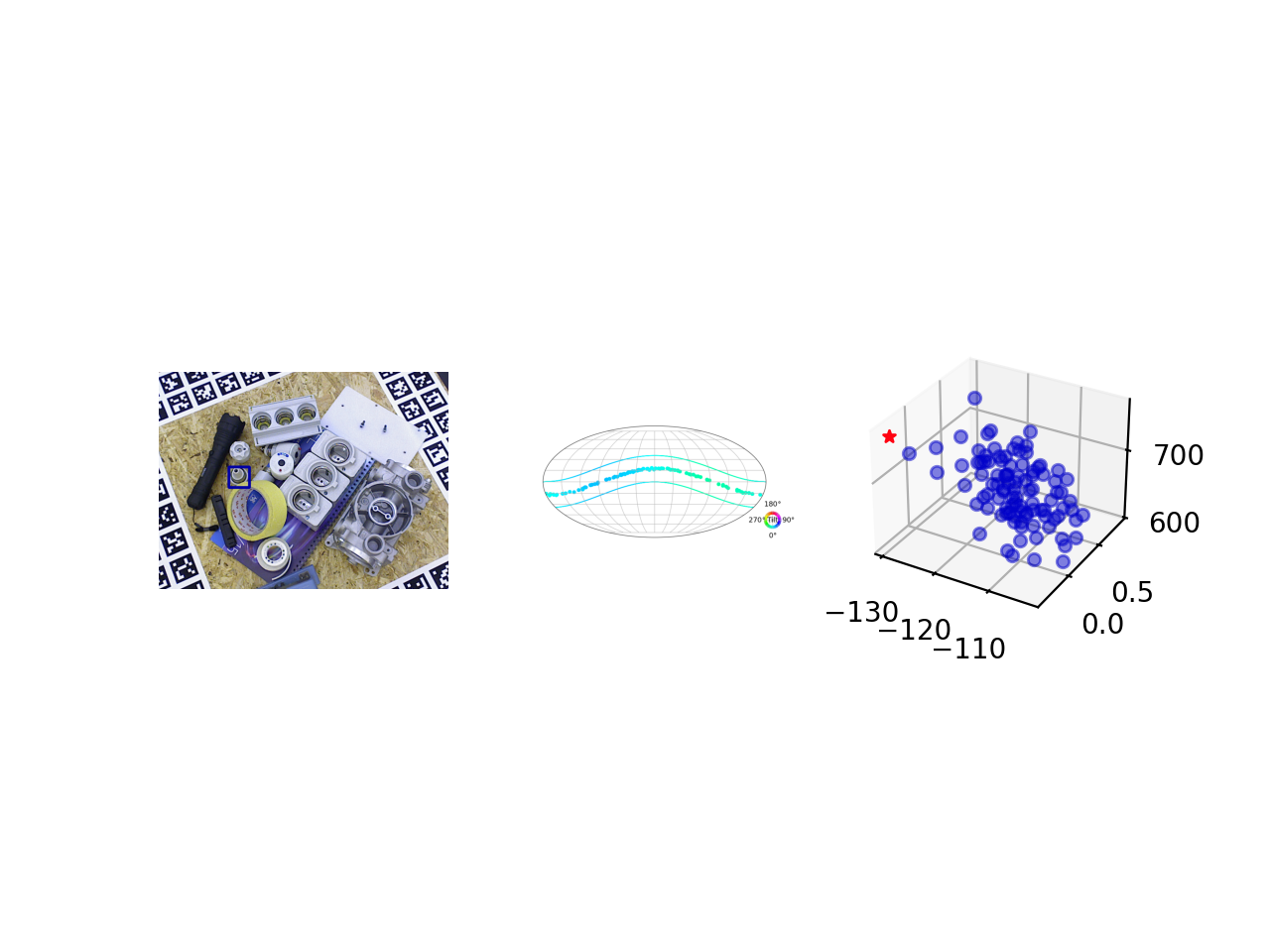}
  
  \vspace{-4cm}
  \includegraphics[width=0.45\textwidth]{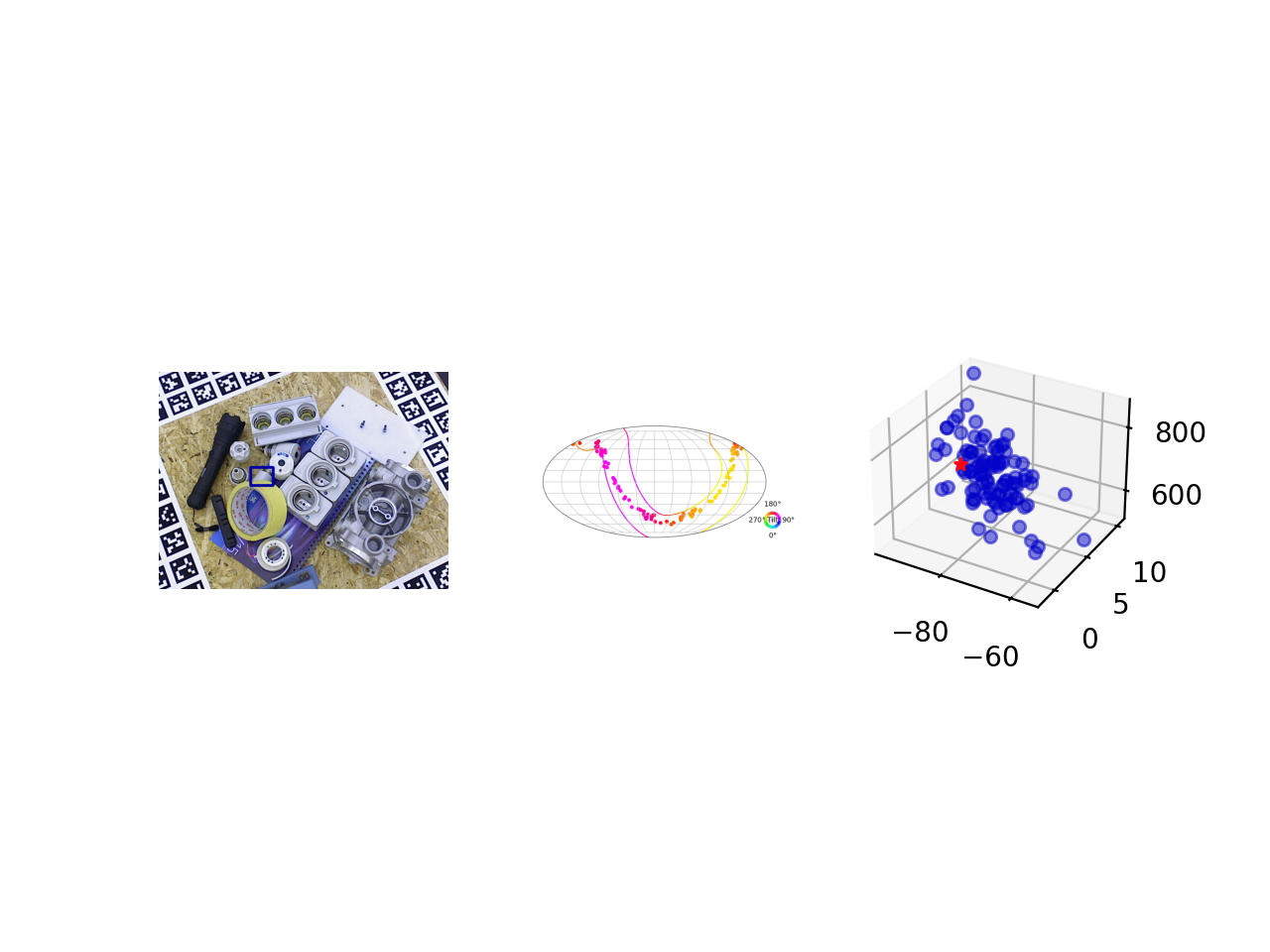}
    
  \vspace{-4cm}
  \includegraphics[width=0.45\textwidth]{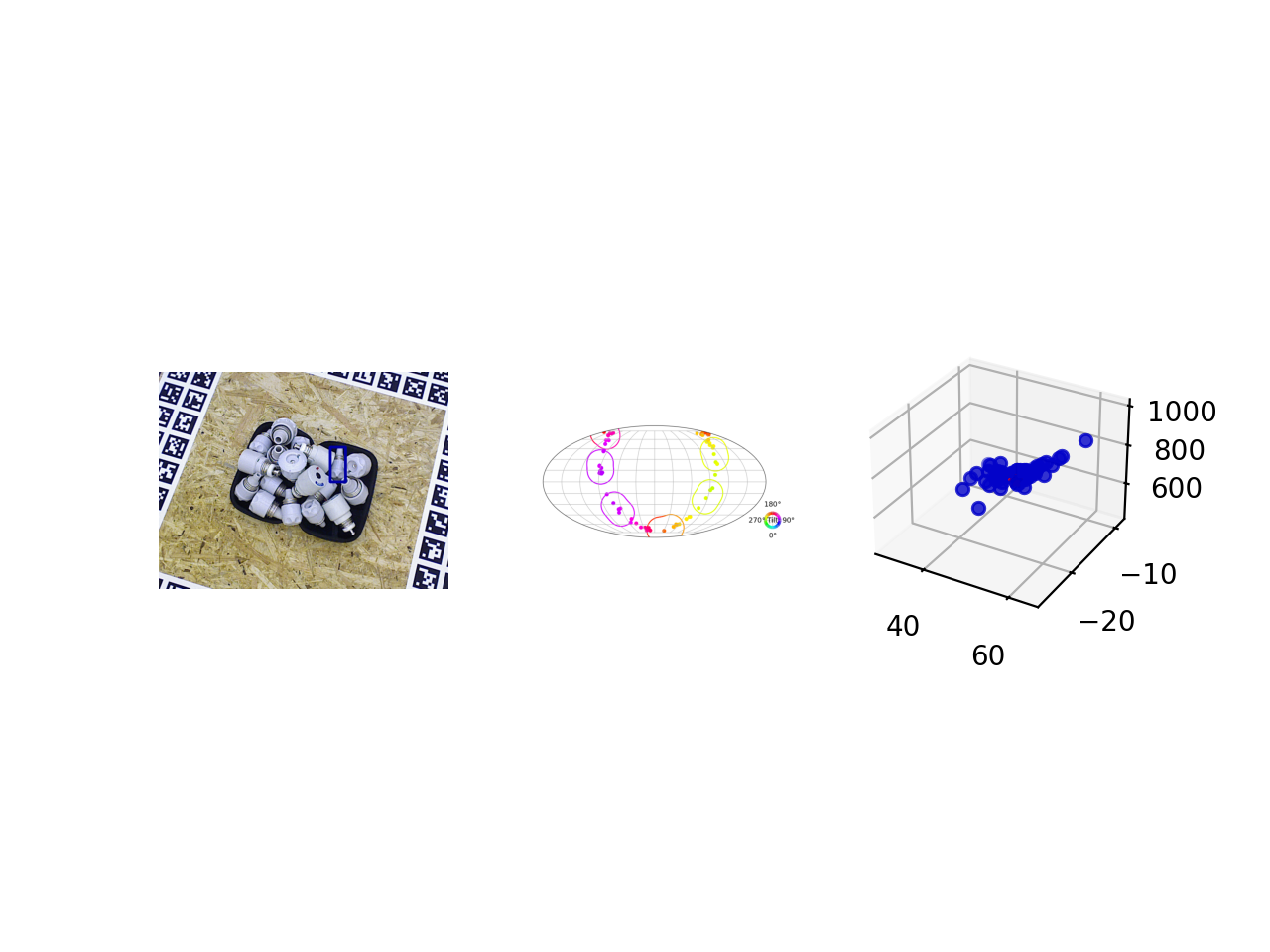}
  
  \caption{\textbf{Illustration of LiePose~\cite{hsiao2024confronting} results on T-LESS.}. We show the distribution estimates for the 3 instances of Object 1 (in the bounding box). The ground truth distribution is displayed as an envelop for the rotation part and as red star for the translation part.} 
  \label{fig:lie_res}
\end{figure}

We show some of LiePose~\cite{hsiao2024confronting} distribution estimates against our ground truth in Figure~\ref{fig:lie_res} on T-LESS. Similarly to SpyroPose~\cite{haugaard2023spyropose}, LiePose~\cite{hsiao2024confronting} is able to retrieve the single mode of Instance 1, but gets better results when estimating continuous distributions.

\begin{figure*}
  \centering

  \includegraphics[width=\textwidth]{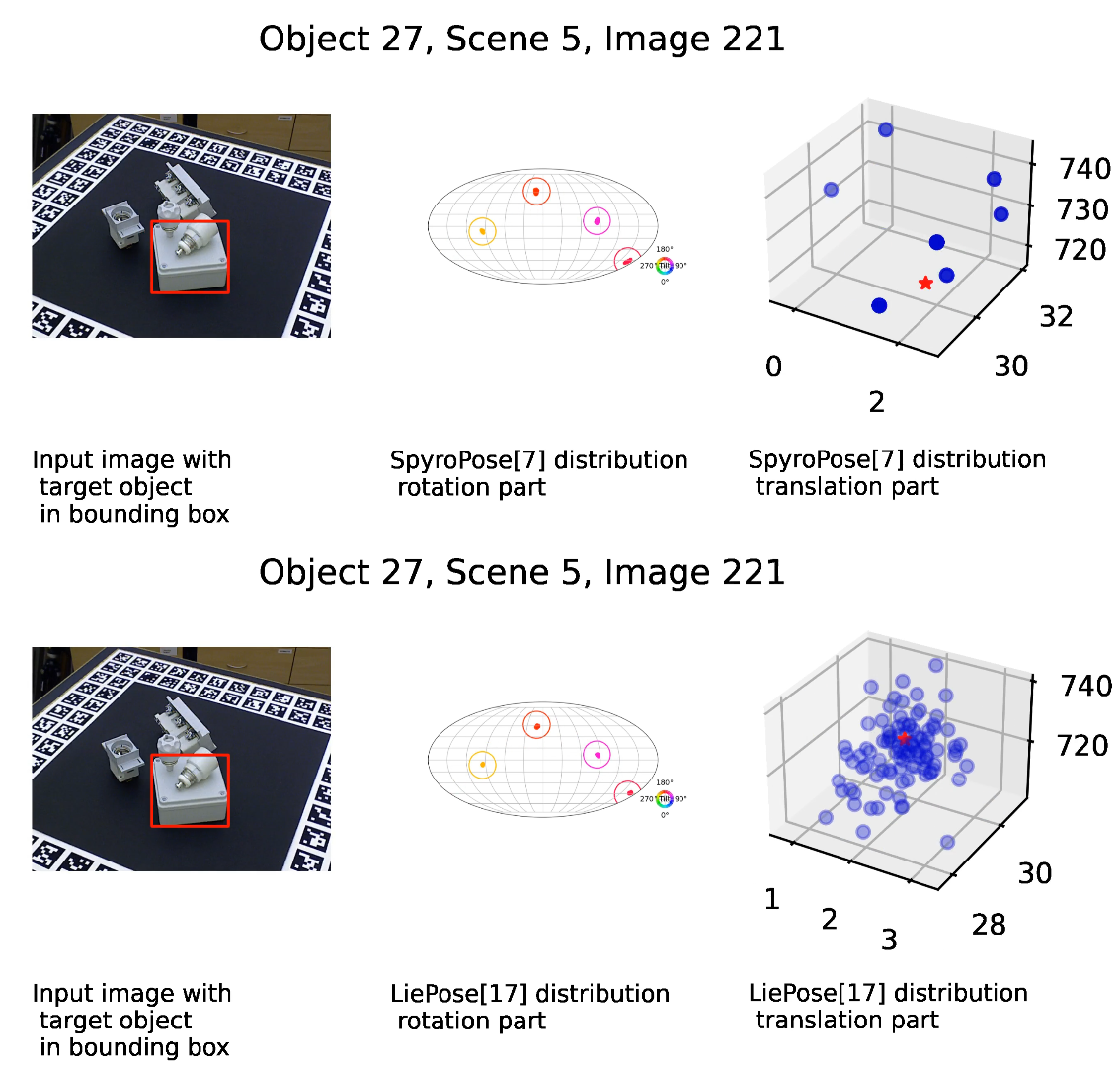}
    \caption{\textbf{Visualizing SpyroPose~\cite{haugaard2023spyropose} (top row) and LiePose~\cite{hsiao2024confronting} (bottom row) distribution results for object 27 (four rotation modes).} Each example features an object of interest, in the bounding box in the left image, and the methods distribution estimation, split between the rotation part (center) and translation part (right). Both methods are able to retrieve the four rotation modes of the object. The envelop in the rotation part represents our BOP-Distrib annotation. We provide much more examples in the accompanying video.}
  \label{fig:distComp1}
\end{figure*}

\begin{figure*}
  \centering
  
  \includegraphics[width=\textwidth]{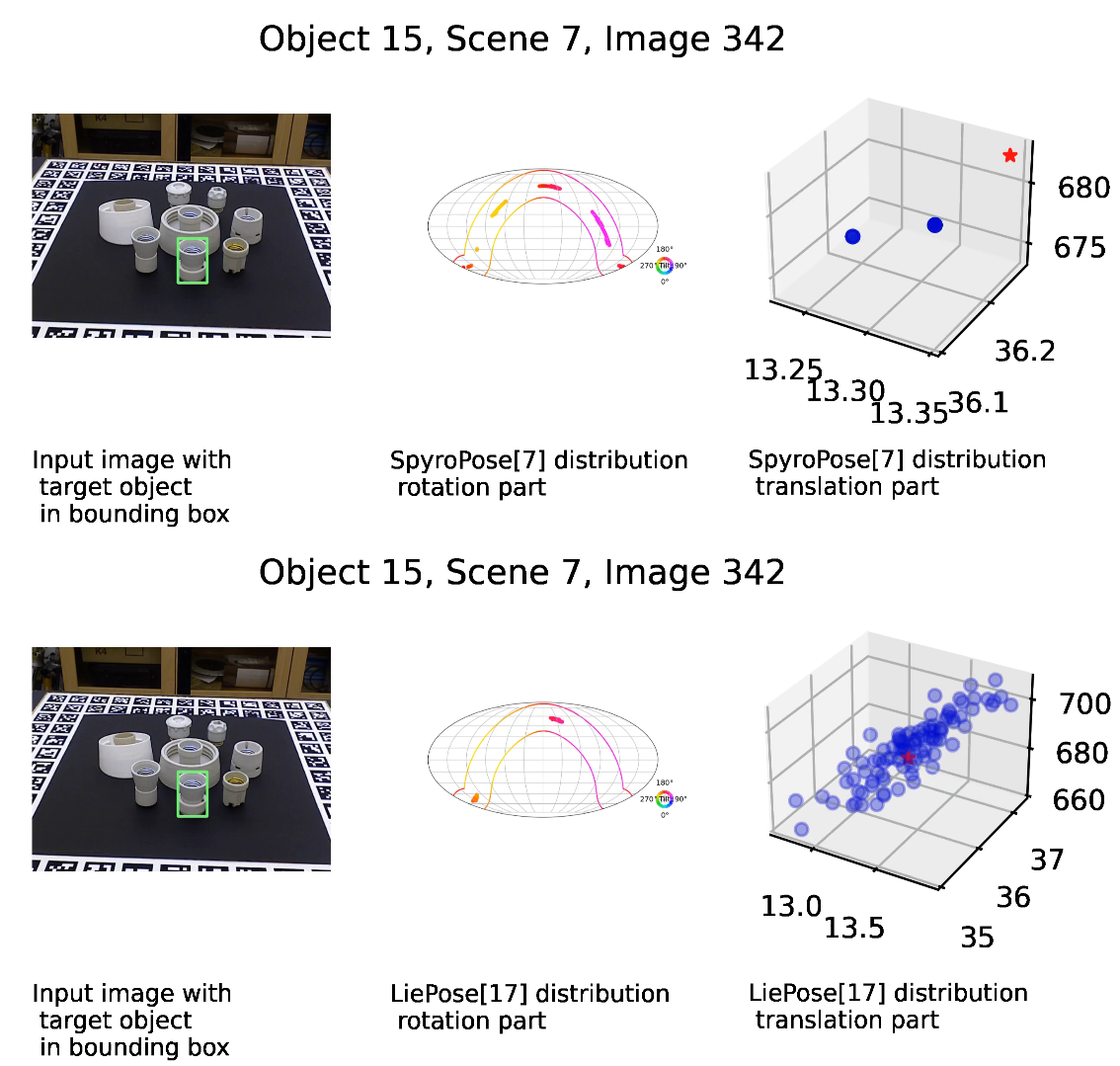}
    \caption{\textbf{Visualizing SpyroPose~\cite{haugaard2023spyropose} (top row) and LiePose~\cite{hsiao2024confronting} (bottom row) distribution results for object 15 (continuous rotation).} Each example features an object of interest, in the bounding box in the left image, and the methods distribution estimation, split between the rotation part (center) and translation part (right). Both methods fail to generate the target continuous rotation, although SpyroPose produces more correct rotations. The envelop in the rotation part represents our BOP-Distrib annotation. We provide much more examples in the accompanying video.}
  \label{fig:distComp2}
\end{figure*}

\begin{figure*}
  \centering
  \includegraphics[width=\textwidth]{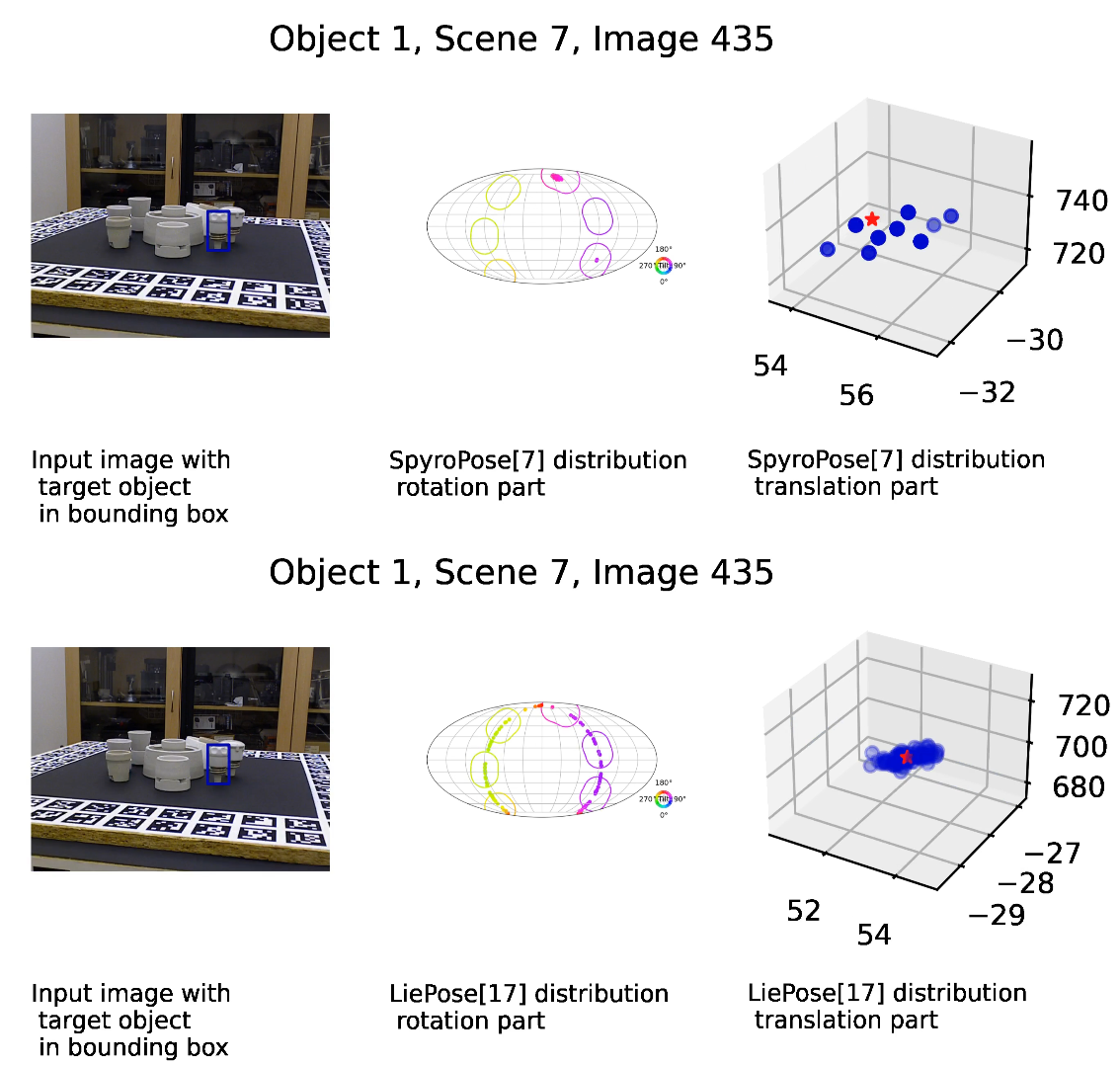}
  \caption{\textbf{Visualizing SpyroPose~\cite{haugaard2023spyropose} (top row) and LiePose~\cite{hsiao2024confronting} (bottom row) distribution results for object 1 (six rotation modes).} Each example features an object of interest, in the bounding box in the left image, and the methods distribution estimation, split between the rotation part (center) and translation part (right). For this case of six rotations modes, SpyroPose is able to retrieve only two of them, whereas LiePose tends to a continuous distribution, thus generating false rotations. The envelop in the rotation part represents our BOP-Distrib annotation. We provide much more examples in the accompanying video.}
  \label{fig:distComp3}
\end{figure*}

\begin{figure*}
  \centering
  \includegraphics[width=\textwidth]{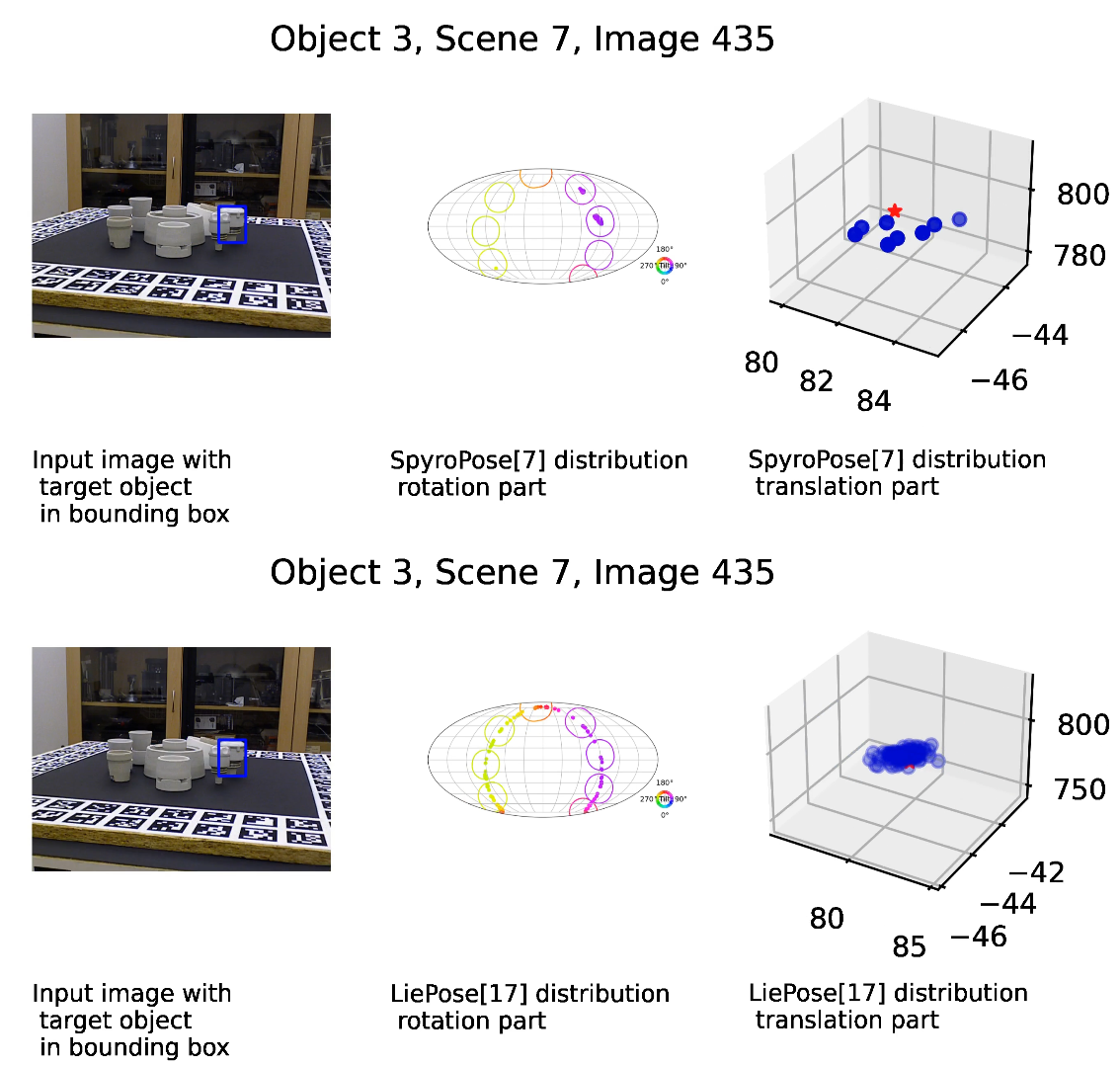}
  \caption{\textbf{Visualizing SpyroPose~\cite{haugaard2023spyropose} (top row) and LiePose~\cite{hsiao2024confronting} (bottom row) distribution results for object 3 (eight rotation modes).} Each example features an object of interest, in the bounding box in the left image, and the methods distribution estimation, split between the rotation part (center) and translation part (right). For this case of eight rotations modes, SpyroPose is able to retrieve only two of them, whereas LiePose tends to a continuous distribution, thus generating false rotations. The envelop in the rotation part represents our BOP-Distrib annotation. We provide much more examples in the accompanying video.}
  \label{fig:distComp4}
\end{figure*}

Figures~\ref{fig:distComp1}, \ref{fig:distComp2}, \ref{fig:distComp3} and \ref{fig:distComp4} compare SpyroPose~\cite{haugaard2023spyropose} and LiePose~\cite{hsiao2024confronting} results on objects with discrete and continuous symmetries. SpyroPose~\cite{haugaard2023spyropose} rotations tends to be more precise than LiePose~\cite{hsiao2024confronting}, but misses some of the modes. LiePose~\cite{hsiao2024confronting} tends to estimate continuous symmetries when the image produces discrete ones. These images are taken from a video compilation of SpyroPose~\cite{haugaard2023spyropose} and LiePose~\cite{hsiao2024confronting} results also provided as supplementary material (\url{BOP\_Distrib\_id8513\_supp\_distribution\_comparison\_SpyroPose\_LiePose.mp4}). Scenes with single instance of objects have been chosen, to facilitate visualization. We invite the reader to stop on some frames and check the differences in the estimates. Our ground truth distribution is displayed as the envelop for the rotation part and as red stars for the translation part.



\section{Metrics for evaluation: extended discussion}
Rotation and translation errors are model-independent. \cite{hodanEvaluation6DObject2016}, which led to BOP, states that \textit{"fitness of object surface alignment is the main indicator of object pose quality, model-dependent pose error functions should be therefore preferred."} However, translation and rotation errors, as definer by~\cite{hodanEvaluation6DObject2016}) with \autoref{eq:hodanTranslationError} \& \autoref{eq:hodanRotationError} can be exploited with our definitions of precision and recall over distributions (\autoref{eq:precisionDist} \& \autoref{eq:recallDist}).
\begin{align}
\mathbf{d}_{\text{TE}}(\hat{\textbf{t}}, \bar{\textbf{t}}) = \Vert\bar{\textbf{t}} - \hat{\textbf{t}}\Vert_2,
\label{eq:hodanTranslationError}
\end{align}
\begin{align}
\mathbf{d}_{\text{RE}}(\hat{\textbf{R}}, \bar{\textbf{R}}) = \text{arccos} \left( \text{Tr}(\hat{\textbf{R}}\bar{\textbf{R}}-1)/2\right),
\label{eq:hodanRotationError}
\end{align}
where $\hat{\textbf{R}}$ and $\hat{\textbf{t}}$ are respectively the ground truth rotation and translation, and where $\bar{\textbf{R}}$ and $\bar{\textbf{t}}$ are respectively the estimated rotation and translation.

Table~\ref{tab:precisionRecallOverRotTrans} is a reprocessing of pose distribution estimation method results, with precision and recall over distribution, as defined by \autoref{eq:precisionDist} \& \autoref{eq:recallDist}, with translation and rotation errors from \autoref{eq:hodanTranslationError} \& \autoref{eq:hodanRotationError}.
\begin{table}
 \begin{center}
 \begin{tabular}{c c c c c}
 \toprule
  Methods & $\mathbf{P_{RE}}$ & $\mathbf{R_{RE}}$ & $\mathbf{P_{TE}}$ & $\mathbf{R_{TE}}$ \\
 \midrule
     SpyroPose~\cite{haugaard2023spyropose}     &    \textbf{73.2}        &       69.1      &  43 &  66.9     \\
     LiePose~\cite{hsiao2024confronting}  &    68        &      \textbf{91.1}       &      \textbf{46.2} &  \textbf{92.9}     \\
  \bottomrule
 \end{tabular}
 \end{center}
 \caption{\textbf{Precision/Recall over distribution for separate rotation and translation errors.} We reprocess methods pose distribution estimates with decoupled rotation and translation errors.}
 \label{tab:precisionRecallOverRotTrans}
\end{table}

\section{Downstream Tasks Discussion}
Our work implication on downstream task is two-fold.
First, as highlighted by the BOP ranking changes in table 2, removing erroneous poses from  the BOP ground truth implies a more reliable  performance evaluation. Indeed, as illustrated by Fig. 11 of supp. mat., performance variations are related to poses that the current BOP ground-truth unduly classify as valid but are properly classified as invalid by our ground truth. This is due to overly lax ground truth annotation: some images are  annotated as having multiple pose solutions due to the object symmetries whereas disambiguating elements breaking those symmetries are visible in the image. For downstream tasks, such as grasping or Augmented Reality, reliable ranking permits to choose the real best performer, and thus a higher success rate of the task.

\begin{figure*}
  \includegraphics[width=\textwidth]{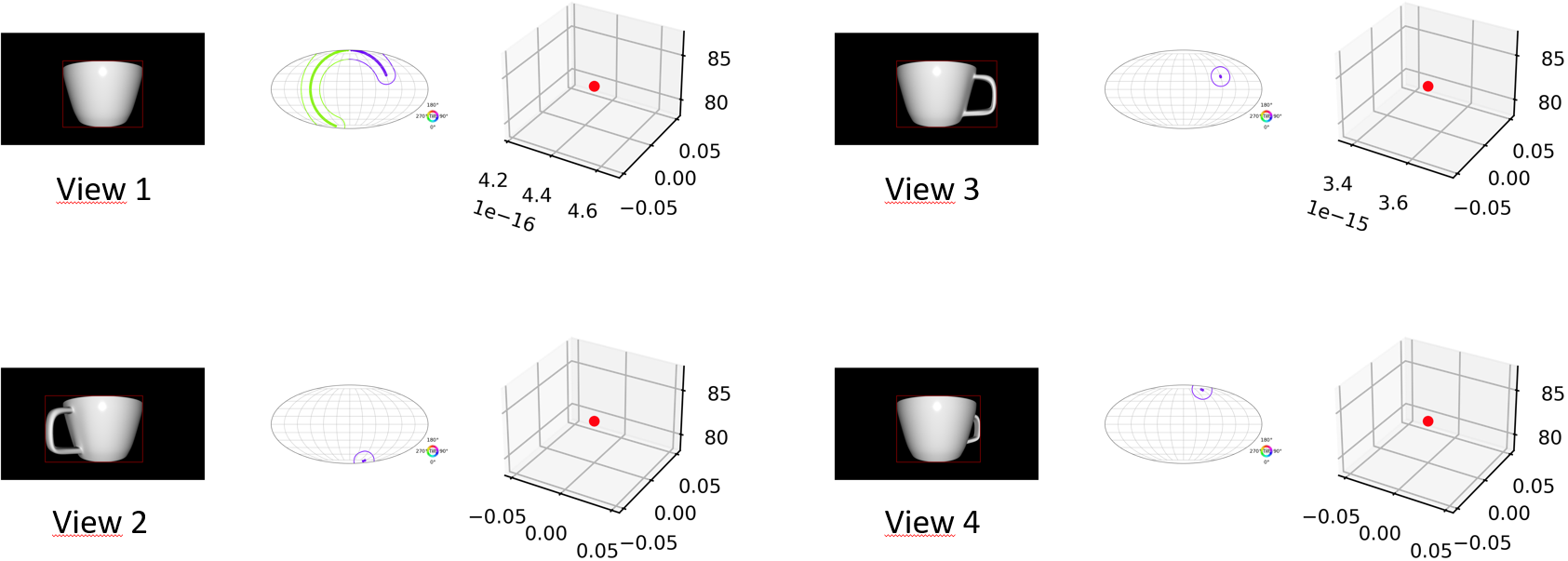}
  \caption{\textbf{Additional experiment for mug grasping task.} We display multiple views of mug, associated to our pose distribution annotation. We are interested at pose estimation downstream tasks. We take the example of robotic grasping. We want the robot to grasp the mug by the handle. On view 1, the pose distribution is multi-modal, \ie the object pose is ambiguous and the handle position in space cannot be accessed. For view 2, 3 and 4, the pose distribution is uni-modal, the image allows to estimate the pose of the mug without ambiguity. In such case, the downstream task of robotic handle grasping becomes feasible. Now, when it comes to evaluating pose distribution estimation methods against our ground truth, our recall metric evaluates the capacity of the estimation method to retrieve multi-modal distributions. This metric is a key indicator of pose distribution estimation methods for downstream tasks usability.} 
  \label{fig:mug_downstream}
\end{figure*}

The second implication on downstream tasks is related to Table 3, \ie evaluating the ability of a  method to determine the complete distribution of poses that explain the observed image. For applications such as grasping or Augmented Reality, if the object includes disambiguating elements, it implies that only one pose is valid for the task (\eg a robot that should grab a mug at a specific position on the handle). However, if the observed image can be explained by multiple pose, it implies that disambiguating elements are not visible (\eg the handle of the mug is not visible) and the downstream task is impossible to achieve from this unique viewpoint. A method that outputs the full distribution of poses provides the downstream task with the ability to determine if the task can be achieved (case of uni-modal distribution) or not (case of multi-modal solution). In such situation, a method that outputs a maximum of one pose does not permit to determine if the task can be achieved or not. Moreover, in such situation the complete distribution can help to determine the next best viewpoint to make the task feasible. Our distribution recall metric~\ref{eq:recallDist} evaluates the capacity of the estimation method to retrieve multi-modal distributions. This metric is a key indicator of pose distribution estimation methods for downstream tasks usability Figure~\ref{fig:mug_downstream} illustrates the robotic mug handle grabbing case.

\section{Using Our Per-Image Annotations for Other Pose Estimation Datasets}

Among BOP datasets, ITODD~\cite{drostIntroducingMVTecITODD2017} and HomeBrewedDB~\cite{kaskmanHomebrewedDBRGBDDataset2019} are the two other presenting object symmetries. They could easily be processed by our method, however their ground truth poses, needed as input to our method, are not public.

\begin{figure}
  \includegraphics[width=0.45\textwidth]{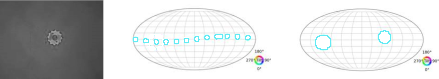}
  \includegraphics[width=0.12\textwidth]{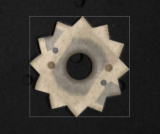}
  \caption{\textbf{Illustration of ITODD symmetries on the validation set (with public GT).} For the star image (left, first row), BOP symmetries display 12 rotation modes (middle), whereas our annotation method keeps only 2 rotation modes (right), which align the two holes (size was set to one over the number of modes, hence the bigger modes on the right). We also show (second row left) a pose estimate of GPose~\cite{gpose2023}, ranked first at BOP 2023, for the star object overlayed on its image. We observe that the holes are not correctly aligned. \textbf{MSPD} and \textbf{MSSD} metrics validate this estimate, whereas \textbf{MSPD} and \textbf{MSSD} metrics with our new annotations would have penalized it.} 
  \label{fig:itodd_syms}
\end{figure}

We give here an illustration of the interest to reprocess their symmetries patterns. We take ITODD's small validation set, for which the ground truth is public. Figure~\ref{fig:itodd_syms} presents our result for the star object new ground truth, as well as one case of the current best performer for Single Pose estimation (gpose2023~\cite{gpose2023}) failing to align the holes. In the current version of BOP evaluation, this pose estimation is validated. With our annotations, it would be penalized.
New efforts at proposing challenging pose estimation datasets, such as~\cite{zhang2024omni6d}, could be processed by our method. However, texture disambiguate a lot of the 3D models, and the scenes do not present much occlusion between objects.

\section{Discussion on Alignist~\cite{vutukur2024alignist}}
\label{suppmat:alignistDiscussion}
Alignist~\cite{vutukur2024alignist} proposes to estimate rotation distribution for ambiguous object shapes from images. It was the first method that introduced a solution for supervising the training with a pseudo-ground truth generated rotation distribution. To do so, it resorts to a precomputation of such rotation distibution based on ground truth pose, rotation sampling, and SDF (Signed Distance Function) and Surfemb~\cite{haugaardSurfEmbDenseContinuous2022} features comparison. This precomputation is performed on renderings of single objects, and used to train a double MLP network to infer these distributions. The translation part of the pose is not considered. The test of the method on T-LESS~\cite{hodanTLESSRGBDDataset2017} is conducted following Gilitschenski~\cite{gilitschenski2019deep} protocol: it only processes single isolated objects on black background, and is evaluated with log likelihood.

In contrast, our annotation procedure does not rely on SurfEmb~\cite{haugaardSurfEmbDenseContinuous2022} comparisons which results are not guaranteed but it uses geometrical comparisons (see \autoref{eq:epsymVM}). Moreover and unlike Alignist~\cite{vutukur2024alignist}, our annotation procedure has a rejection mechanism for false visible points and false occluded points, as illustrated in Section~\ref{suppmat:discussionLimitEps} (see Section~\ref{subsec:refine}). This point is crucial to be able to generate a proper ground truth annotation. Finally, our approach does not need to retrain SurfEmb~\cite{haugaardSurfEmbDenseContinuous2022} to annotate a new dataset.

Alignist~\cite{vutukur2024alignist} and other pose distributions estimation methods would benefit from our more accurate pose distributions for their trainings.
Finally, Alignist~\cite{vutukur2024alignist} method would benefit from our evaluation framework (see Section~\ref{subsec:expDistpose}). For that though, Alignist~\cite{vutukur2024alignist} would need to be tested on the full T-LESS~\cite{hodanTLESSRGBDDataset2017} test set (and not just Gilitschenski~\cite{gilitschenski2019deep} protocol that excludes external occlusions). We could not conduct such tests as the codes and results are not public.

\end{document}